\documentclass{article}

\PassOptionsToPackage{numbers,sort&compress}{natbib}
\usepackage[preprint]{neurips_2026}

\usepackage[utf8]{inputenc} 
\usepackage[T1]{fontenc}    
\usepackage{hyperref}       
\usepackage{url}            
\usepackage{booktabs}       
\usepackage{amsmath}        
\usepackage{amsfonts}       
\usepackage{nicefrac}       
\usepackage{microtype}      
\usepackage{xcolor}         
\usepackage{graphicx}       
\usepackage{subcaption}     
\usepackage{rotating}       
\usepackage[ruled,vlined,linesnumbered]{algorithm2e}
\SetAlgoCaptionSeparator{:}
\SetKwInput{KwInput}{Input}
\SetKwInput{KwOutput}{Output}
\SetKw{Return}{Return}

\IncMargin{1.5em}
\SetNlSkip{0.4em}

\title{PHIDA: Persistence-Guided Node-to-Cluster Mapping for Online Clustering}

\newcommand{\phidaauthorblock}{%
  \begin{minipage}{0.96\textwidth}
  \centering
  Naoki Masuyama$^{1}$\footnotemark \quad
  Yusuke Nojima$^{1}$ \quad
  Stefan Wermter$^{2}$ \\
  Yuichiro Toda$^{3}$ \quad
  Hisao Ishibuchi$^{4}$ \quad
  Chu Kiong Loo$^{5}$ \\
  \vspace{0.55em}
  \normalfont\small
  $^{1}$Graduate School of Informatics, Osaka Metropolitan University\par
  $^{2}$Knowledge Technology Research Group, Department of Informatics, University of Hamburg\par
  $^{3}$Graduate School of Environmental, Life, Natural Science and Technology, Okayama University\par
  $^{4}$Department of Computer Science and Engineering, Southern University of Science and Technology\par
  $^{5}$Department of Artificial Intelligence, Faculty of Computer Science and Information Technology, Universiti Malaya\par
  \vspace{0.45em}
  \texttt{masuyama@omu.ac.jp}, \texttt{nojima@omu.ac.jp}\par
  \texttt{stefan.wermter@uni-hamburg.de}, \texttt{ytoda@okayama-u.ac.jp}\par
  \texttt{hisao@sustech.edu.cn}, \texttt{ckloo.um@um.edu.my}\par
  \end{minipage}
}

\author{\phidaauthorblock}

\begin{document}

\maketitle
\footnotetext{Corresponding Author: Naoki Masuyama, \texttt{masuyama@omu.ac.jp}.}

\begin{abstract}
Online clustering methods that adaptively create and update nodes as data arrive often make node learning explicit, whereas the mapping from the learned node state to output clusters often remains implicit or simplified. Implicit mappings make output clusters sensitive to weak graph bridges or local relations based on distance in the graph over learned nodes, leaving no explicit constraint on which node groups remain intact during mapping. This paper addresses this gap by proposing PHIDA, a persistence-guided node-to-cluster mapping method for online clustering with learned nodes. PHIDA implements this mapping within Adaptive Resonance Theory (ART)-based online clustering by combining Inverse-Distance ART (IDA) node learning with node-to-cluster mapping constrained by Persistent Homology (PH). Experiments on 24 benchmark datasets show that PHIDA achieves the best average ranks in stationary comparisons that include the recent stationary-only clustering methods, while also improving aggregate performance in the nonstationary setting over the evaluated online methods that adaptively create and update nodes. Ablations and comparisons with conventional node-to-cluster mappings indicate that the observed gains are associated with PH-constrained mapping that preserves raw PH components, together with the use of the PH component view during node learning. Source code is available at \url{https://github.com/Masuyama-lab/PHIDA}
\end{abstract}

\section{Introduction}
\label{sec:introduction}

For a long time, online clustering has used learned nodes, prototypes, and graph states to summarize streams \citep{fritzke95,furao07,wiwatcharakoses20}. Adaptive Resonance Theory (ART)-based clustering provides another line of online prototype learning and adaptation \citep{carpenter91b,masuyama22b,masuyama26}. These states of learned nodes are updated as new samples arrive, but node construction is not the same as deriving output clusters. Clustering performance depends on both the state of learned nodes and the mapping from learned nodes to output clusters.

This paper focuses on the mapping from learned nodes to output clusters within ART-based online clustering. The mapping rule takes learned nodes, support counts, and geometric relations as inputs and derives output clusters through explicit node-to-cluster mapping. This viewpoint distinguishes node construction from output cluster selection while keeping both roles inside a single online learning procedure.

The proposed method, PHIDA, combines Inverse-Distance ART (IDA) node learning with node-to-cluster mapping constrained by Persistent Homology (PH)~\citep{zomorodian04} within a single online learning process. IDA creates and updates learned nodes from the stream. PHIDA periodically builds a graph from the current state of learned nodes and applies density-guided $H_0$ persistence to identify raw PH components \citep{zomorodian04,chazal13,harada17,huber25}. The PH graph identifies isolated learned nodes that are eligible for removal, and the PH component view contributes to updates of the vigilance parameter in the IDA component. These operations modify the state of learned nodes as samples are processed, distinguishing PHIDA from a mapper applied after IDA node learning is complete. Across all benchmark experiments in this paper, the same PHIDA implementation configuration is used for all datasets, and PHIDA settings are not selected through a hyperparameter search specific to each dataset. The PH graph is built from the evolving state of learned nodes, raw PH components are extracted from that graph, and the node-to-cluster mapping is determined from the resulting PH component view.

The main contributions of this paper are summarized as follows.
\begin{enumerate}
\item[(i)] This paper proposes node-to-cluster mapping as an explicit role in online clustering methods that construct learned nodes or prototypes. This role is different from node construction because the learned node state does not by itself determine which node groups should remain intact as output clusters. PHIDA keeps node construction, PH component updates, and mapping within one online learning process. The PH component view is fed back to node learning rather than being used only after node learning is completed.

\item[(ii)] PHIDA implements the node-to-cluster mapping in an online clustering procedure based on ART. PHIDA applies density-guided $H_0$ persistence to the graph of learned nodes to identify raw PH components. During learning, the PH graph and the PH component view are used to prune isolated nodes and update a vigilance parameter in IDA.

\item[(iii)] Experiments on 24 benchmark datasets show that PHIDA is competitive in the stationary setting and achieves better aggregate clustering performance than the evaluated online methods that adaptively create and update nodes in the nonstationary setting. Ablations and mapping comparisons based on the IDA component further indicate that these gains are tied to the extraction and preservation of PH components.
\end{enumerate}

The remainder of this paper is organized as follows. Section~\ref{sec:related_work} reviews related work, Section~\ref{sec:method} presents the proposed PHIDA algorithm, Section~\ref{sec:results} reports empirical results and limitations, and Section~\ref{sec:conclusion} concludes with future directions.

\section{Related Work}
\label{sec:related_work}

Several online clustering methods maintain learned nodes, prototypes, or graph states while processing data sequentially, including Growing Neural Gas~\citep{fritzke95}, SOINN-type methods~\citep{furao07,wiwatcharakoses20}, and ART-based clustering~\citep{carpenter91b,masuyama22b,masuyama26}. The broader literature on online self-organizing models also contains related mechanisms for growing or adapting prototype structures~\citep{marsland02,hung23}. This paper focuses on node-to-cluster mapping for methods that explicitly maintain learned nodes, prototypes, or graph states.

ART-based clustering has been studied for incremental category learning and prototype adaptation \citep{meng13,majeed18}. Most methods in this family focus on node creation, update, merge, or removal during online learning. These methods update learned nodes, prototypes, or graph states, and output cluster labels are often induced by learned graph connectivity or node relations based on distance. The drawback is that weak graph bridges or local distance relations can determine cluster labels without an explicit constraint on which groups of learned nodes remain intact. In this sense, the mapping role is implicit even when the method specifies an operational rule for extracting clusters. IDAT is a recent ART-based reference that forms clusters as connected components of a learned graph with promoted edges \citep{masuyama25}. PHIDA differs by making the mapping role explicit within the same ART-based online learning procedure. The PH graph and PH component view also affect learning through the removal of eligible isolated learned nodes and updates of the vigilance parameter in the IDA component.

PH provides tools for extracting stable topological structure across scales \citep{zomorodian04,chazal13}. Clustering methods that use structural information and methods based on PH have also shown useful behavior in batch or graph settings \citep{harada17,ran23,huber25}. PHIDA uses PH differently. It does not apply topology directly to raw samples. Instead, PHIDA applies density-guided $H_0$ persistence to a graph built from learned nodes and uses the resulting raw PH components as the partition.

Most clustering benchmarks emphasize stationary settings \citep{ding24,wang24,yu24,zhou24bsurvey,zhou24survey,ren24}. These evaluations are valuable, but they do not fully capture streams where data distributions and learned node structures evolve over time. Related continual and federated clustering studies highlight the need for stable mappings under sequential updates \citep{kinoshita24}. PHIDA targets ART-based online clustering with learned nodes. The central question is the mapping from evolving states of learned nodes to output clusters. A remaining gap is the lack of an explicit online mapping rule for converting evolving learned node states into output clusters while preserving stable component structure during online updates.

\section{Method}
\label{sec:method}

\begin{figure*}[htbp]
  \centering
  \includegraphics[width=1.0\textwidth]{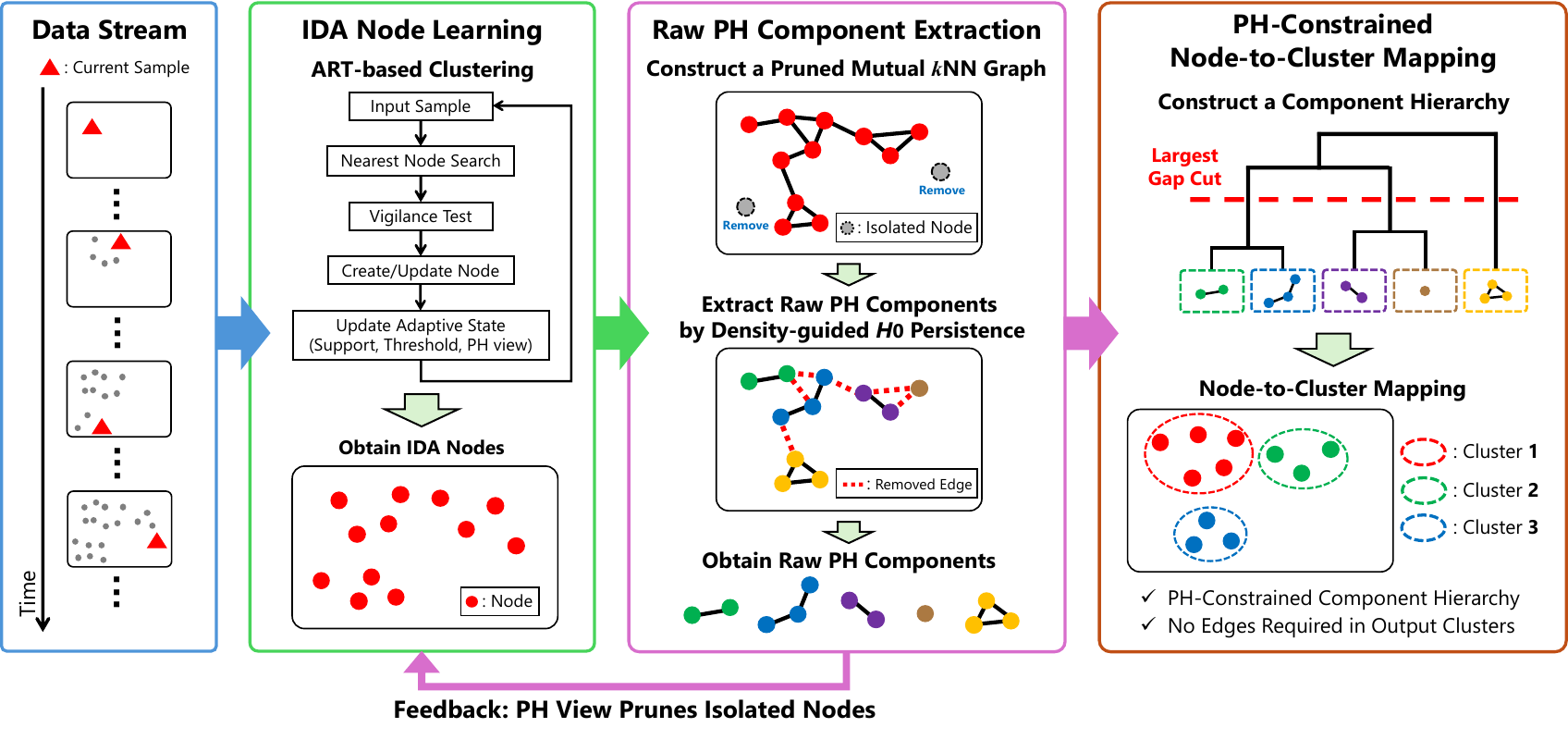}
  \caption{Overview of PHIDA algorithm. IDA summarizes streaming samples as learned nodes. PHIDA extracts raw PH components from a graph over learned nodes using density-guided $H_0$ persistence, maps them as units to output clusters through a PH-constrained component hierarchy, and feeds the PH component view back to IDA during learning.}
  \label{fig:phida_pipeline}
\end{figure*}

PHIDA combines IDA node learning with PH-constrained node-to-cluster mapping. The IDA component maintains learned nodes and support counts online. The PH-constrained stage maps learned nodes to output clusters by assigning each raw PH component as a unit and cutting a PH-constrained component hierarchy whose merge cost uses raw PH component support totals. PHIDA uses the same implementation configuration across datasets, and its PH graph and mapping decisions are computed from the stream and the current state of learned nodes. The PH graph is built as an automatically pruned mutual $k$ Nearest Neighbor (kNN) graph over learned nodes. Figure~\ref{fig:phida_pipeline} summarizes the pipeline.

The subsections below define the IDA component, PH construction, node-to-cluster mapping, out-of-sample assignment, and computational complexity.

\subsection{IDA Node Learning Component within PHIDA}

The IDA component creates and updates learned nodes from the stream using inverse-distance matching and adaptive vigilance. It maintains representatives of learned nodes $Y=\{\mathbf{y}_i\}_{i=1}^{K}$, support counts $\{M_i\}_{i=1}^{K}$, flags for nodes active for prediction, and states for the adaptive feature transformation and the vigilance parameter. For each sample, IDA applies the current adaptive feature transformation, computes distances to representatives of learned nodes in the transformed space, and converts those distances into inverse-distance similarities. It then selects the winner by the nearest distance and applies the vigilance test. If no node passes the test, IDA creates a new node. Otherwise, it updates the selected node and also updates the runner-up when it passes the vigilance test and is compatible with the current PH component view. The state $(Y,\{M_i\})$ defines the graph and component view used by PHIDA.

The PH component view affects IDA updates through the vigilance parameter and optional secondary updates. It also constrains node-to-cluster mapping by requiring nodes in the same raw PH component to receive the same output cluster label. Appendix~\ref{app:ida_details} gives the detailed update equations and implementation settings.

\subsection{Problem Setup: Node Construction and Node-to-Cluster Mapping}

PHIDA treats the node-to-cluster mapping as the step that maps the state of learned nodes to output clusters. Table~\ref{tab:phida_notation} of Appendix~\ref{app:ida_details} summarizes the notation. The IDA component outputs representatives of learned nodes $Y=\{\mathbf{y}_i\}_{i=1}^{K}$, support counts $M_i \in \mathbb{N}_{>0}$, indicators for nodes active for prediction, and states for the feature transformation and threshold. The mapping component takes this state and returns output clusters over the learned nodes.

\textbf{Definition 1 (Node-to-cluster mapping).}
Let $\mathcal{K}=\{1,\dots,K\}$ be indices of learned nodes.
A node-to-cluster mapping is a function
$g: \mathcal{K} \rightarrow \{1,\dots,C\}$, where $C$ denotes the number of output clusters.
The induced output clustering is
$\Pi(g)=\{\mathcal{C}_c(g)\}_{c=1}^{C}$, where
$\mathcal{C}_c(g)=\{i \in \mathcal{K} \mid g(i)=c\}$.

PHIDA constructs $g$ under a constraint that preserves raw PH components. All learned nodes in the same raw PH component receive the same output cluster label. Equivalently, each raw PH component is assigned as a whole to a single output cluster, and each output cluster contains one or more raw PH components. The following subsections define the PH graph, the raw PH components, and the hierarchy cut used for this assignment.

\subsection{PH-Constrained Raw Components}

PHIDA selects PH input nodes from the current IDA state $(Y,\{M_i\})$. Nodes active for prediction are used when available, and all current nodes are used otherwise. The maintenance step removes eligible nodes with support $M_i=1$ and prunes isolated nodes in the PH graph, while retaining at least two nodes with the highest density when needed.

The PH graph is rebuilt from representatives of learned nodes rather than using the current IDA edge set. Let $\mathcal{A}\subseteq \mathcal{K}$ be the node indices used by the PH builder and let $K_{\mathcal{A}}=|\mathcal{A}|$. The adaptive feature transformation is
\begin{equation}
  \Phi_{\mathcal{A}}(\mathbf{x})_j
  =
  \frac{x_j-\operatorname{med}_j(\mathcal{A})}
       {\widehat{\sigma}_j(\mathcal{A})^{\gamma_{\mathcal{A}}}},
  \qquad
  \widehat{\sigma}_j(\mathcal{A})=\frac{\operatorname{IQR}_j(\mathcal{A})}{1.349},
  \label{eq:phida_adaptive_transform}
\end{equation}
with
\begin{equation}
  \gamma_{\mathcal{A}}
  =
  \max\left\{1-\frac{1}{\operatorname{cv}(\widehat{\boldsymbol{\sigma}}(\mathcal{A}))^2},\,0\right\},
  \label{eq:phida_scale_exponent}
\end{equation}
where $\operatorname{med}_j(\mathcal{A})$ and $\operatorname{IQR}_j(\mathcal{A})$ denote the Hazen median and interquartile range of the $j$-th coordinates of $\{\mathbf{y}_i:i\in\mathcal{A}\}$, and $\widehat{\boldsymbol{\sigma}}(\mathcal{A})=(\widehat{\sigma}_1(\mathcal{A}),\ldots,\widehat{\sigma}_d(\mathcal{A}))$. Here, $\operatorname{cv}(\mathbf{v})=\operatorname{std}(\mathbf{v})/\operatorname{mean}(\mathbf{v})$ for positive finite entries, and $\gamma_{\mathcal{A}}=0$ when the scale coefficient of variation is zero. The transformation preserves Euclidean distances up to centering when feature scales are nearly homogeneous and progressively normalizes feature scales as heterogeneity in feature scale increases.

PHIDA then constructs an automatically pruned mutual $k$NN graph
\begin{equation}
  \mathcal{G}_{\mathrm{kNN}}=(\mathcal{A},\mathcal{E}_{\mathrm{kNN}}).
  \label{eq:phida_knn_graph}
\end{equation}
For $K_{\mathcal{A}}>1$, the neighborhood size is
\begin{equation}
  k_{\mathcal{A}}
  =
  \max\left\{
    1,\,
    \min\left(
      \left\lceil\sqrt{\log K_{\mathcal{A}}}\right\rceil,
      K_{\mathcal{A}}-1
    \right)
  \right\}.
  \label{eq:phida_knn_size}
\end{equation}
When compatible node-wise feature weights are available, PHIDA uses them only to add a nonnegative local metric penalty to the transformed Euclidean candidate distances. PHIDA then prunes each directed candidate row by the smaller of a row-wise Hazen median plus $1.5$IQR threshold and a global threshold computed over all directed candidates among the $k_{\mathcal{A}}$ nearest neighbors. Let $N_i$ denote the retained directed neighbor set of node $i$ after this pruning, with the nearest candidate retained when pruning would otherwise leave the row empty. PHIDA forms $\mathcal{E}_{\mathrm{kNN}}$ only from pairs of nodes that mutually retain each other as directed neighbors. Appendix~\ref{app:phida_knn_pruning} gives the candidate distance, candidate set, and pruning definitions.

On this graph, PHIDA runs density-guided $H_0$ persistence using the density based on node support
\begin{equation}
  \rho_i=\log M_i.
  \label{eq:phida_node_support_density}
\end{equation}
Nodes are activated in descending $\rho_i$. When an activated edge joins two active components, the mode with lower density dies into the mode with higher density, and its persistence is the birth density minus the current activation level. Surviving modes have infinite persistence. The present implementation does not use features in higher dimensions \citep{zomorodian04,chazal13}.

Let $0<a_1<\dots<a_R$ be the positive finite persistence levels, with $a_0=0$. If $R\le 1$, PHIDA sets $\varepsilon_{\mathrm{PH}}=0$. Otherwise, PHIDA chooses the largest gap index and persistence threshold as
\begin{equation}
  r^*
  =
  \arg\max_{1\le r\le R}(a_r-a_{r-1}),
  \qquad
  \varepsilon_{\mathrm{PH}}=a_{r^*-1},
  \label{eq:phida_ph_largest_gap_threshold}
\end{equation}
where ties in the maximization over $r$ are broken toward the larger upper level $a_r$. PHIDA obtains the raw PH component label for each node by following parent links until the current mode persistence exceeds $\varepsilon_{\mathrm{PH}}$. This produces the raw PH component partition
\begin{equation}
  \mathcal{B}=\{B_a\}_{a=1}^{K_{\mathrm{PH}}}.
  \label{eq:phida_raw_ph_component_partition}
\end{equation}
where each $B_a\subseteq\mathcal{A}$ is a raw PH component and $K_{\mathrm{PH}}$ is the resulting number of raw PH components. PH construction uses only the current state of learned nodes. PHIDA does not use true labels during learning or node-to-cluster mapping. True labels are used only to compute evaluation metrics. Section~\ref{sec:results} and Appendix~\ref{app:benchmark_protocol} describe baseline tuning.

The raw PH components define the partition used by the subsequent node-to-cluster mapping. Each raw PH component is assigned as a whole to a single output cluster. An output cluster may contain one or more raw PH components.

\subsection{PH-Constrained Node-to-Cluster Mapping in PHIDA}

PHIDA obtains output clusters through a PH-constrained component hierarchy whose merge cost uses raw PH component support totals. Let $S_a=\sum_{i\in B_a}M_i$ be the support of raw PH component $B_a$, and let $\pi_a$ be its support-weighted persistence summary. PHIDA forms the PH-stable component mass
\begin{equation}
  S_a^*
  =
  S_a\,\phi(\pi_a),
  \qquad
  \phi(\pi)=\frac{\pi}{\pi+\pi_{\mathrm{ref}}},
  \label{eq:phida_ph_stable_mass}
\end{equation}
where $\pi_{\mathrm{ref}}$ is the median positive component persistence. If the PH-stable masses are invalid or give a nonpositive merge weight, PHIDA falls back to the component support totals $S_a$. The normalized masses define the entropy effective component count
\begin{equation}
  C_{\mathrm{ent}}
  =
  \exp\left(-\sum_{a=1}^{K_{\mathrm{PH}}} p_a\log p_a\right),
  \qquad
  p_a=\frac{S_a^*}{\sum_b S_b^*}.
  \label{eq:phida_entropy_effective_count}
\end{equation}
The minimum retained cluster count is
\begin{equation}
  C_{\min}
  =
  \min\{\lceil C_{\mathrm{ent}}\rceil,K_{\mathrm{PH}}\},
  \qquad
  C_{\min}\ge 1.
  \label{eq:phida_min_retained_cluster_count}
\end{equation}

PHIDA computes a support-weighted centroid for each raw PH component in the transformed feature space and builds another automatically pruned mutual $k$NN graph over these centroids. PHIDA performs deterministic agglomeration only along the edges of this component graph. For adjacent groups $A$ and $B$, let $W_A$ and $W_B$ denote their total merge weights, and let $\boldsymbol{\mu}_A$ and $\boldsymbol{\mu}_B$ denote their weighted centroids in the transformed feature space. PHIDA computes the merge height as
\begin{equation}
  q(A,B)
  =
  \frac{W_AW_B}{W_A+W_B}
  \left\|\boldsymbol{\mu}_A-\boldsymbol{\mu}_B\right\|_2^2.
  \label{eq:phida_ward_merge_height}
\end{equation}
This merge cost is the increase in weighted dispersion within groups for the two adjacent groups, evaluated with the weights and centroids used by PHIDA.
At each step, PHIDA merges the eligible adjacent pair with the smallest $q(A,B)$, updates the component graph, and records the partition and merge height. Agglomeration stops at $C_{\min}$ groups or when no eligible adjacent pair remains.

Let $\mathcal{H}_{\mathrm{W}}=\{\mathcal{P}^{(0)},\dots,\mathcal{P}^{(L)}\}$ be the recorded component hierarchy over raw PH components, with merge heights $q_\ell$. PHIDA considers the eligible levels whose number of groups is at least $C_{\min}$. The hierarchy level immediately before the largest adjacent jump between merge heights is
\begin{equation}
  \Delta_\ell = q_{\ell+1}-q_\ell,
  \qquad
  \ell^*=\arg\max_{\ell:\,|\mathcal{P}^{(\ell)}|\ge C_{\min}}
  \Delta_\ell.
  \label{eq:phida_merge_height_gap_level}
\end{equation}
Ties are broken toward the larger upper merge height. Expanding $\mathcal{P}^{(\ell^*)}$ back to learned nodes yields the node-to-cluster mapping $g^{(\ell^*)}$ induced by the cut. The hierarchy cut is determined from merge heights during learning.

This hierarchy defines the feasible set of PH-constrained node-to-cluster mappings
\begin{equation}
  \mathcal{M}_{\mathrm{PH}}(\mathcal{H}_{\mathrm{W}})=
  \{g \mid \Pi(g)\text{ is induced by a cut of }\mathcal{H}_{\mathrm{W}}\}.
  \label{eq:phida_ph_mapping_set}
\end{equation}
Within PHIDA, node-to-cluster mapping is restricted to $\mathcal{M}_{\mathrm{PH}}(\mathcal{H}_{\mathrm{W}})$. The restriction is tied to the online state because IDA also uses the PH component view for compatibility checks and vigilance parameter updates. Thus, PHIDA is not an independent node learning procedure followed by a detachable mapper. The PHIDA-noPH ablation removes the raw PH component partition from learning and mapping while keeping the remaining matching, maintenance, and mapping procedure comparable.

\paragraph{Relation to conventional mappings.}
The PH-constrained stage operates on the state of learned nodes, not raw samples. Connected components on the graph over learned nodes can join density modes through weak graph bridges. Ward hierarchical clustering~\citep{ward63} on learned nodes does not enforce raw PH component preservation. HDBSCAN~\citep{campello15} or ToMATo~\citep{chazal13} applied at the sample level addresses a different density or persistence problem. Appendix~\ref{app:ida_component_mapping_control_cd} separately evaluates HDBSCAN applied to learned nodes produced by the IDA component. PHIDA also differs during learning because it uses the PH component view for compatibility checks and vigilance parameter updates.

\subsection{Out-of-Sample Cluster Assignment}
\label{sec:prediction}

For out-of-sample cluster assignment, PHIDA assigns a query sample to an output cluster rather than to a raw learned node. The rule combines the distance to the nearest learned node after applying the learned feature transformation with a cluster radius term based on support. For an output cluster with a single node, the rule reduces to the squared distance in the transformed space to that node. Appendix~\ref{app:out_of_sample_assignment} gives the full rule. The assignment uses only representatives of learned nodes, support counts, the selected mapping $g^{(\ell^\ast)}$, and the learned state of the feature transformation. Ground-truth labels are not used.

\subsection{Computational Complexity}
\label{sec:complexity}

Let $T$ be the number of streamed samples, $d$ the dimension, $K$ the maximum number of learned nodes, $\lambda$ the average interval for recalculating the vigilance parameter, and $\tau_{\mathrm{PH}}$ the effective average interval between PH component view refreshes, including the refresh after learning ends. The online matching term costs $\mathcal{O}(Kd)$ per sample. The update of the vigilance parameter contributes $\mathcal{O}(\lambda d+\lambda^2)$ amortized per sample. A PH refresh costs $\mathcal{O}(K^2d+K^2\log K)$ per refresh, yielding the amortized term in Eq.~\eqref{eq:phida_total_complexity} when refreshes occur every $\tau_{\mathrm{PH}}$ samples on average.

The combined big-$\mathcal{O}$ expression for $T$ streamed samples is
\begin{equation}
  \mathcal{O}\left(
  T\cdot
  \max\left\{
  Kd,\ \lambda d+\lambda^2,\ 
  \frac{K^2d+K^2\log K}{\tau_{\mathrm{PH}}}
  \right\}
  \right).
  \label{eq:phida_total_complexity}
\end{equation}

This expression characterizes implementation cost and accounts for the additional cost of periodic PH refreshes. Empirically, PHIDA can still produce fewer clusters at the final evaluation point on some datasets.

\section{Results}
\label{sec:results}

\subsection{Compared Algorithms}

The experiments compare PHIDA with baselines from several clustering paradigms. The nonstationary comparison includes GNG~\citep{fritzke95}, SOINN+~\citep{wiwatcharakoses20}, DDVFA~\citep{da20}, CAEA~\citep{masuyama22a}, CAE~\citep{masuyama26}, IDAT~\citep{masuyama25}, and PHIDA. The stationary comparison additionally includes TBM-P~\citep{yelugam23}, TC~\citep{yang25}, SR-PCM-HDP~\citep{hu25}, TableDC~\citep{rauf25}, G-CEALS~\citep{rabbani25}, and AuToMATo~\citep{huber25}. AuToMATo is evaluated only in the stationary setting and has no tunable hyperparameters in this study.

All methods follow the same dataset collection, preprocessing, seed identities, evaluation measures, and handling of failed runs. All reported means and standard deviations are computed over 30 independent runs, indexed by shared random seed identities for each dataset and setting. For methods supporting the nonstationary setting, staged streams also follow that comparison. PHIDA is evaluated with fixed implementation settings. No PHIDA hyperparameter or internal selection rule is selected by maximizing Adjusted Rand Index (ARI)~\citep{hubert85} with true labels. The quantities used by PHIDA for adaptive feature transformation, adaptive updates of the vigilance parameter, PH graph construction, persistence threshold selection, component hierarchy construction, minimum retained cluster count, cut selection from merge heights, node-to-cluster mapping, and out-of-sample assignment are computed from the input stream and state of learned nodes during learning. Thus, PHIDA is not included in the hyperparameter search loop used for tunable baselines.

Baseline hyperparameter handling depends on method type. GNG, DDVFA, and CAEA are tunable in both settings. TBM-P and SR-PCM-HDP are tunable when included in the stationary comparison. Bayesian optimization selects these baseline hyperparameters by maximizing ARI under the shared protocol. IDAT, SOINN+, CAE, TC, and AuToMATo are run without explicit external hyperparameter optimization. TC and AuToMATo are evaluated only in the stationary setting. TableDC and G-CEALS use fixed deep clustering implementations with the dataset class count as required by their method specifications, but they are not selected by Bayesian optimization. Ground-truth labels are used for PHIDA only to compute evaluation metrics. Appendix~\ref{app:benchmark_protocol} gives the detailed protocol.

\subsection{Evaluation Metrics}

ARI \citep{hubert85} and Adjusted Mutual Information (AMI) \citep{vinh10} are the main evaluation metrics. Both compare the obtained clustering with ground-truth classes after chance correction, and higher values indicate better clustering performance.

In the nonstationary setting, let $Q_j$ denote the value of a generic evaluation measure $Q$ after the $j$-th incremental stage, where $j = 1, \dots, J$ and $Q \in \{\mathrm{ARI}, \mathrm{AMI}\}$. The streamwise summary is defined by
\begin{equation}
  \operatorname{avgInc}_{Q}
  =
  \frac{1}{J}\sum_{j=1}^{J} Q_j,
  \qquad
  Q \in \{\mathrm{ARI}, \mathrm{AMI}\}.
  \label{eq:phida_avginc_metric}
\end{equation}
Accordingly, the reported streamwise summaries are $\operatorname{avgInc}_{\mathrm{ARI}}$ and $\operatorname{avgInc}_{\mathrm{AMI}}$.

Critical difference diagrams use Friedman tests followed by Nemenyi post-hoc comparisons over per-dataset ranks~\citep{demvsar06}. Focused comparisons use two-sided Wilcoxon signed-rank tests on paired per-dataset ARI and AMI values~\citep{wilcoxon45}. Holm correction is applied separately within each reported PHIDA-vs-IDAT comparison family: four nonstationary performance summaries and two stationary final score summaries. The focused comparisons also include PHIDA versus SOINN+ and PHIDA-noPH. Node counts and final cluster counts are descriptive structural quantities, not superiority objectives. Unavailable or all-NaN results receive the worst rank, while Wilcoxon tests use finite paired values only. The significance level is $0.05$. Appendix~\ref{app:benchmark_protocol} gives the detailed protocol and handling of failed runs.

\subsection{Nonstationary Comparison}

The primary empirical evidence comes from the nonstationary regime, where PHIDA has a favorable rank-based profile across the reported performance summaries.

\begin{figure*}[htbp]
\centering
\begin{subfigure}[t]{0.48\textwidth}
  \centering
  \includegraphics[width=\linewidth]{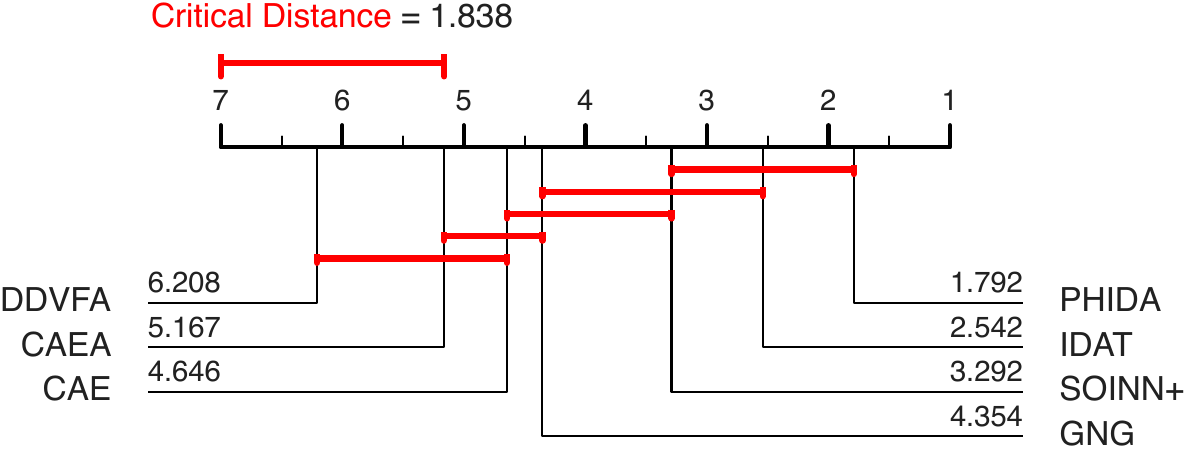}
  \caption{Final ARI}
\end{subfigure}\hfill
\begin{subfigure}[t]{0.48\textwidth}
  \centering
  \includegraphics[width=\linewidth]{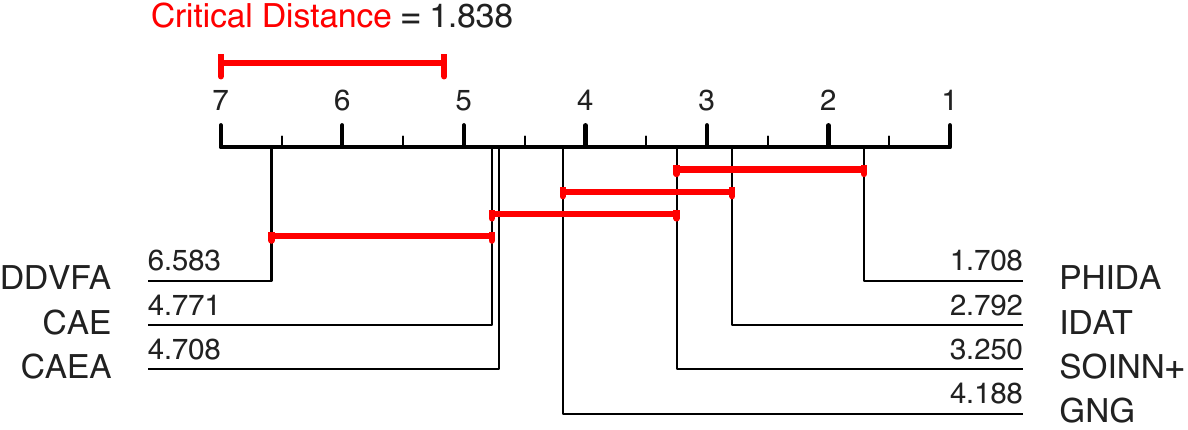}
  \caption{Final AMI}
\end{subfigure}

\vspace{0.6em}

\begin{subfigure}[t]{0.48\textwidth}
  \centering
  \includegraphics[width=\linewidth]{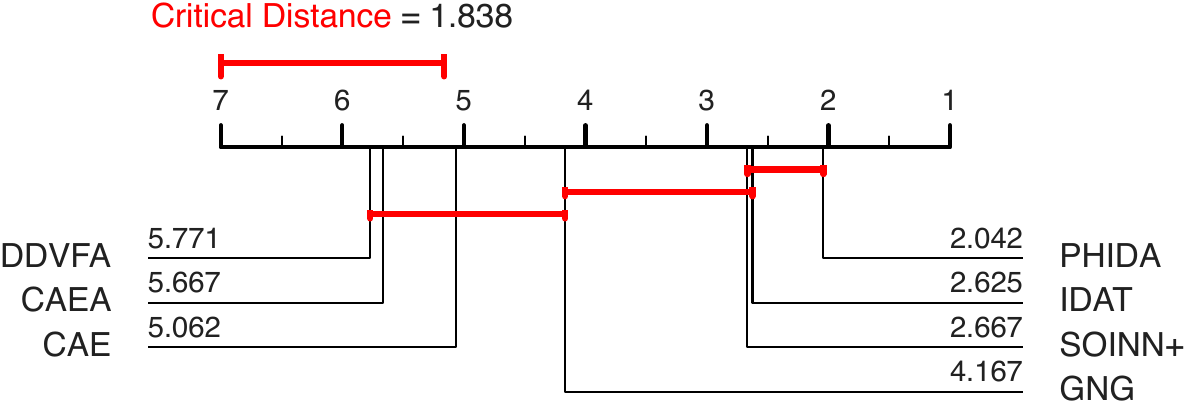}
  \caption{avgInc\_ARI}
\end{subfigure}\hfill
\begin{subfigure}[t]{0.48\textwidth}
  \centering
  \includegraphics[width=\linewidth]{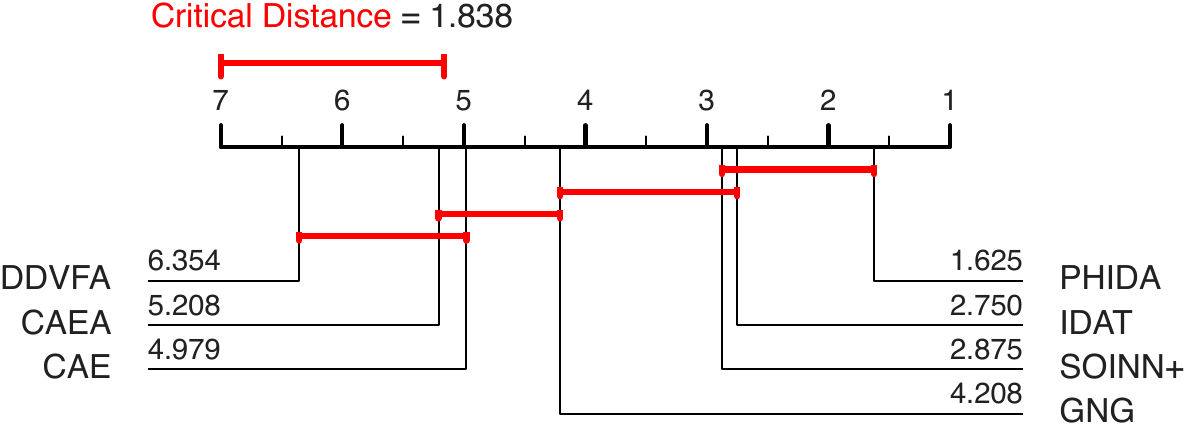}
  \caption{avgInc\_AMI}
\end{subfigure}
\caption{Critical difference diagrams based on final ARI, final AMI, avgInc\_ARI, and avgInc\_AMI in the nonstationary setting. Smaller average rank is better.}
\label{fig:cd_nonstationary_main}
\end{figure*}

Figure~\ref{fig:cd_nonstationary_main} provides an overview of the main nonstationary results across final ARI, final AMI, avgInc\_ARI, and avgInc\_AMI. Tables~\ref{tab:app_nonstationary_final_quality} and~\ref{tab:app_nonstationary_streamwise} of Appendix~\ref{app:benchmark_tables} provide the exact per-dataset values.

Focused two-sided Wilcoxon signed-rank comparisons against IDAT are significant for final ARI ($p=0.013$), final AMI ($p=0.001$), avgInc\_ARI ($p=0.008$), and avgInc\_AMI ($p<0.001$). These four PHIDA-vs-IDAT comparisons remain significant after Holm correction across the four nonstationary performance summaries, with adjusted $p$-values $0.016$, $0.003$, $0.016$, and $0.001$, respectively. Because IDAT is a paired ART-based reference, this comparison is especially informative. PHIDA also significantly improves over SOINN+ in final ARI ($p=0.003$) and final AMI ($p=0.003$).

Across the 24 nonstationary datasets, PHIDA obtains the best final ARI and final AMI on 16 datasets each. Together with Figure~\ref{fig:cd_nonstationary_main} and the paired tests, these counts indicate an aggregate advantage in the nonstationary setting. Dataset exceptions are reported in Tables~\ref{tab:app_nonstationary_final_quality} and~\ref{tab:app_nonstationary_streamwise} of Appendix~\ref{app:benchmark_tables}, with discussion by dataset in Appendix~\ref{app:structure_quality_tradeoffs}. This does not imply per-dataset dominance.

\subsection{Stationary Comparison}

\begin{figure}[htbp]
\centering
\begin{subfigure}[t]{0.48\textwidth}
  \centering
  \includegraphics[width=\linewidth]{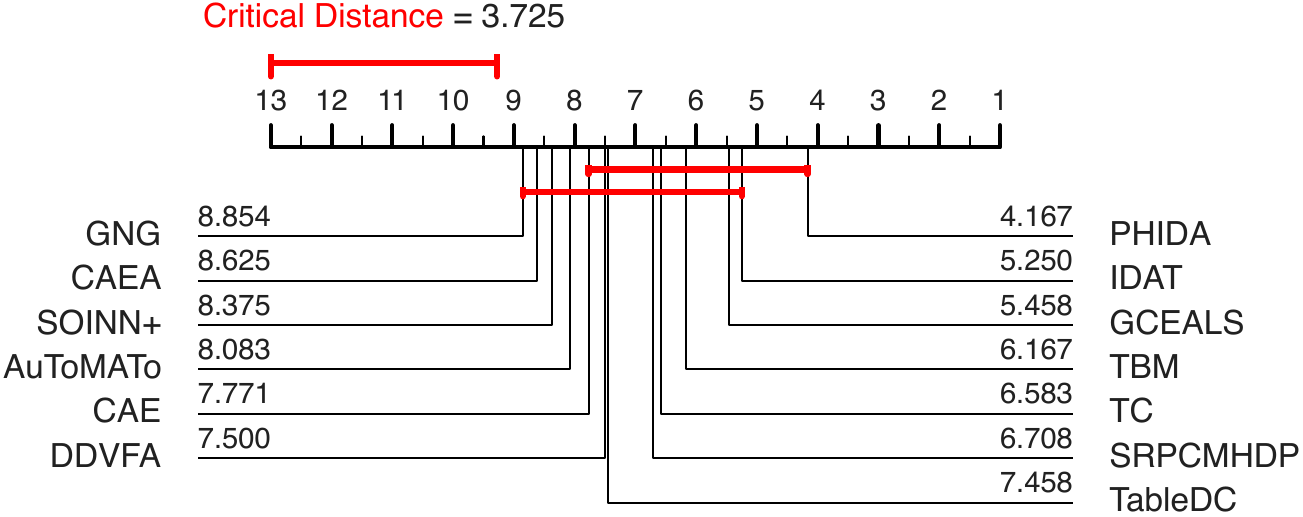}
  \caption{Final ARI}
\end{subfigure}\hfill
\begin{subfigure}[t]{0.48\textwidth}
  \centering
  \includegraphics[width=\linewidth]{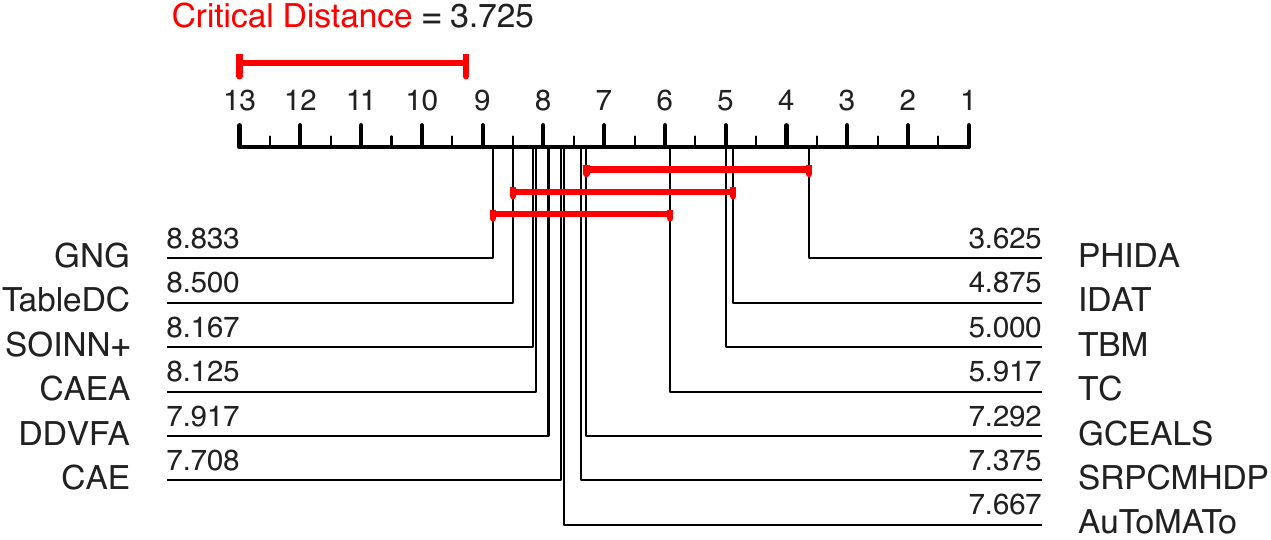}
  \caption{Final AMI}
\end{subfigure}
\caption{Critical difference diagrams based on final ARI and final AMI in the stationary setting. Smaller average rank is better.}
\label{fig:cd_stationary_main}
\end{figure}

Stationary results evaluate conventional clustering, while nonstationary results evaluate online functionality. Figure~\ref{fig:cd_stationary_main} provides an overview of final ARI and AMI in the stationary setting.

PHIDA has the best average rank under both stationary evaluation metrics. In the stationary setting, paired two-sided Wilcoxon signed-rank tests against IDAT over the 24 per-dataset final scores indicate significant gains in final ARI (raw $p=0.023$, Holm-adjusted $p=0.023$) and final AMI (raw $p=0.003$, Holm-adjusted $p=0.005$). Holm correction is applied only across these two stationary PHIDA-vs-IDAT comparisons, and both remain significant at $\alpha=0.05$. The exact values are reported in Table~\ref{tab:app_stationary_quality} of Appendix~\ref{app:benchmark_tables}. Some entries favor other methods. Therefore, the claim is aggregate competitiveness rather than uniform dominance. Appendix~\ref{app:structure_quality_tradeoffs} discusses counts of learned nodes, output granularity, and ARI and AMI values.

\subsection{Ablation and Mapping Comparisons}

PHIDA-noPH removes the raw PH component view from PHIDA learning, including compatibility behavior during learning and node-to-cluster mapping. This differs from changing only the final node-to-cluster mapping after learning is complete.

The secondary ablations remove one mechanism at a time relative to PHIDA. noRefresh disables the periodic refresh of the PH component view during learning while retaining the refresh after learning ends needed to build the output assignment view. noDelete disables physical deletion of learned nodes during maintenance. noPrune disables filtering of isolated nodes from the PH input while keeping the mutual $k$NN candidate distance pruning rule unchanged. Appendix~\ref{app:ablation_variant_protocol} maps these switches to Algorithms~\ref{alg:ida_online_update}--\ref{alg:phida_maintenance_refresh}.

The mapping comparisons in Appendix~\ref{app:ida_component_mapping_control_cd} test a separate question. They evaluate whether conventional mappings applied after node construction by IDA match PHIDA when those mappings receive information that PHIDA does not use during learning. IDA+Ward, IDA+HDBSCAN, and IDA+KM use learned nodes produced by the IDA component without the PH component view. Ward and K-means use the number of classes in the ground-truth labels as the target number of clusters. In the nonstationary setting, the target is the number of classes observed up to the current evaluation stage. HDBSCAN uses the corresponding target to choose a hierarchy cut. These methods are therefore additional comparisons rather than baselines evaluated without using the true number of classes.

In the four nonstationary CD diagrams for the mapping comparisons, PHIDA has the best average rank for final ARI, final AMI, and avgInc\_AMI. IDA+KM has the best average rank for avgInc\_ARI, followed by IDA+Ward and PHIDA. IDA+HDBSCAN has the weakest average rank profile across these diagrams.

These results indicate that conventional mappings applied to the state of learned nodes produced by IDA, even with the target number derived from the ground-truth labels, do not uniformly match PHIDA in the nonstationary setting. They complement the noPH ablation by testing the absence of the PH component view during learning and the final mapping rule.

\begin{figure}[htbp]
  \centering
  \begin{subfigure}[t]{0.48\textwidth}
    \centering
    \includegraphics[width=\linewidth]{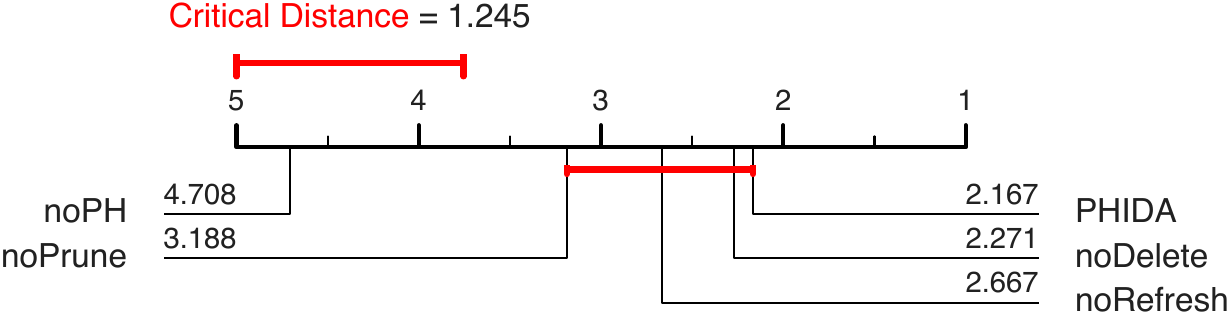}
    \caption{Final ARI}
  \end{subfigure}\hfill
  \begin{subfigure}[t]{0.48\textwidth}
    \centering
    \includegraphics[width=\linewidth]{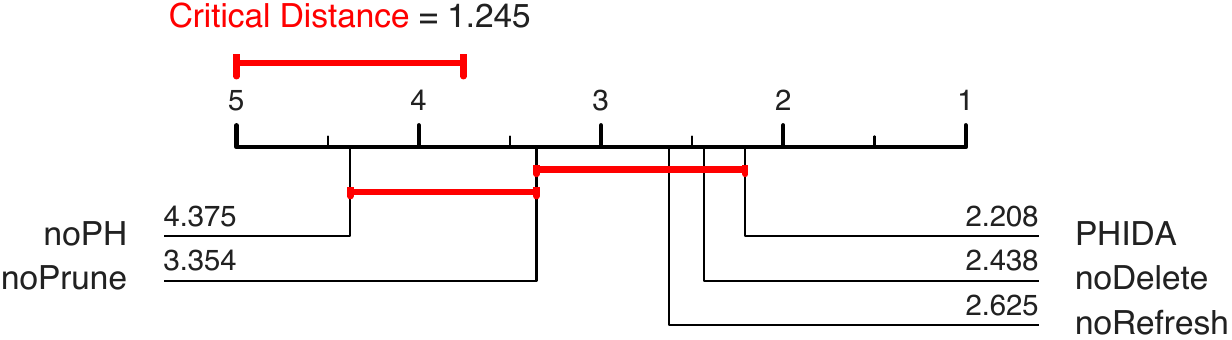}
    \caption{Final AMI}
  \end{subfigure}
  \caption{Nonstationary ablation CD diagrams. Lower rank is better. Ablation switches are defined in Appendix~\ref{app:ablation_variant_protocol}.}
  \label{fig:ablation_cd_nonstationary}
\end{figure}

Figure~\ref{fig:ablation_cd_nonstationary} summarizes final ARI and AMI ablation ranks, where noPH has the worst rank profile and PHIDA has the best.

PHIDA beats PHIDA-noPH on 23 of 24 datasets in ARI and 21 of 24 in AMI. PHIDA-noPH produces many small output clusters and lower ARI and AMI values. The noDelete variant remains close to PHIDA in both diagrams, suggesting that physical deletion of learned nodes is not the sole source of the gains when the PH component view and PH-constrained mapping are retained. Other ablations are secondary analyses, with stationary diagrams in Appendix~\ref{app:ablation_cd_stationary}.

\subsection{Limitations}
\label{sec:limitations}

The empirical claims are aggregate claims over the benchmark suites. The strongest empirical advantage appears in the nonstationary setting, while the stationary setting still shows aggregate competitiveness and significant gains over IDAT in final ARI and final AMI. Because PHIDA explicitly maps learned nodes to output clusters, structural summaries and the complexity characterization are reported together with ARI and AMI values. The experiments evaluate the PHIDA procedure implemented in this paper, including the PH component view used during learning and the preservation of raw PH components during mapping. They also include targeted mapping comparisons based on the IDA component, but these comparisons are not exhaustive over all possible mappings. However, the benchmark focuses on methods that explicitly maintain learned nodes, prototypes, or graph states, matching the node-to-cluster mapping scope of PHIDA. Appendix~\ref{app:phida_properties} gives finite-sample algorithmic properties rather than statistical consistency or optimality theory.

\section{Concluding Remarks}
\label{sec:conclusion}

This paper proposes node-to-cluster mapping as an explicit role in ART-based online clustering. PHIDA implements this role together with node construction inside a single online learning procedure. The PH-constrained mapping in PHIDA assigns raw PH components as units during node-to-cluster mapping and uses a PH-constrained component hierarchy whose merge cost uses raw PH component support totals. Across 24 benchmark datasets, the strongest empirical advantage appears in the nonstationary setting. In that setting, PHIDA has favorable rank profiles, paired tests with Holm correction indicate higher ARI and AMI values than IDAT, and noPH ablations show the importance of the PH component view. Mapping comparisons based on the IDA component further show that conventional mappings given the true number of classes do not uniformly match PHIDA. These results indicate that preserving raw PH components is a useful constraint for node-to-cluster mapping in online clustering with learned nodes.

A future research direction is to connect the finite-sample algorithmic properties in Appendix~\ref{app:phida_properties} with stability guarantees for the PH component view under streaming updates. Another future direction is to extend PHIDA to use persistence summaries at several scales during learning.

\section*{Acknowledgments}
This research was supported by Japan Society for the Promotion of Science (JSPS) KAKENHI Grant Number JP25K03187.

\bibliographystyle{unsrtnat}
\bibliography{myref}

\clearpage
\appendix
\setcounter{table}{0}
\setcounter{figure}{0}
\renewcommand{\thetable}{A\arabic{table}}
\renewcommand{\thefigure}{A\arabic{figure}}

\section{Algorithmic Details of PHIDA}
\label{app:ida_details}

This appendix provides detailed algorithmic specifications for the IDA update, PHIDA node-to-cluster mapping, maintenance and refresh during learning, and out-of-sample cluster assignment procedures used by the main method.

\begin{table}[htbp]
\centering
\normalsize
\setlength{\tabcolsep}{3pt}
\caption{Main notation used in PHIDA}
\label{tab:phida_notation}
{\renewcommand{\arraystretch}{1.08}%
\makebox[\textwidth][c]{%
\begin{tabular}{@{}ll@{}}
\toprule
Notation & Description \\
\midrule
$X=\{\mathbf{x}_t\}_{t=1}^{T}$ & input stream containing $T$ samples, with each $\mathbf{x}_t\in\mathbb{R}^d$ \\
$Y=\{\mathbf{y}_i\}_{i=1}^{K}$ & representative set of learned nodes returned by IDA \\
$\mathbf{y}_i \in \mathbb{R}^d$ & representative of learned node $i$ \\
$M_i \in \mathbb{N}_{>0}$ & support count of learned node $i$ \\
$s_{i,t}$ & stored node scale for learned node $i$ \\
$\alpha_{i,t}$ & inverse-distance scale $1/\max\{s_{i,t-1},10^{-12}\}$ \\
$\mathcal{K}=\{1,\dots,K\}$ & index set of learned nodes \\
$\mathcal{A}\subseteq\mathcal{K}$ & indices of learned nodes used by the PH builder \\
$B_t$ & ring buffer of recent samples used by the adaptive vigilance update \\
$L_t^{B}$ & logical buffer retention length, equal to $2\lambda_t$ after interval updates \\
$W_t^{(m)}$ & $m$ most recent samples extracted from $B_t$, ordered oldest to newest \\
$\rho_i=\log M_i$ & density based on node support used by the density-guided $H_0$ persistence step \\
$\mathcal{G}_{\mathrm{kNN}}$ & pruned mutual $k$NN graph over learned nodes \\
$\theta_p^{\mathrm{loc}}$ & local Hazen median plus $1.5$IQR pruning threshold for row $p$ \\
$\theta_{\mathcal{A}}^{\mathrm{glob}}$ & global Hazen median plus $1.5$IQR threshold over directed candidates \\
$\mathcal{B}=\{B_a\}_{a=1}^{K_{\mathrm{PH}}}$ & raw PH component partition \\
$\varepsilon_{\mathrm{PH}}$ & persistence threshold used to form raw PH components \\
$S_a=\sum_{i\in B_a}M_i$ & support total of raw PH component $B_a$ \\
$\pi_a$ & persistence summary assigned to raw PH component $B_a$ \\
$\mathcal{H}_{\mathrm{W}}=\{\mathcal{P}^{(\ell)}\}_{\ell=0}^{L}$ & component hierarchy over raw PH components under the PH constraint \\
$C_{\min}$ & minimum number of retained groups in the PH-constrained hierarchy \\
$q_\ell$ & merge height associated with hierarchy level $\ell$ \\
$\Delta_\ell=q_{\ell+1}-q_\ell$ & adjacent gap between merge heights \\
$\Delta_t^{(m)}$ & Cholesky determinant proxy for the buffer window $W_t^{(m)}$ \\
$\Delta_{\det}$ & threshold for the Cholesky determinant proxy, equal to $10^{-6}$ \\
$\bar r_t$ & smoothed cluster-to-node ratio used as the Hazen quantile level \\
$a_{b_t,t}$ & runner-up inverse-distance similarity \\
$\chi_{b_t,t}$ & runner-up confidence factor \\
$\ell^\ast$ & selected cut level before the largest eligible gap in merge height \\
$\mathcal{M}_{\mathrm{PH}}(\mathcal{H}_{\mathrm{W}})$ & feasible set of PH-constrained node-to-cluster mappings \\
$g:\mathcal{K}\to\{1,\dots,C\}$ & node-to-cluster mapping, where $C$ is the number of output clusters \\
$\Pi(g)=\{\mathcal{C}_c(g)\}_{c=1}^{C}$ & output clustering induced by mapping $g$ \\
$g^{(\ell^\ast)}$ & selected node-to-cluster mapping at cut level $\ell^\ast$ \\
\bottomrule
\end{tabular}%
}
}
\end{table}

\subsection{State Variables and Outputs}

At time step $t$, IDA maintains the following state variables.
\begin{itemize}
    \item representatives of learned nodes $Y_t=\{\mathbf{y}_{i,t}\}_{i=1}^{K_t}$,
  \item node support counts $M_{i,t}\in\mathbb{N}_{>0}$,
  \item indicators for nodes active for prediction,
  \item adaptive feature transformation state, buffer state, vigilance parameter state, and PH component view state.
\end{itemize}

For an input stream $\{\mathbf{x}_t\}_{t=1}^{T}$, IDA outputs the state of learned nodes $(Y_T,\{M_{i,T}\})$ and the associated states for active nodes, adaptive feature transformation, and the PH component view used by PHIDA.

\subsection{\texorpdfstring{Mutual $k$NN Graph Pruning}{Mutual kNN Graph Pruning}}
\label{app:phida_knn_pruning}

This subsection specifies the automatic mutual $k$NN builder used for the PH graph over learned nodes. PHIDA uses the same rule for the component graph in the PH-constrained hierarchy, with node-wise feature weights unavailable in that call. Let $\mathcal{A}=\{i_1,\ldots,i_n\}$, with $n=K_{\mathcal{A}}>1$, and let $k_{\mathcal{A}}$ be given by Eq.~\eqref{eq:phida_knn_size}. PHIDA applies the adaptive transformation in Eq.~\eqref{eq:phida_adaptive_transform} and computes the transformed representatives and base distances as
\begin{equation}
  \mathbf{z}_p
  =
  \Phi_{\mathcal{A}}(\mathbf{y}_{i_p}),
  \qquad
  g_{pq}
  =
  \|\mathbf{z}_p-\mathbf{z}_q\|_2,
  \label{eq:phida_graph_base_distance}
\end{equation}
where $p,q\in\{1,\ldots,n\}$.

When compatible node-wise feature weights are available, PHIDA computes the averaged feature weights, correction for effective dimensionality, and local weighted distance as
\begin{align}
  \bar{\boldsymbol{\omega}}_{pq}
  &=
  \frac{1}{2}
  \left(
    \boldsymbol{\omega}_{i_p}
    +
    \boldsymbol{\omega}_{i_q}
  \right),
  \\
  \kappa_{pq}
  &=
  \operatorname{clip}
  \left(
    \frac{1}
         {\sum_{\ell=1}^{d}\bar{\omega}_{pq,\ell}^{2}},
    1,
    d
  \right),
  \\
  \ell_{pq}
  &=
  \left(
    \kappa_{pq}
    \sum_{\ell=1}^{d}
    \bar{\omega}_{pq,\ell}
    (z_{p,\ell}-z_{q,\ell})^2
  \right)^{1/2},
  \label{eq:phida_graph_local_distance}
\end{align}
where $\boldsymbol{\omega}_{i_p}$ and $\boldsymbol{\omega}_{i_q}$ denote the compatible node-wise feature weight vectors for nodes $i_p$ and $i_q$. PHIDA sets the directed candidate distances as
\begin{equation}
  \delta_{pq}
  =
  \begin{cases}
    g_{pq},
    & \text{if compatible node-wise feature weights are unavailable},\\
    g_{pq}
    +
    \gamma_{\mathcal{A}}
    \max\{\ell_{pq}-g_{pq},0\},
    & \text{otherwise},
  \end{cases},
  \label{eq:phida_graph_candidate_distance}
\end{equation}
where PHIDA sets $\delta_{pp}=\infty$, and where the second branch only adds a nonnegative local metric penalty to the transformed Euclidean distance.

For each row $p$, PHIDA lets $C_p$ contain the $k_{\mathcal{A}}$ indices with the smallest finite values of $\delta_{pq}$. PHIDA gathers the row-wise and global distance multisets for directed candidates as
\begin{equation}
  \mathcal{D}_p
  =
  \{\delta_{pq}:q\in C_p\},
  \qquad
  \mathcal{D}_{\mathcal{A}}
  =
  \bigcup_{p=1}^{n}\mathcal{D}_p,
  \label{eq:phida_graph_candidate_sets}
\end{equation}
where $\mathcal{D}_{\mathcal{A}}$ is a multiset union, so every directed candidate distance contributes once.

For a finite multiset $S$, PHIDA uses the Hazen interquartile range with
\begin{equation}
  \operatorname{IQR}^{\mathrm{Hazen}}(S)
  =
  Q_{0.75}^{\mathrm{Hazen}}(S)
  -
  Q_{0.25}^{\mathrm{Hazen}}(S),
  \label{eq:phida_hazen_iqr}
\end{equation}
where $Q_q^{\mathrm{Hazen}}(S)$ denotes the Hazen $q$-quantile and $\operatorname{med}^{\mathrm{Hazen}}(S)=Q_{0.5}^{\mathrm{Hazen}}(S)$. PHIDA computes the row-wise and global pruning thresholds as
\begin{equation}
  \theta_p^{\mathrm{loc}}
  =
  \operatorname{med}^{\mathrm{Hazen}}(\mathcal{D}_p)
  +
  1.5\,\operatorname{IQR}^{\mathrm{Hazen}}(\mathcal{D}_p),
  \qquad
  \theta_{\mathcal{A}}^{\mathrm{glob}}
  =
  \operatorname{med}^{\mathrm{Hazen}}(\mathcal{D}_{\mathcal{A}})
  +
  1.5\,\operatorname{IQR}^{\mathrm{Hazen}}(\mathcal{D}_{\mathcal{A}}),
  \label{eq:phida_graph_pruning_thresholds}
\end{equation}
Here, PHIDA computes the global threshold over all directed candidates among the $k_{\mathcal{A}}$ nearest neighbors, not over all pairwise distances.

PHIDA retains directed candidates as
\begin{equation}
  N_p
  =
  \left\{
    q\in C_p:
    \delta_{pq}
    \le
    \min\{
      \theta_p^{\mathrm{loc}},
      \theta_{\mathcal{A}}^{\mathrm{glob}}
    \}
  \right\},
  \label{eq:phida_graph_retained_neighbors}
\end{equation}
where, if $N_p=\emptyset$, PHIDA replaces $N_p$ by the singleton containing the nearest candidate in $C_p$. PHIDA includes undirected edges by mutual retention as
\begin{equation}
  (i_p,i_q)\in\mathcal{E}_{\mathrm{kNN}}
  \quad\Longleftrightarrow\quad
  q\in N_p,\ p\in N_q,
  \label{eq:phida_mutual_edge_rule}
\end{equation}
where PHIDA stores each unordered edge once.

\subsection{Similarity Computation for Nearest Node Selection}

For each input $\mathbf{x}_t$, IDA applies the current adaptive feature transformation and computes distances between $\mathbf{x}_t$ and the current representatives of learned nodes in the transformed feature space. With node-wise scale parameter $\alpha_{i,t}$, the inverse-distance similarity for learned node $i$ is
\begin{equation}
  a_{i,t}
  =
  \frac{1}{1+\alpha_{i,t}d_{i,t}},
  \label{eq:phida_appendix_match_signal}
\end{equation}
where $d_{i,t}$ is the distance in the transformed space between $\mathbf{x}_t$ and learned node representative $\mathbf{y}_{i,t-1}$. When local node-wise feature weights are available, PHIDA uses them only as a nonnegative penalty relative to the global distance in the transformed space.

PHIDA stores a scalar scale $s_{i,t}$ for each node. PHIDA uses the inverse-distance scale as
\begin{equation}
  \alpha_{i,t}
  =
  \frac{1}
       {\max\{s_{i,t-1},\,10^{-12}\}},
  \label{eq:phida_appendix_alpha_scale}
\end{equation}
where PHIDA uses the nonfinite or nonpositive fallback value $1$. Since stored scales are floored at $10^{-6}$, this fallback is only a numerical guard. PHIDA obtains the stored scale from a global Welford estimate over raw streamed samples, not from transformed samples. Let $\widehat{\sigma}_{j,t}^{\mathrm{raw}}$ be the global standard deviation in the raw space for feature $j$ after incorporating the current sample into the global Welford state. Before two samples are available this standard deviation is zero. PHIDA defines
\begin{equation}
  s_t^\star
  =
  \max\left\{
    \max_{1\le j\le d}
    \widehat{\sigma}_{j,t}^{\mathrm{raw}},
    10^{-6}
  \right\}.
  \label{eq:phida_appendix_node_scale_state}
\end{equation}
When a new node is created, PHIDA sets $s_{K_t,t}\leftarrow s_t^\star$. During the cold start with two nodes, after the second node is created, the first node scale is reset to match the second, $s_{1,t}\leftarrow s_{2,t}$. When the winner node is updated, its stored scale is refreshed as $s_{r_t,t}\leftarrow s_t^\star$. All other node scales are inherited unless explicitly updated, $s_{i,t}\leftarrow s_{i,t-1}$. The runner-up secondary update does not refresh this global scale state. Therefore, if $b_t$ is updated only as the runner-up, then $s_{b_t,t}\leftarrow s_{b_t,t-1}$.

PHIDA selects the winner node index by the smallest distance in the transformed space,
\begin{equation}
  r_t=\arg\min_i d_{i,t}.
  \label{eq:phida_appendix_winner_node}
\end{equation}
The nearest node is selected by distance, whereas the vigilance test is evaluated using the corresponding inverse-distance similarity.

\subsection{Vigilance Test and Node Learning}

IDA applies the vigilance test with the current vigilance parameter,
\begin{equation}
  a_{r_t,t} \ge \tau_t,
  \label{eq:phida_appendix_acceptance_condition}
\end{equation}
where $\tau_t$ is the vigilance parameter, which PHIDA updates adaptively from recent inverse-distance similarities and the current cluster-to-node ratio used by PHIDA.

PHIDA creates nodes directly until the initial state with three nodes is available. After that, if $a_{r_t,t}<\tau_t$, IDA creates a new node as
\begin{equation}
  K_t\leftarrow K_{t-1}+1,
  \quad
  \mathbf{y}_{K_t,t}\leftarrow\mathbf{x}_t,
  \quad
  M_{K_t,t}\leftarrow 1,
  \quad
  s_{K_t,t}\leftarrow s_t^\star.
  \label{eq:phida_appendix_new_node_update}
\end{equation}
Existing node states are inherited for indices $i<K_t$.

If $a_{r_t,t}\ge\tau_t$, IDA updates the winner node as
\begin{align}
  \mathbf{y}_{r_t,t}
  &\leftarrow
  \mathbf{y}_{r_t,t-1}
  +\eta_{r_t,t}(\mathbf{x}_t-\mathbf{y}_{r_t,t-1}),
  \label{eq:phida_appendix_winner_centroid_update} \\
  M_{r_t,t}&\leftarrow M_{r_t,t-1}+1,
  \label{eq:phida_appendix_winner_support_update}
\end{align}
where the default winner update uses the learning rate based on support, $\eta_{r_t,t}=1/M_{r_t,t}$. Let
\begin{equation}
  b_t
  =
  \arg\min_{i\ne r_t} d_{i,t}.
  \label{eq:phida_appendix_runnerup_condition}
\end{equation}
PHIDA considers the runner-up update only when
\begin{equation}
  a_{b_t,t}>\tau_t,
  \qquad
  b_t\ne r_t,
  \qquad
  \zeta_t(r_t,b_t)=1.
  \label{eq:phida_appendix_runnerup_gate}
\end{equation}
and the condition for compatibility with the PH component view holds. Here, $\zeta_t(r_t,b_t)\in\{0,1\}$ denotes the PH compatibility gate for the winner and runner-up pair. PHIDA sets $\zeta_t(r_t,b_t)=1$ if either no current PH component view is available, the number of raw PH components is less than $2$, $K_t<\max\{2\lambda_t,8\}$, or both $r_t$ and $b_t$ occur in the current PH component map and have the same raw PH component id. Otherwise, PHIDA sets $\zeta_t(r_t,b_t)=0$.

When the condition holds, PHIDA updates the runner-up support, confidence factor, and secondary learning rate as
\begin{align}
  M_{b_t,t}
  &=
  M_{b_t,t-1}+1, \\
  \chi_{b_t,t}
  &=
  \operatorname{clip}
  \left(
    \frac{a_{b_t,t}-\tau_t}
         {\max\{1-\tau_t,10^{-12}\}},
    0,
    1
  \right), \\
  \eta^{(2)}_{b_t,t}
  &=
  \frac{\chi_{b_t,t}}{M_{b_t,t}}.
  \label{eq:phida_appendix_runnerup_confidence}
\end{align}
PHIDA then updates the runner-up representative as
\begin{equation}
  \mathbf{y}_{b_t,t}
  \leftarrow
  \mathbf{y}_{b_t,t-1}
  +
  \eta^{(2)}_{b_t,t}
  \left(
    \mathbf{x}_t-\mathbf{y}_{b_t,t-1}
  \right).
  \label{eq:phida_appendix_runnerup_update}
\end{equation}
This secondary update increments the runner-up support and updates its prototype and moment state local to the node, but it does not update the global raw Welford state or the stored node scale $s_{b_t,t}$. Other nodes are carried over unchanged.

\subsection{Adaptive Vigilance and PH Refresh}

PHIDA initializes the vigilance state as
\begin{equation}
  \lambda_0=2,
  \qquad
  \tau_0=0.
  \label{eq:phida_appendix_vigilance_init}
\end{equation}
It maintains a ring buffer $B_t$ of recent raw samples. After interval updates, PHIDA sets the logical retention length as
\begin{equation}
  L_t^{B}=2\lambda_t.
  \label{eq:phida_appendix_buffer_retention}
\end{equation}
Thus, after insertion and trimming, $|B_t|\le L_t^{B}=2\lambda_t$, and $B_t$ contains the most recent samples in order from oldest to newest. The physical storage may allocate extra headroom, but this has no algorithmic effect. During the initial cold start, before the first eligible interval update, the buffer may remain untrimmed. Under $\lambda_0=2$ and the cold start with three nodes, this affects only the first few samples. For any $m$, let $W_t^{(m)}$ denote the $m$ most recent samples retrieved from $B_t$ in order from oldest to newest. In the decremental scan, PHIDA processes the current $\lambda_t$ window from newest to oldest, so the tested subwindow $W_t^{(m)}$ contains the $m$ most recent samples.

PHIDA defines the buffer scale as
\begin{equation}
  s_t^{\max}
  =
  \max_{1\le i\le K_t}s_{i,t},
  \qquad
  \alpha_t^{B}
  =
  \frac{1}
       {\max\{s_t^{\max},10^{-12}\}}.
  \label{eq:phida_appendix_buffer_scale}
\end{equation}
When no valid node scale is available, PHIDA uses the fallback $s_t^{\max}=1$. After applying the current adaptive feature transformation used during training, $\Phi_t$, to the samples in $W_t^{(m)}=(\mathbf{v}_1,\dots,\mathbf{v}_m)$, PHIDA computes the buffer similarity matrix as
\begin{equation}
  R_t^{(m)}[p,q]
  =
  \frac{
    1
  }{
    1+
    \alpha_t^{B}
    \left\|
      \Phi_t(\mathbf{v}_p)-\Phi_t(\mathbf{v}_q)
    \right\|_2
  },
  \label{eq:phida_appendix_buffer_similarity_matrix}
\end{equation}
where $\mathbf{v}_p$ and $\mathbf{v}_q$ denote raw samples from the selected recent window, and where PHIDA uses transformed Euclidean distance only and does not add the node-wise local feature penalty used in node matching and PH graph construction.

PHIDA declares a window $W_t^{(m)}$ stable if Cholesky factorization succeeds,
\begin{equation}
  R_t^{(m)}
  =
  L_t^{(m)}
  L_t^{(m)\top},
\end{equation}
and
\begin{equation}
  \Delta_t^{(m)}
  =
  \left(
    \prod_{\ell=1}^{m}
    L_t^{(m)}[\ell,\ell]
  \right)^2
  \ge
  \Delta_{\det},
  \qquad
  \Delta_{\det}=10^{-6},
  \label{eq:phida_appendix_stability_proxy}
\end{equation}
where PHIDA declares the window unstable if Cholesky factorization fails or if the determinant proxy is below $\Delta_{\det}$.

For a selected window $W_t^{(m)}$, PHIDA sets the diagonal of $R_t^{(m)}$ to $-\infty$ and computes the nearest neighbor similarities as
\begin{equation}
  u_p^{(m)}
  =
  \max_{q\ne p}R_t^{(m)}[p,q],
  \qquad
  p=1,\dots,m.
  \label{eq:phida_appendix_window_nn_similarity}
\end{equation}
Let $K_t$ be the number of current learned nodes and let $C_t$ be the current cluster count used for the threshold update. PHIDA uses the current output cluster count when it is available, and otherwise falls back to the raw PH component count, the current component count, or an effective count based on support. PHIDA computes the current cluster-to-node ratio as
\begin{equation}
  r_t
  =
  \operatorname{clip}
  \left(
    \frac{C_t}{K_t},
    0,
    1
  \right).
  \label{eq:phida_appendix_cluster_node_ratio}
\end{equation}
Given an interval argument $L_{\mathrm{mix}}$, PHIDA updates the smoothed ratio and vigilance threshold as
\begin{equation}
  \bar r_t
  =
  \operatorname{clip}
  \left(
    \frac{1}{L_{\mathrm{mix}}}r_t
    +
    \left(1-\frac{1}{L_{\mathrm{mix}}}\right)\bar r_{t-1},
    0,
    1
  \right),
  \qquad
    \tau_t
  =
  Q_{\bar r_t}^{\mathrm{Hazen}}
  \left(
    \{u_p^{(m)}\}_{p=1}^{m}
  \right).
  \label{eq:phida_appendix_similarity_threshold_update}
\end{equation}
Thus, the PH component view affects IDA node learning through the cluster-to-node ratio used to update the internal similarity threshold. In the decremental scan and in the helper used for tests of larger windows, $L_{\mathrm{mix}}$ is the current interval before reassignment. When an unstable larger window is followed by recomputation on the last stable smaller window, $L_{\mathrm{mix}}$ is the length of that smaller window.

At a recalculation event, let $\lambda=\lambda_t$. Recalculation is attempted only when the recalculation counter reaches $\lambda$ and the model has more than two learned nodes. PHIDA first runs the decremental scan on the latest $\lambda$ samples. It initializes
\begin{align}
  m_0=\min\{\lambda,|B_t|\},
  \\
  \lambda_{\mathrm{last}}\leftarrow\lambda,
  \\
  \tau_{\mathrm{last}}\leftarrow\tau_t.
\end{align}
For $m=2,\dots,m_0$, PHIDA tests the stability of $W_t^{(m)}$. If $W_t^{(m)}$ is unstable, it sets $\lambda_{t+}=\lambda_{\mathrm{last}}$ and $\tau_{t+}=\tau_{\mathrm{last}}$, and it does not run the incremental phase. Otherwise it updates $\lambda_{\mathrm{last}}\leftarrow m$ and recomputes $\tau_{\mathrm{last}}$ from $W_t^{(m)}$ using Eq.~\eqref{eq:phida_appendix_similarity_threshold_update}.

If no instability is detected in the decremental scan, PHIDA runs the incremental scan with
\begin{equation}
  m_{\max}=\min\{2\lambda,|B_t|\}.
\end{equation}
If $|B_t|\le\lambda$, PHIDA sets $\lambda_{t+}=\lambda$ and $\tau_{t+}=\tau_t$. Otherwise PHIDA initializes the lower tested length, upper tested length, and expansion stride as
\begin{align}
  \ell\leftarrow\lambda,
  \\
  h\leftarrow\min\{\lambda+1,m_{\max}\},
  \\
  \xi\leftarrow1.
\end{align}
While $h\le m_{\max}$, PHIDA tests $W_t^{(h)}$. If $W_t^{(h)}$ is unstable, it sets $\lambda_{t+}=\ell$ and recomputes $\tau_{t+}$ from $W_t^{(\ell)}$, then stops. If $W_t^{(h)}$ is stable, it sets
\begin{align}
  \ell\leftarrow h,
  \\
  \xi
  \leftarrow
  \min\{2\xi,\,m_{\max}-\lambda\}, \\
  h
  \leftarrow
  \min\{\lambda+\xi,\,m_{\max}\}.
\end{align}
If this update would make $h\le\ell$, the loop stops. If no unstable larger window is found, PHIDA sets $\lambda_{t+}=m_{\max}$ and recomputes $\tau_{t+}$ from $W_t^{(m_{\max})}$. This is the exhaustion case. Finally,
\begin{equation}
  \lambda_t\leftarrow\lambda_{t+},
  \qquad
  \tau_t\leftarrow\tau_{t+},
  \qquad
  L_t^{B}\leftarrow2\lambda_t.
  \label{eq:phida_appendix_lambda_update}
\end{equation}
PHIDA then resets the recalculation counter.

PHIDA also maintains flags for nodes active for prediction. A node becomes active for prediction when it resonates with the current sample and its support count is at least the median support count among nodes with support greater than one. PHIDA refreshes the PH partition and cached assignment view periodically according to $\lambda_t$ and once again after learning ends.

\subsection{Largest Gap Cut Rules}
\label{app:phida_cut_rule}

For completeness, PHIDA uses two cut rules based on largest gaps. PHIDA uses Eq.~\eqref{eq:phida_ph_largest_gap_threshold} to cut the density-guided $H_0$ persistence tree, where ties in the largest gap selection are broken toward the larger upper persistence level. This rule yields the raw PH components. PHIDA then uses Eq.~\eqref{eq:phida_merge_height_gap_level} to cut the PH-constrained component hierarchy, where only levels that retain at least $C_{\min}$ groups are eligible. Ties are again broken toward the larger upper merge height. PHIDA expands the selected component partition back to learned nodes and caches the resulting assignment view.

\subsection{Out-of-Sample Cluster Assignment}
\label{app:out_of_sample_assignment}

The out-of-sample cluster assignment procedure uses the cached PHIDA assignment view built during learning. This cached state stores the output clusters induced by $g^{(\ell^*)}$, the transformed representatives of learned nodes used for query assignment, and the information on node support needed to compute $h_c(\mathbf{x})$.

For out-of-sample cluster assignment, PHIDA uses the cached output clustering induced by $g^{(\ell^*)}$. Let
\begin{equation}
  \mathcal{C}_c
  =
  \{\, i \in \mathcal{K} \mid g^{(\ell^*)}(i) = c \,\},
  \qquad c = 1,\dots,C,
  \label{eq:phida_pred_clusters}
\end{equation}
denote the output clusters induced by $g^{(\ell^*)}$. Let $\tilde{\mathbf{x}}$ and $\tilde{\mathbf{y}}_i$ denote the query and representatives of learned nodes in the transformed feature space.

For each output cluster, PHIDA computes the total support as
\begin{equation}
  S_c
  =
  \sum_{i \in \mathcal{C}_c} M_i,
  \label{eq:phida_pred_support}
\end{equation}
and the normalized share of cluster support as
\begin{equation}
  p_c
  =
  \frac{S_c}{\sum_{r=1}^{C} S_r}.
  \label{eq:phida_pred_share}
\end{equation}

PHIDA computes the support concentration fraction as
\begin{equation}
  q
  =
  \sum_{c=1}^{C} p_c^2.
  \label{eq:phida_pred_concentration}
\end{equation}

For a query $\mathbf{x}$, PHIDA computes the distance to the nearest node in output cluster $c$ as
\begin{equation}
  d_{\min,c}(\mathbf{x})
  =
  \min_{i \in \mathcal{C}_c}
  \left\|
    \tilde{\mathbf{x}} - \tilde{\mathbf{y}}_i
  \right\|_2,
  \label{eq:phida_pred_dmin}
\end{equation}
where the norm is evaluated in the transformed feature space. Let $i_{(1)}, i_{(2)}, \dots, i_{(|\mathcal{C}_c|)}$ be the indices in $\mathcal{C}_c$ ordered by increasing distance from $\tilde{\mathbf{x}}$. The smallest index whose cumulative support reaches the fraction $q$ of the cluster support is
\begin{equation}
  j_c(\mathbf{x})
  =
  \min
  \left\{
    j \mid
    \sum_{r=1}^{j} M_{i_{(r)}} \ge q\, S_c
  \right\},
  \label{eq:phida_pred_jk}
\end{equation}
and PHIDA computes the corresponding cluster radius based on support as
\begin{equation}
  d_{q,c}(\mathbf{x})
  =
  \left\|
    \tilde{\mathbf{x}} - \tilde{\mathbf{y}}_{i_{(j_c(\mathbf{x}))}}
  \right\|_2.
  \label{eq:phida_pred_dq}
\end{equation}

For output clusters with more than one member node, PHIDA computes the assignment score as
\begin{equation}
  h_c(\mathbf{x})
  =
  d_{\min,c}(\mathbf{x}) + d_{q,c}(\mathbf{x}).
  \label{eq:phida_pred_score}
\end{equation}

For an output cluster containing a single learned node, $\mathcal{C}_c=\{i\}$, PHIDA uses the squared distance in the transformed space
\begin{equation}
  h_c(\mathbf{x})
  =
  \left\|
    \tilde{\mathbf{x}}-\tilde{\mathbf{y}}_i
  \right\|_2^2.
  \label{eq:phida_pred_singleton_score}
\end{equation}

PHIDA assigns the output cluster index by
\begin{equation}
  \hat{c}(\mathbf{x})
  =
  \min\left\{
  c\in\{1,\ldots,C\}
  \mid
  h_c(\mathbf{x})=\min_{r\in\{1,\ldots,C\}}h_r(\mathbf{x})
  \right\}.
  \label{eq:phida_pred_label}
\end{equation}

The first term in $h_c(\mathbf{x})$ captures proximity to the nearest cluster member. The second term incorporates the local support structure inside the same output cluster.

\subsection{Pseudocode}
\label{app:appendix_pseudocode}

\setcounter{algocf}{0}
\renewcommand{\thealgocf}{A\arabic{algocf}}

The following algorithms summarize the detailed PHIDA procedures used by the main method.

Algorithm~\ref{alg:ida_online_update} shows the online IDA update component used within PHIDA.

\begin{algorithm}[htbp]
\caption{IDA online update}
\label{alg:ida_online_update}
\KwInput{stream $\{\mathbf{x}_t\}_{t=1}^{T}$, state of the vigilance parameter, and maintenance flags for refresh, deletion, and pruning of the PH input}
\KwOutput{state of learned nodes $(Y_T,\{M_{i,T}\})$ and cached PHIDA view state}
Initialize $Y_0=\emptyset$, node states, flags for nodes active for prediction, and vigilance state using Eqs.~\eqref{eq:phida_appendix_vigilance_init} and \eqref{eq:phida_appendix_buffer_retention}\;
\ForEach{sample $\mathbf{x}_t$}{
  Append $\mathbf{x}_t$ to $B_t$ using Eq.~\eqref{eq:phida_appendix_buffer_retention}\;
  Apply the current transform by Eq.~\eqref{eq:phida_adaptive_transform}\;
  Compute distances and similarities using Eqs.~\eqref{eq:phida_appendix_alpha_scale} and \eqref{eq:phida_appendix_match_signal}\;
  Select $r_t$ by Eq.~\eqref{eq:phida_appendix_winner_node}, and evaluate Eq.~\eqref{eq:phida_appendix_acceptance_condition}\;
  \eIf{the initial state with three nodes is not yet available or no active node passes the vigilance test under the current vigilance parameter}{
    Create a new node by Eq.~\eqref{eq:phida_appendix_new_node_update}\;
  }{
    Update the winner using Eqs.~\eqref{eq:phida_appendix_node_scale_state} and \eqref{eq:phida_appendix_winner_centroid_update}--\eqref{eq:phida_appendix_winner_support_update}\;
    \If{the runner-up gate in Eq.~\eqref{eq:phida_appendix_runnerup_gate} holds}{
      Compute the runner-up confidence factor and apply the secondary update by Eqs.~\eqref{eq:phida_appendix_runnerup_confidence}--\eqref{eq:phida_appendix_runnerup_update}\;
    }
  }
  \If{the recalculation counter reaches $\lambda_t$}{
    Compute the buffer scale and Cholesky stability proxy by Eqs.~\eqref{eq:phida_appendix_buffer_scale}--\eqref{eq:phida_appendix_stability_proxy}\;
    Update $\lambda_t$, $\tau_t$, and the buffer retention rule by Eqs.~\eqref{eq:phida_appendix_similarity_threshold_update}--\eqref{eq:phida_appendix_lambda_update}\;
  }
  \If{the refresh flag is enabled and a periodic PH refresh is due}{
    Rebuild the PH graph and raw PH components using Eqs.~\eqref{eq:phida_knn_graph}--\eqref{eq:phida_raw_ph_component_partition}\;
    Refresh the assignment view using Eqs.~\eqref{eq:phida_ph_stable_mass}--\eqref{eq:phida_merge_height_gap_level}\;
  }
}
\Return{$(Y_T,\{M_{i,T}\})$ and the cached PHIDA view}\;
\end{algorithm}

Algorithm~\ref{alg:phida_cluster_structure} shows the PHIDA node-to-cluster mapping procedure.

\begin{algorithm}[htbp]
\caption{PHIDA node-to-cluster mapping}
\label{alg:phida_cluster_structure}
\KwInput{representatives of learned nodes $Y=\{\mathbf{y}_i\}_{i=1}^{K}$, support counts $\{M_i\}$, flags for nodes active for prediction, adaptive feature transformation state, and flag for pruning the PH input}
\KwOutput{mapping $g^{(\ell^*)}$ and the induced output clustering}
Select nodes active for prediction if available, and otherwise use all current nodes\;
Build the mutual $k$NN graph using Eq.~\eqref{eq:phida_knn_size} and Appendix~\ref{app:phida_knn_pruning}\;
If the flag for pruning the PH input is enabled, remove isolated nodes used as PH input subject to the minimum retention rule\;
Set $\rho_i=\log M_i$ by Eq.~\eqref{eq:phida_node_support_density}, and activate nodes in descending $\rho_i$\;
Compute the density-guided $H_0$ persistence parent tree\;
Choose the persistence threshold by Eq.~\eqref{eq:phida_ph_largest_gap_threshold}\;
Cut the persistence tree to obtain $\mathcal{B}$ in Eq.~\eqref{eq:phida_raw_ph_component_partition}\;
Compute PH-stable masses, $C_{\mathrm{ent}}$, and $C_{\min}$ by Eqs.~\eqref{eq:phida_ph_stable_mass}--\eqref{eq:phida_min_retained_cluster_count}\;
Build the component $k$NN graph using Appendix~\ref{app:phida_knn_pruning}\;
\While{more than $C_{\min}$ groups remain and at least one eligible adjacent pair exists}{
  Merge the eligible adjacent groups with the smallest height in Eq.~\eqref{eq:phida_ward_merge_height}\;
  Update the component graph and record the merge height\;
}
Select the partition before the largest eligible jump using Eq.~\eqref{eq:phida_merge_height_gap_level}\;
Expand the selected component partition back to learned nodes\;
\Return{the induced mapping $g^{(\ell^*)}$ in Eq.~\eqref{eq:phida_ph_mapping_set}}\;
\end{algorithm}

Algorithm~\ref{alg:phida_maintenance_refresh} shows the maintenance and refresh cycle in PHIDA during learning.

\begin{algorithm}[htbp]
\caption{PHIDA maintenance and refresh cycle during learning}
\label{alg:phida_maintenance_refresh}
\KwInput{current IDA state, maintenance interval, refresh flag, deletion flag, and flag for pruning the PH input}
\KwOutput{refreshed node state and cached PHIDA assignment view}
\If{the refresh flag is enabled and maintenance is triggered}{
  \If{the deletion flag is enabled}{
    Remove nodes whose support remains $M_i=1$ under the IDAT style rule for low support nodes, while preserving sparse PH components when required\;
  }
  Rebuild the raw PH component partition using Eqs.~\eqref{eq:phida_knn_graph}--\eqref{eq:phida_raw_ph_component_partition}, with filtering of isolated nodes from the PH input controlled by the flag for pruning the PH input\;
  \If{the deletion flag is enabled and an isolated node pruned by the PH step has support $M_i=1$ and is inactive for prediction}{
    Physically remove the node from the active state\;
  }
  Rebuild and cache the PH-constrained assignment view using Eqs.~\eqref{eq:phida_ph_stable_mass}--\eqref{eq:phida_merge_height_gap_level}\;
}
\If{learning end is reached}{
  Perform the final PH build and rebuild of the assignment view using the same deletion flag and flag for pruning the PH input before returning the PHIDA output state\;
}
\Return{the refreshed node state and cached PHIDA assignment view}\;
\end{algorithm}

Algorithm~\ref{alg:phida_out_of_sample_assignment} shows the out-of-sample cluster assignment procedure in Appendix~\ref{app:out_of_sample_assignment}.

\begin{algorithm}[htbp]
\caption{Out-of-sample cluster assignment in PHIDA}
\label{alg:phida_out_of_sample_assignment}
\KwInput{sample $\mathbf{x}\in\mathbb{R}^d$ and cached cluster assignment state with output clusters $\{\mathcal{C}_c\}_{c=1}^{C}$, transformed representatives of learned nodes $\{\tilde{\mathbf{y}}_i\}$, node support counts $\{M_i\}$, and the learned feature transformation}
\KwOutput{assigned output cluster index $\hat{c}(\mathbf{x})$}
Transform the query sample into the learned metric space and denote it by $\tilde{\mathbf{x}}$\;
Recover the output clusters induced by $g^{(\ell^*)}$ by Eq.~\eqref{eq:phida_pred_clusters}\;
Compute the cluster support totals and support shares by Eqs.~\eqref{eq:phida_pred_support} and \eqref{eq:phida_pred_share}\;
Compute the support concentration depth $q$ by Eq.~\eqref{eq:phida_pred_concentration}\;
\ForEach{output cluster $c\in\{1,\ldots,C\}$}{
  Compute the distance from $\tilde{\mathbf{x}}$ to the nearest node in cluster $c$ by Eq.~\eqref{eq:phida_pred_dmin}\;
  For clusters with multiple nodes, compute the index for reaching the required support and the support radius distance by Eqs.~\eqref{eq:phida_pred_jk} and \eqref{eq:phida_pred_dq}\;
  Compute the cluster score $h_c(\mathbf{x})$ by Eq.~\eqref{eq:phida_pred_singleton_score} for singleton clusters and by Eq.~\eqref{eq:phida_pred_score} otherwise\;
}
Assign the final cluster label by Eq.~\eqref{eq:phida_pred_label}\;
\Return{$\hat{c}(\mathbf{x})$}\;
\end{algorithm}

\section{Finite-Sample Algorithmic Properties}
\label{app:phida_properties}

This appendix records supporting finite-sample algorithmic properties of PHIDA. The statements do not claim statistical consistency, distributional optimality, or global optimality of the returned clustering. They clarify what follows from preserving raw PH components during node-to-cluster mapping and from using the implemented merge and assignment rules.

\paragraph{Proposition 1 (Hierarchy preserving raw components and reduced cut search).}
Let $\mathcal{K}_{\mathrm{ret}}\subseteq\{1,\ldots,K\}$ be the retained index set of learned nodes used by the PHIDA node-to-cluster mapping stage, and let $m=|\mathcal{K}_{\mathrm{ret}}|$. Let $\mathcal{B}=\{B_a\}_{a=1}^{K_{\mathrm{PH}}}$ be the raw PH component partition of $\mathcal{K}_{\mathrm{ret}}$, and let $\mathcal{H}_{\mathrm{W}}=\{\mathcal{P}^{(\ell)}\}_{\ell=0}^{L}$ be the component hierarchy under the PH constraint initialized at $\mathcal{P}^{(0)}=\mathcal{B}$. Assume that every transition $\mathcal{P}^{(\ell)}\rightarrow\mathcal{P}^{(\ell+1)}$ is obtained only by merging current groups and never by splitting a raw PH component. Then every mapping induced by a cut in $\mathcal{M}_{\mathrm{PH}}(\mathcal{H}_{\mathrm{W}})$ is constant on each raw PH component. Equivalently, every output cluster is a union of one or more raw PH components. The hierarchy is a nested chain of coarsenings, and the returned mapping $g^{(\ell^*)}$ belongs to $\mathcal{M}_{\mathrm{PH}}(\mathcal{H}_{\mathrm{W}})$. Moreover, the rule that cuts at the largest eligible gap between merge heights searches at most $L+1\le K_{\mathrm{PH}}\le m\le K$ candidate partitions, instead of the unrestricted set of all $\operatorname{Bell}(m)$ partitions of the $m$ retained learned nodes.

\textit{Proof.}
The initial partition is the raw PH component partition $\mathcal{B}$. Every later hierarchy level is produced only by merging existing groups. Therefore no later level splits any $B_a$. Each group at every hierarchy level is a union of raw PH components, and every mapping induced by a cut is constant on each $B_a$. The construction using only merges also implies that $\mathcal{P}^{(\ell+1)}$ is a coarsening of $\mathcal{P}^{(\ell)}$. Hence the hierarchy is nested and the number of groups is nonincreasing. Since $g^{(\ell^*)}$ is induced by one selected cut of $\mathcal{H}_{\mathrm{W}}$, it belongs to $\mathcal{M}_{\mathrm{PH}}(\mathcal{H}_{\mathrm{W}})$ by definition. If consecutive recorded levels are strict coarsenings, each transition decreases the number of groups by at least one. Starting from $K_{\mathrm{PH}}\le m$ raw PH components, the hierarchy has at most $K_{\mathrm{PH}}$ recorded cuts. Hence $L+1\le K_{\mathrm{PH}}\le m\le K$. The unrestricted number of partitions of the retained set of learned nodes is $\operatorname{Bell}(m)$. The implemented cut search is therefore over the recorded hierarchy cuts rather than over all unrestricted partitions. \hfill$\square$

\paragraph{Proposition 2 (PHIDA merge cost equals weighted dispersion increase under the PH constraint).}
Consider a current PH-constrained component partition. Let $w_a>0$ be the actual merge weight used by PHIDA for raw PH component $B_a$ after applying the fallback rule if necessary. For a current group $G$, define $W_G=\sum_{a\in G}w_a>0$ and let $\boldsymbol{\mu}_G$ be its weighted centroid in the transformed feature space. Define the weighted dispersion within groups of a partition $\mathcal{P}$ over raw PH component centroids as
\begin{equation}
  \operatorname{WCSS}(\mathcal{P})
  =
  \sum_{G\in\mathcal{P}}
  \sum_{a\in G}
  w_a\left\|\mathbf{z}_a-\boldsymbol{\mu}_G\right\|_2^2,
  \label{eq:phida_wcss_definition}
\end{equation}
where $\mathbf{z}_a$ is the centroid in the transformed space of raw PH component $B_a$. If two eligible adjacent groups $A$ and $B$ are merged, the increase in $\operatorname{WCSS}$ is
\begin{equation}
  \operatorname{WCSS}(\mathcal{P}_{A\cup B})
  -
  \operatorname{WCSS}(\mathcal{P})
  =
  \frac{W_AW_B}{W_A+W_B}
  \left\|\boldsymbol{\mu}_A-\boldsymbol{\mu}_B\right\|_2^2.
  \label{eq:phida_ward_wcss_increase}
\end{equation}
Consequently, at each agglomeration step, the PHIDA merge rule greedily selects the eligible adjacent merge with the smallest finite increase in weighted dispersion within components among the current candidate pairs.

\textit{Proof.}
Only the two groups $A$ and $B$ change when they are merged. The standard weighted centroid identity gives
\begin{equation}
  \sum_{a\in A\cup B} w_a\left\|\mathbf{z}_a-\boldsymbol{\mu}_{A\cup B}\right\|_2^2
  -
  \sum_{a\in A} w_a\left\|\mathbf{z}_a-\boldsymbol{\mu}_{A}\right\|_2^2
  -
  \sum_{a\in B} w_a\left\|\mathbf{z}_a-\boldsymbol{\mu}_{B}\right\|_2^2
  =
  \frac{W_AW_B}{W_A+W_B}
  \left\|\boldsymbol{\mu}_A-\boldsymbol{\mu}_B\right\|_2^2.
  \label{eq:phida_ward_decomposition}
\end{equation}
All other groups make the same contribution before and after the merge. This is the merge height used by PHIDA. The same algebraic form appears in Ward's weighted dispersion criterion, but PHIDA applies it to raw PH components and restricts candidate pairs to eligible adjacent groups. Therefore the greedy PHIDA step minimizes the dispersion increase over the constrained candidate set available at that step. \hfill$\square$

\paragraph{Proposition 3 (Deterministic out-of-sample cluster assignment and margin stability).}
Assume that every output cluster has positive total support and that all transformed query and node representative vectors are finite. Then, for every query $\mathbf{x}$, the numerical values $h_c(\mathbf{x})$ used by the PHIDA cluster assignment rule are finite for all output clusters $c$. The implemented rule for resolving ties assigns the smallest cluster index among the minimizers of $h_c(\mathbf{x})$. Therefore the learned PHIDA clusterer returns exactly one output cluster index for every query. Moreover, if the minimizing output cluster $c^*(\mathbf{x})=\arg\min_c h_c(\mathbf{x})$ is unique with margin
\begin{equation}
  \mu(\mathbf{x})
  =
  \min_{c\neq c^*(\mathbf{x})}
  \bigl(h_c(\mathbf{x})-h_{c^*(\mathbf{x})}(\mathbf{x})\bigr)
  >0,
  \label{eq:phida_assignment_margin}
\end{equation}
and perturbed values $\widetilde{h}_c(\mathbf{x})$ satisfy
\begin{equation}
  |\widetilde{h}_c(\mathbf{x})-h_c(\mathbf{x})|<\mu(\mathbf{x})/2
  \qquad\text{for all }c,
  \label{eq:phida_assignment_score_perturbation}
\end{equation}
then the assigned output cluster is unchanged.

\textit{Proof.}
Positive output cluster support implies $S_c>0$ for all $c$, and the normalized support shares satisfy $p_c>0$ and $\sum_c p_c=1$. Hence $q=\sum_c p_c^2$ satisfies $0<q\le 1$. For an output cluster containing more than one learned node, the cumulative support over the nodes ordered by distance increases monotonically from a positive value to $S_c$. Since $0<q\le1$, the smallest index whose cumulative support reaches $qS_c$ exists. The finiteness of transformed query and node representative vectors implies that $d_{\min,c}(\mathbf{x})$, $d_{q,c}(\mathbf{x})$, and $h_c(\mathbf{x})$ are finite. For an output cluster containing a single learned node, PHIDA uses the finite squared distance in the transformed space to that node. Therefore all numerical values used by the assignment rule are finite. The deterministic rule for resolving ties then returns exactly one output cluster index.

For margin stability, suppose the minimizer is unique with margin $\mu(\mathbf{x})>0$. For any $c\neq c^*(\mathbf{x})$,
\begin{equation}
  \widetilde{h}_{c^*(\mathbf{x})}(\mathbf{x})
  <
  h_{c^*(\mathbf{x})}(\mathbf{x})+\mu(\mathbf{x})/2
  <
  h_c(\mathbf{x})-\mu(\mathbf{x})/2
  <
  \widetilde{h}_{c}(\mathbf{x}).
  \label{eq:phida_assignment_stability_inequality}
\end{equation}
Thus $c^*(\mathbf{x})$ remains the unique minimizer under the perturbed values. The assigned output cluster therefore remains unchanged. \hfill$\square$

\paragraph{Relation to connected components and separate mapping variants.}
The preceding properties distinguish PHIDA from clustering by connected components on the graph over learned nodes. A connected graph containing two dense node groups linked by a weak bridge has one connected component, whereas density-guided $H_0$ persistence may preserve two raw PH components, and the component hierarchy under the PH constraint may cut before the merge induced by the bridge. The PHIDA output therefore may differ from connected components on the same retained node state. Changing only the final node-to-cluster mapping after a completed PHIDA run evaluates a separate protocol because the state of learned nodes has already been influenced by the PH component view. Additional mapping comparisons assess states of learned nodes and mapping rules under separately defined protocols. Appendix~\ref{app:ida_component_mapping_control_protocol} and Appendix~\ref{app:ida_component_mapping_control_cd} report targeted comparisons based on the IDA component.

\section{Experimental Protocol and Reproducibility}
\label{app:reproducibility}

\subsection{Benchmark Protocol and Hyperparameter Selection}
\label{app:benchmark_protocol}

The experiments follow the benchmark protocol used for the IDAT comparison where applicable. All methods use the same dataset collection, dataset loaders, preprocessing assumptions, evaluation metrics, and handling of failed runs. Each result for a dataset, method, and setting is evaluated over 30 independent runs, indexed by the same 30 random seed identities across methods. The stationary setting evaluates each method on the same preprocessed datasets under this shared seed schedule. The nonstationary setting uses a staged stream construction. For a fixed seed, the class order, stage assignment, sample ordering, and evaluation stages are identical across methods.

Hyperparameter handling is separated by method type. PHIDA is evaluated with fixed implementation settings specified by the method description and the released configuration files. No PHIDA hyperparameter or PHIDA internal selection rule is selected by maximizing ARI with true labels. PHIDA computes the adaptive feature transformation, quantities for the vigilance parameter, PH graph, persistence threshold, component hierarchy, minimum retained cluster count, hierarchy cut, node-to-cluster mapping, and out-of-sample assignment from the input stream and state of learned nodes during learning.

For baseline methods with tunable benchmark configurations, the search space follows the IDAT comparison or the corresponding specification for each method used in that comparison. This group includes GNG, DDVFA, and CAEA in both stationary and nonstationary settings. It also includes TBM-P and SR-PCM-HDP in the stationary setting when they are included in the reported stationary comparison. Bayesian optimization selects these baseline hyperparameters by maximizing ARI under the common benchmark protocol.

IDAT, SOINN+, CAE, TC, and AuToMATo are run without explicit external hyperparameter optimization. TC and AuToMATo are evaluated only in the stationary setting in this study. TableDC and G-CEALS are run with the fixed deep clustering implementations described in the benchmark materials. These implementations use the dataset class count to specify cluster initialization or the number of clusters, but they are not selected by Bayesian optimization.

Ground-truth sample labels are not used inside any online clustering procedure during learning. For PHIDA, labels are used only to compute evaluation metrics. For tunable baselines, labels are used for hyperparameter selection and evaluation at the benchmark level. For TableDC and G-CEALS, the dataset class count is used as required by their fixed method specifications.

All reported stationary and nonstationary evaluations use the same 24 benchmark datasets summarized in Table~\ref{tab:app_dataset_stats}. The stationary protocol reports final ARI and AMI after learning. The exact per-dataset stationary values appear in Table~\ref{tab:app_stationary_quality}. The nonstationary protocol reports final ARI and AMI in Table~\ref{tab:app_nonstationary_final_quality} together with the streamwise summaries avgInc\_ARI and avgInc\_AMI in Table~\ref{tab:app_nonstationary_streamwise}. Table~\ref{tab:app_nonstationary_bwt} gives supplementary backward transfer summaries only. Table~\ref{tab:app_nodes_clusters_stationary} and Table~\ref{tab:app_nodes_clusters_nonstationary} report descriptive summaries of node counts and summaries of the number of clusters produced at the final evaluation point used for structure interpretation. Additional appendix CD diagrams report results for the mapping comparisons based on the IDA component. These diagrams compare PHIDA with IDA+Ward, IDA+HDBSCAN, and IDA+KM under stationary and nonstationary settings. The three additional methods use the true number of classes in their mapping step. They are therefore additional comparisons rather than baselines evaluated without using the true number of classes. All existing result and structural summary tables except Table~\ref{tab:app_dataset_stats} report means and standard deviations over the same repeated protocol with 30 runs.

Implementation details for reproducing PHIDA are distributed across the main Method section and Appendix Sections~\ref{app:appendix_pseudocode} and~\ref{app:phida_cut_rule}. In particular, PHIDA constructs an automatically pruned mutual $k$NN graph from the representatives of learned nodes, applies density-guided $H_0$ persistence using the density based on node support, $\rho_i=\log M_i$, cuts the resulting persistence tree by the largest persistence gap, and then builds the output cluster assignment view with a PH-constrained component hierarchy whose merge cost uses raw PH component support totals and a cut selected from the largest eligible gap between merge heights. Unavailable results remain marked as N/A in the appendix tables. For rank-based summaries, failed, unavailable, or all-NaN results receive the worst rank for the corresponding dataset and evaluation measure. Focused Wilcoxon tests use only finite paired ARI and AMI values. Holm correction is applied within each stated comparison family; the stationary PHIDA-vs-IDAT tests use a separate family of two comparisons from the nonstationary PHIDA-vs-IDAT tests.

\subsection{Ablation Variant Protocol}
\label{app:ablation_variant_protocol}

The ablation diagrams compare PHIDA with removals of one mechanism at a time. Unless a row states otherwise, each variant keeps the same IDA node update, vigilance update, PH-constrained mapping, out-of-sample assignment, benchmark protocol, and evaluation metrics as PHIDA.

\begin{table}[htbp]
\centering
\small
\setlength{\tabcolsep}{4pt}
\caption{Ablation variants used in Figures~\ref{fig:ablation_cd_nonstationary} and~\ref{fig:ablation_cd_stationary}.}
\label{tab:phida_ablation_variants}
\begin{tabular}{@{}p{0.12\textwidth}p{0.43\textwidth}p{0.39\textwidth}@{}}
\toprule
Variant & Disabled PHIDA step & Steps retained \\
\midrule
noRefresh
&
Disables the periodic refresh of the PH component view during the stream, i.e., the branch for periodic refresh in Algorithm~\ref{alg:ida_online_update} and the refresh cycle triggered by maintenance in Algorithm~\ref{alg:phida_maintenance_refresh}.
&
Retains the PH build after learning ends and the rebuild of the assignment view. Node learning, vigilance updates, and out-of-sample assignment are unchanged. \\
\addlinespace
noDelete
&
Disables physical deletion of learned nodes during maintenance, including the deletion step for low support nodes and the physical removal of eligible isolated nodes in Algorithm~\ref{alg:phida_maintenance_refresh}.
&
Retains periodic PH refresh, filtering of isolated nodes from the PH input, PH-constrained component mapping, and the final assignment view. Nodes may be excluded from the PH input view but are not physically removed by the maintenance deletion rules. \\
\addlinespace
noPrune
&
Disables filtering of isolated nodes from the PH input in Algorithm~\ref{alg:phida_cluster_structure} and Algorithm~\ref{alg:phida_maintenance_refresh}.
&
Retains the mutual $k$NN candidate distance pruning rule in Appendix~\ref{app:phida_knn_pruning}, periodic PH refresh, physical deletion rules, and PH-constrained mapping. Thus, noPrune refers to pruning of isolated nodes from the PH input, not to the local/global distance pruning inside the mutual $k$NN builder. \\
\bottomrule
\end{tabular}
\end{table}

PHIDA-noPH is defined in Section~\ref{sec:results}: it removes the raw PH component view from learning and mapping, including the behavior controlled by compatibility with the PH component view during learning and the PH-constrained node-to-cluster mapping rule. The noRefresh, noDelete, and noPrune variants are secondary ablations that keep the raw PH component view except for the disabled step listed in Table~\ref{tab:phida_ablation_variants}.

\subsection{Protocol for Mapping Comparisons Based on the IDA Component}
\label{app:ida_component_mapping_control_protocol}

The additional comparison methods are evaluated in both stationary and nonstationary settings on the same 24 datasets. They use the IDA component without the PH component view, PH component view refreshes, or the PHIDA node-to-cluster mapping rule. The resulting representatives of learned nodes are then mapped to output clusters by Ward hierarchical clustering~\citep{ward63}, HDBSCAN~\citep{campello15}, or K-means~\citep{lloyd82}.

IDA+Ward applies Ward hierarchical clustering~\citep{ward63} to the representatives of learned nodes. In the stationary setting, the target number of output clusters is the true number of classes. In the nonstationary setting, the target at each evaluation stage is the number of classes observed up to that stage.

IDA+KM applies K-means~\citep{lloyd82} to the representatives of learned nodes. The number of K-means clusters is set to the same target used by IDA+Ward.

IDA+HDBSCAN constructs an HDBSCAN hierarchy~\citep{campello15} from the representatives of learned nodes. Because HDBSCAN does not directly take a target cluster count as input, the hierarchy is cut at the cluster count closest to the target among cuts with at least the target number of clusters. If no such cut is available, the closest available cut is used. In the nonstationary setting, the target is the number of classes observed up to the current evaluation stage.

These methods use the true number of classes. They are therefore additional comparisons rather than baselines evaluated without using the true number of classes. They are included to test whether conventional mapping applied after node construction by IDA matches PHIDA when the conventional mapping is given the true number of classes. For PHIDA, true labels remain used only to compute evaluation metrics.

\section{Additional Experimental Results}
\label{app:additional_experimental_results}

This section collects supplementary experimental material for Section~\ref{sec:results}. It reports benchmark dataset statistics, per-dataset values, auxiliary summaries for counts of learned nodes, output granularity, and runtime, discussion by dataset, and additional critical difference diagrams.

\subsection{Benchmark Datasets and Per-Dataset Values}
\label{app:benchmark_tables}

This subsection reports benchmark dataset statistics and per-dataset values for the stationary and nonstationary settings. Table~\ref{tab:app_dataset_stats} summarizes the benchmark dataset statistics. Table~\ref{tab:app_stationary_quality} reports the per-dataset stationary ARI and AMI values. Table~\ref{tab:app_nonstationary_final_quality} reports the per-dataset nonstationary final ARI and AMI values. Table~\ref{tab:app_nonstationary_streamwise} reports the nonstationary streamwise summaries. Table~\ref{tab:app_nonstationary_bwt} reports supplementary nonstationary backward transfer summaries. Across all auxiliary tables, \textnormal{N/A} follows the availability convention in Tables~\ref{tab:app_stationary_quality} and~\ref{tab:app_nonstationary_final_quality}. When a result for a dataset and method is unavailable because of infeasibility, timeout, memory failure, invalid hyperparameters, or failure to produce a predictive model, all corresponding auxiliary entries are also reported as \textnormal{N/A}.

\begin{table}[htbp]
\centering
\footnotesize
\setlength{\tabcolsep}{6pt}
\renewcommand{\arraystretch}{1.0}
\caption{Statistics of the benchmark datasets.}
\label{tab:app_dataset_stats}
\begin{tabular}{lrrr}
\toprule
Dataset & \#Samples & \#Attributes & \#Classes \\
\midrule
Iris & 150 & 4 & 3 \\
Seeds & 210 & 7 & 3 \\
Dermatology & 366 & 34 & 6 \\
Pima & 768 & 8 & 2 \\
Mice Protein & 1,077 & 77 & 8 \\
Binalpha & 1,404 & 320 & 36 \\
Yeast & 1,484 & 8 & 10 \\
Semeion & 1,593 & 256 & 10 \\
MSRA25 & 1,799 & 64 & 12 \\
Image Seg. & 2,310 & 19 & 7 \\
Rice & 3,810 & 7 & 2 \\
TUANDROMD & 4,464 & 241 & 2 \\
Phoneme & 5,404 & 5 & 2 \\
Texture & 5,500 & 40 & 11 \\
OptDigits & 5,620 & 64 & 10 \\
Statlog & 6,435 & 36 & 6 \\
Anuran Calls & 7,195 & 22 & 4 \\
Isolet & 7,797 & 617 & 26 \\
STL10 & 13,000 & 2,048 & 10 \\
MNIST10K & 10,000 & 4,096 & 10 \\
PenBased & 10,992 & 16 & 10 \\
Letter & 20,000 & 16 & 26 \\
Shuttle & 58,000 & 9 & 7 \\
Skin & 245,057 & 3 & 2 \\
\bottomrule
\end{tabular}
\par\vspace{0.35em}
\parbox{0.48\linewidth}{\footnotesize\raggedright
Anuran Calls dataset uses only the Family label.}
\end{table}

\begin{sidewaystable*}[htbp]
\centering
\scriptsize
\setlength{\tabcolsep}{3pt}
\renewcommand{\arraystretch}{0.95}
\caption{Per-dataset stationary clustering performance.}
\label{tab:app_stationary_quality}
\resizebox{\textheight}{!}{%
\begin{tabular}{llccccccccccccc}
\toprule
Dataset & Measure & GNG & SOINN+ & DDVFA & CAEA & TBM-P & CAE & IDAT & TC & SR-PCM-HDP & TableDC & G-CEALS & AuToMATo & PHIDA \\
\midrule
Iris & ARI & 0.480 (0.231) & 0.501 (0.096) & 0.567 (0.002) & 0.098 (0.011) & 0.568 (0.000) & 0.182 (0.142) & 0.624 (0.084) & 0.568 (0.000) & 0.713 (0.009) & 0.649 (0.093) & 0.348 (0.246) & 0.568 (0.000) & 0.702 (0.090) \\
 & AMI & 0.536 (0.242) & 0.603 (0.056) & 0.727 (0.012) & 0.357 (0.015) & 0.732 (0.000) & 0.369 (0.134) & 0.693 (0.054) & 0.732 (0.000) & 0.757 (0.006) & 0.731 (0.080) & 0.450 (0.304) & 0.732 (0.000) & 0.742 (0.059) \\
\midrule
Seeds & ARI & 0.016 (0.087) & 0.276 (0.191) & 0.395 (0.058) & 0.061 (0.008) & 0.595 (0.141) & 0.009 (0.007) & 0.493 (0.087) & 0.411 (0.000) & 0.702 (0.008) & 0.072 (0.088) & 0.075 (0.139) & 0.789 (0.000) & 0.587 (0.088) \\
 & AMI & 0.019 (0.103) & 0.351 (0.192) & 0.459 (0.033) & 0.291 (0.015) & 0.606 (0.127) & 0.064 (0.041) & 0.562 (0.051) & 0.527 (0.000) & 0.670 (0.007) & 0.127 (0.114) & 0.112 (0.164) & 0.747 (0.000) & 0.625 (0.044) \\
\midrule
Dermatology & ARI & 0.001 (0.004) & 0.080 (0.054) & 0.661 (0.124) & 0.368 (0.144) & 0.472 (0.116) & 0.099 (0.165) & 0.049 (0.035) & 0.328 (0.000) & 0.089 (0.000) & 0.038 (0.014) & 0.223 (0.093) & 0.000 (0.000) & 0.513 (0.116) \\
 & AMI & 0.001 (0.008) & 0.220 (0.082) & 0.763 (0.077) & 0.538 (0.124) & 0.624 (0.070) & 0.190 (0.185) & 0.161 (0.071) & 0.518 (0.000) & 0.182 (0.000) & 0.090 (0.030) & 0.351 (0.089) & 0.000 (0.000) & 0.628 (0.097) \\
\midrule
Pima & ARI & 0.001 (0.003) & 0.022 (0.018) & 0.046 (0.014) & 0.001 (0.000) & 0.072 (0.038) & 0.008 (0.004) & 0.037 (0.029) & 0.018 (0.000) & 0.020 (0.000) & 0.031 (0.027) & 0.060 (0.009) & 0.012 (0.000) & 0.080 (0.044) \\
 & AMI & 0.000 (0.000) & 0.025 (0.015) & 0.022 (0.008) & 0.037 (0.004) & 0.060 (0.019) & 0.040 (0.012) & 0.058 (0.020) & 0.039 (0.000) & 0.045 (0.000) & 0.016 (0.011) & 0.021 (0.004) & 0.001 (0.000) & 0.047 (0.027) \\
\midrule
Mice Protein & ARI & 0.024 (0.041) & 0.105 (0.098) & 0.039 (0.001) & 0.113 (0.009) & 0.111 (0.007) & 0.014 (0.025) & 0.241 (0.051) & 0.393 (0.000) & 0.242 (0.014) & 0.291 (0.028) & 0.253 (0.047) & 0.001 (0.000) & 0.238 (0.037) \\
 & AMI & 0.069 (0.094) & 0.257 (0.153) & 0.354 (0.004) & 0.405 (0.016) & 0.438 (0.013) & 0.111 (0.148) & 0.402 (0.046) & 0.610 (0.000) & 0.379 (0.018) & 0.462 (0.034) & 0.431 (0.054) & 0.024 (0.000) & 0.392 (0.036) \\
\midrule
Binalpha & ARI & 0.000 (0.000) & 0.019 (0.023) & 0.115 (0.056) & 0.194 (0.022) & 0.169 (0.009) & 0.176 (0.030) & 0.031 (0.037) & 0.120 (0.000) & 0.000 (0.000) & 0.251 (0.011) & 0.211 (0.014) & 0.006 (0.009) & 0.038 (0.037) \\
 & AMI & 0.000 (0.000) & 0.182 (0.075) & 0.389 (0.053) & 0.412 (0.027) & 0.400 (0.009) & 0.409 (0.018) & 0.205 (0.111) & 0.383 (0.000) & 0.000 (0.000) & 0.490 (0.010) & 0.447 (0.014) & 0.052 (0.056) & 0.200 (0.115) \\
\midrule
Yeast & ARI & 0.104 (0.065) & 0.007 (0.018) & 0.054 (0.041) & 0.038 (0.011) & 0.069 (0.022) & 0.079 (0.026) & 0.115 (0.058) & 0.041 (0.000) & 0.156 (0.000) & 0.048 (0.049) & 0.101 (0.039) & 0.012 (0.000) & 0.130 (0.038) \\
 & AMI & 0.198 (0.064) & 0.031 (0.035) & 0.132 (0.035) & 0.103 (0.023) & 0.197 (0.020) & 0.133 (0.027) & 0.203 (0.059) & 0.091 (0.000) & 0.197 (0.000) & 0.114 (0.076) & 0.187 (0.065) & 0.049 (0.000) & 0.226 (0.033) \\
\midrule
Semeion & ARI & 0.002 (0.008) & 0.083 (0.079) & 0.008 (0.002) & 0.120 (0.009) & 0.159 (0.019) & 0.107 (0.021) & 0.267 (0.135) & 0.048 (0.000) & 0.000 (0.000) & 0.193 (0.026) & 0.288 (0.029) & 0.244 (0.020) & 0.180 (0.084) \\
 & AMI & 0.008 (0.041) & 0.281 (0.097) & 0.066 (0.003) & 0.435 (0.008) & 0.451 (0.014) & 0.360 (0.010) & 0.473 (0.108) & 0.375 (0.000) & 0.000 (0.000) & 0.359 (0.020) & 0.426 (0.026) & 0.446 (0.015) & 0.377 (0.097) \\
\midrule
MSRA25 & ARI & 0.063 (0.029) & 0.287 (0.079) & 0.567 (0.049) & 0.201 (0.081) & 0.348 (0.031) & 0.239 (0.107) & 0.328 (0.086) & 0.653 (0.000) & 0.030 (0.000) & 0.243 (0.020) & 0.181 (0.053) & 0.118 (0.015) & 0.332 (0.082) \\
 & AMI & 0.242 (0.090) & 0.604 (0.065) & 0.810 (0.022) & 0.391 (0.103) & 0.677 (0.022) & 0.559 (0.095) & 0.631 (0.067) & 0.864 (0.000) & 0.101 (0.000) & 0.497 (0.020) & 0.393 (0.058) & 0.396 (0.033) & 0.618 (0.070) \\
\midrule
Image Seg. & ARI & 0.048 (0.052) & 0.156 (0.103) & 0.312 (0.106) & 0.199 (0.021) & 0.389 (0.058) & 0.263 (0.089) & 0.299 (0.086) & 0.102 (0.000) & 0.345 (0.024) & 0.014 (0.043) & 0.312 (0.067) & 0.381 (0.029) & 0.463 (0.076) \\
 & AMI & 0.163 (0.176) & 0.400 (0.085) & 0.533 (0.090) & 0.512 (0.013) & 0.596 (0.026) & 0.520 (0.046) & 0.545 (0.067) & 0.347 (0.000) & 0.535 (0.018) & 0.042 (0.097) & 0.445 (0.061) & 0.593 (0.009) & 0.611 (0.069) \\
\midrule
Rice & ARI & 0.267 (0.113) & 0.195 (0.064) & 0.110 (0.032) & 0.019 (0.001) & 0.037 (0.006) & 0.006 (0.007) & 0.296 (0.152) & 0.053 (0.000) & 0.586 (0.000) & 0.025 (0.014) & 0.084 (0.090) & 0.000 (0.000) & 0.395 (0.099) \\
 & AMI & 0.314 (0.096) & 0.276 (0.024) & 0.207 (0.011) & 0.186 (0.002) & 0.208 (0.004) & 0.109 (0.042) & 0.307 (0.090) & 0.205 (0.000) & 0.483 (0.000) & 0.063 (0.026) & 0.175 (0.059) & 0.000 (0.000) & 0.382 (0.048) \\
\midrule
TUANDROMD & ARI & -0.070 (0.041) & 0.071 (0.046) & 0.017 (0.009) & 0.054 (0.001) & 0.308 (0.123) & 0.093 (0.072) & 0.227 (0.308) & 0.026 (0.000) & 0.000 (0.000) & 0.178 (0.069) & 0.400 (0.323) & 0.051 (0.006) & 0.199 (0.222) \\
 & AMI & 0.154 (0.044) & 0.215 (0.042) & 0.050 (0.006) & 0.203 (0.004) & 0.235 (0.033) & 0.175 (0.086) & 0.238 (0.214) & 0.219 (0.000) & 0.000 (0.000) & 0.139 (0.085) & 0.352 (0.217) & 0.181 (0.011) & 0.269 (0.145) \\
\midrule
Phoneme & ARI & -0.012 (0.026) & -0.003 (0.022) & 0.060 (0.000) & 0.028 (0.010) & 0.125 (0.032) & 0.086 (0.029) & 0.105 (0.046) & 0.025 (0.000) & 0.065 (0.000) & -0.042 (0.068) & 0.050 (0.086) & 0.019 (0.077) & 0.057 (0.015) \\
 & AMI & 0.034 (0.022) & 0.031 (0.026) & 0.166 (0.000) & 0.102 (0.006) & 0.113 (0.011) & 0.112 (0.007) & 0.121 (0.021) & 0.103 (0.000) & 0.142 (0.000) & 0.108 (0.024) & 0.092 (0.055) & 0.037 (0.034) & 0.117 (0.006) \\
\midrule
Texture & ARI & 0.423 (0.046) & 0.153 (0.033) & 0.222 (0.016) & 0.373 (0.037) & 0.243 (0.021) & 0.508 (0.043) & 0.434 (0.100) & 0.925 (0.000) & 0.519 (0.005) & 0.237 (0.069) & 0.204 (0.035) & 0.092 (0.016) & 0.448 (0.108) \\
 & AMI & 0.626 (0.023) & 0.530 (0.045) & 0.527 (0.020) & 0.624 (0.012) & 0.577 (0.013) & 0.674 (0.027) & 0.689 (0.043) & 0.942 (0.000) & 0.670 (0.002) & 0.587 (0.052) & 0.423 (0.048) & 0.408 (0.035) & 0.662 (0.053) \\
\midrule
OptDigits & ARI & 0.413 (0.088) & 0.354 (0.164) & 0.004 (0.009) & 0.324 (0.043) & 0.193 (0.022) & 0.391 (0.063) & 0.670 (0.044) & 0.900 (0.000) & 0.000 (0.000) & 0.650 (0.059) & 0.684 (0.073) & 0.633 (0.002) & 0.758 (0.045) \\
 & AMI & 0.657 (0.039) & 0.623 (0.092) & 0.040 (0.054) & 0.458 (0.035) & 0.566 (0.013) & 0.578 (0.034) & 0.768 (0.019) & 0.922 (0.000) & 0.000 (0.000) & 0.733 (0.037) & 0.745 (0.049) & 0.807 (0.002) & 0.822 (0.020) \\
\midrule
Statlog & ARI & 0.183 (0.137) & 0.377 (0.152) & 0.422 (0.034) & 0.349 (0.075) & 0.164 (0.031) & 0.366 (0.058) & 0.383 (0.091) & 0.376 (0.000) & 0.520 (0.001) & 0.167 (0.053) & 0.411 (0.106) & 0.433 (0.000) & 0.511 (0.084) \\
 & AMI & 0.297 (0.179) & 0.510 (0.088) & 0.394 (0.038) & 0.442 (0.082) & 0.459 (0.008) & 0.499 (0.031) & 0.529 (0.037) & 0.550 (0.000) & 0.610 (0.001) & 0.351 (0.047) & 0.513 (0.064) & 0.622 (0.000) & 0.600 (0.042) \\
\midrule
Anuran Calls & ARI & 0.392 (0.140) & 0.324 (0.068) & 0.459 (0.076) & 0.032 (0.005) & 0.418 (0.063) & 0.204 (0.071) & 0.371 (0.089) & 0.473 (0.000) & 0.491 (0.002) & 0.054 (0.108) & 0.371 (0.091) & 0.510 (0.006) & 0.334 (0.094) \\
 & AMI & 0.385 (0.094) & 0.396 (0.027) & 0.439 (0.019) & 0.242 (0.015) & 0.440 (0.017) & 0.333 (0.027) & 0.424 (0.033) & 0.401 (0.000) & 0.403 (0.005) & 0.065 (0.106) & 0.368 (0.093) & 0.469 (0.015) & 0.444 (0.033) \\
\midrule
Isolet & ARI & 0.355 (0.061) & 0.139 (0.080) & 0.031 (0.001) & 0.222 (0.017) & 0.143 (0.007) & 0.080 (0.052) & 0.227 (0.063) & 0.435 (0.000) & 0.000 (0.000) & 0.252 (0.038) & 0.350 (0.025) & N/A & 0.304 (0.072) \\
 & AMI & 0.663 (0.038) & 0.498 (0.120) & 0.345 (0.003) & 0.521 (0.014) & 0.526 (0.006) & 0.324 (0.205) & 0.656 (0.046) & 0.692 (0.000) & 0.000 (0.000) & 0.569 (0.032) & 0.623 (0.018) & N/A & 0.681 (0.047) \\
\midrule
STL10 & ARI & 0.648 (0.089) & 0.256 (0.132) & 0.219 (0.026) & 0.271 (0.058) & 0.205 (0.020) & 0.000 (0.000) & 0.387 (0.174) & 0.185 (0.000) & 0.000 (0.000) & 0.197 (0.048) & 0.539 (0.069) & N/A & 0.481 (0.177) \\
 & AMI & 0.773 (0.039) & 0.508 (0.147) & 0.408 (0.011) & 0.457 (0.050) & 0.529 (0.011) & 0.000 (0.000) & 0.648 (0.132) & 0.443 (0.000) & 0.000 (0.000) & 0.421 (0.059) & 0.695 (0.024) & N/A & 0.686 (0.122) \\
\midrule
MNIST10K & ARI & 0.565 (0.192) & 0.480 (0.205) & 0.371 (0.090) & 0.666 (0.123) & 0.961 (0.040) & 0.322 (0.179) & 0.677 (0.054) & 0.294 (0.000) & 0.832 (0.065) & 0.313 (0.272) & 0.442 (0.115) & N/A & 0.800 (0.083) \\
 & AMI & 0.774 (0.091) & 0.739 (0.112) & 0.569 (0.103) & 0.826 (0.046) & 0.968 (0.011) & 0.624 (0.151) & 0.806 (0.022) & 0.694 (0.000) & 0.880 (0.020) & 0.583 (0.231) & 0.690 (0.083) & N/A & 0.891 (0.024) \\
\midrule
PenBased & ARI & 0.439 (0.078) & 0.458 (0.128) & 0.253 (0.046) & 0.370 (0.042) & 0.310 (0.042) & 0.565 (0.053) & 0.577 (0.047) & 0.772 (0.000) & 0.430 (0.062) & 0.501 (0.033) & 0.557 (0.040) & 0.631 (0.000) & 0.628 (0.059) \\
 & AMI & 0.666 (0.037) & 0.700 (0.053) & 0.547 (0.010) & 0.561 (0.036) & 0.607 (0.011) & 0.687 (0.027) & 0.693 (0.026) & 0.848 (0.000) & 0.582 (0.044) & 0.650 (0.020) & 0.668 (0.020) & 0.778 (0.000) & 0.743 (0.027) \\
\midrule
Letter & ARI & 0.016 (0.011) & 0.002 (0.002) & 0.034 (0.001) & 0.096 (0.004) & 0.077 (0.003) & 0.124 (0.017) & 0.011 (0.008) & 0.004 (0.000) & 0.001 (0.001) & 0.189 (0.013) & 0.163 (0.014) & 0.148 (0.002) & 0.067 (0.021) \\
 & AMI & 0.249 (0.048) & 0.089 (0.038) & 0.454 (0.002) & 0.459 (0.015) & 0.495 (0.004) & 0.468 (0.022) & 0.178 (0.051) & 0.167 (0.000) & 0.019 (0.013) & 0.412 (0.013) & 0.390 (0.012) & 0.520 (0.003) & 0.251 (0.050) \\
\midrule
Shuttle & ARI & 0.329 (0.217) & 0.689 (0.028) & 0.418 (0.179) & 0.007 (0.000) & 0.341 (0.064) & 0.586 (0.121) & 0.325 (0.175) & 0.356 (0.000) & 0.375 (0.136) & 0.001 (0.001) & 0.316 (0.176) & 0.685 (0.000) & 0.139 (0.049) \\
 & AMI & 0.325 (0.177) & 0.598 (0.022) & 0.494 (0.090) & 0.212 (0.002) & 0.452 (0.087) & 0.522 (0.083) & 0.292 (0.120) & 0.314 (0.000) & 0.287 (0.108) & 0.003 (0.002) & 0.244 (0.161) & 0.636 (0.000) & 0.325 (0.026) \\
\midrule
Skin & ARI & 0.060 (0.027) & 0.083 (0.032) & N/A & 0.006 (0.002) & N/A & 0.646 (0.070) & 0.263 (0.064) & N/A & 0.162 (0.043) & 0.661 (0.176) & 0.266 (0.326) & 0.001 (0.000) & 0.215 (0.062) \\
 & AMI & 0.238 (0.012) & 0.225 (0.016) & N/A & 0.154 (0.009) & N/A & 0.505 (0.050) & 0.333 (0.038) & N/A & 0.306 (0.011) & 0.602 (0.120) & 0.257 (0.283) & 0.117 (0.001) & 0.351 (0.040) \\
\bottomrule
\end{tabular}%
}
\par\vspace{0.35em}
\parbox{0.98\textheight}{\footnotesize\raggedright
The values in parentheses indicate the standard deviation over 30 independent runs. \textnormal{N/A} denotes unavailable results under the convention above.}
\end{sidewaystable*}

\begin{table*}[htbp]
\centering
\scriptsize
\setlength{\tabcolsep}{3pt}
\renewcommand{\arraystretch}{0.95}
\caption{Per-dataset nonstationary final clustering performance.}
\label{tab:app_nonstationary_final_quality}
\resizebox{\textwidth}{!}{%
\begin{tabular}{llccccccc}
\toprule
Dataset & Measure & GNG & SOINN+ & DDVFA & CAEA & CAE & IDAT & PHIDA \\
\midrule
Iris & ARI & 0.379 (0.181) & 0.556 (0.104) & 0.048 (0.050) & 0.108 (0.041) & 0.218 (0.227) & 0.565 (0.135) & 0.704 (0.124) \\
 & AMI & 0.460 (0.197) & 0.582 (0.061) & 0.138 (0.103) & 0.177 (0.070) & 0.262 (0.233) & 0.611 (0.108) & 0.731 (0.092) \\
\midrule
Seeds & ARI & 0.454 (0.195) & 0.307 (0.053) & 0.159 (0.039) & 0.305 (0.080) & 0.000 (0.000) & 0.484 (0.046) & 0.411 (0.082) \\
 & AMI & 0.499 (0.152) & 0.414 (0.028) & 0.180 (0.051) & 0.363 (0.004) & 0.000 (0.000) & 0.550 (0.020) & 0.539 (0.025) \\
\midrule
Dermatology & ARI & 0.031 (0.018) & 0.171 (0.047) & 0.258 (0.072) & 0.273 (0.185) & 0.000 (0.001) & 0.072 (0.063) & 0.486 (0.100) \\
 & AMI & 0.085 (0.028) & 0.373 (0.045) & 0.376 (0.085) & 0.408 (0.178) & 0.003 (0.008) & 0.179 (0.087) & 0.625 (0.086) \\
\midrule
Pima & ARI & 0.033 (0.021) & 0.013 (0.023) & -0.002 (0.013) & 0.003 (0.001) & -0.006 (0.008) & 0.042 (0.008) & 0.062 (0.013) \\
 & AMI & 0.053 (0.011) & 0.016 (0.006) & 0.008 (0.014) & 0.045 (0.001) & 0.020 (0.011) & 0.062 (0.009) & 0.063 (0.016) \\
\midrule
Mice Protein & ARI & 0.196 (0.024) & 0.243 (0.071) & 0.003 (0.003) & 0.113 (0.017) & 0.002 (0.003) & 0.254 (0.028) & 0.285 (0.035) \\
 & AMI & 0.375 (0.028) & 0.493 (0.027) & 0.036 (0.010) & 0.420 (0.019) & 0.035 (0.047) & 0.420 (0.033) & 0.489 (0.032) \\
\midrule
Binalpha & ARI & 0.024 (0.016) & 0.211 (0.037) & 0.002 (0.001) & 0.239 (0.014) & 0.095 (0.046) & 0.095 (0.069) & 0.102 (0.041) \\
 & AMI & 0.202 (0.062) & 0.444 (0.022) & 0.026 (0.014) & 0.472 (0.014) & 0.202 (0.069) & 0.334 (0.116) & 0.356 (0.062) \\
\midrule
Yeast & ARI & 0.094 (0.048) & 0.034 (0.029) & 0.014 (0.013) & 0.013 (0.002) & 0.077 (0.041) & 0.122 (0.067) & 0.132 (0.046) \\
 & AMI & 0.190 (0.043) & 0.080 (0.050) & 0.044 (0.018) & 0.137 (0.011) & 0.120 (0.047) & 0.203 (0.060) & 0.218 (0.045) \\
\midrule
Semeion & ARI & 0.131 (0.067) & 0.295 (0.054) & 0.029 (0.005) & 0.224 (0.023) & 0.085 (0.058) & 0.299 (0.064) & 0.361 (0.090) \\
 & AMI & 0.334 (0.067) & 0.495 (0.034) & 0.148 (0.044) & 0.505 (0.017) & 0.167 (0.070) & 0.463 (0.057) & 0.528 (0.069) \\
\midrule
MSRA25 & ARI & 0.361 (0.042) & 0.324 (0.030) & 0.061 (0.019) & 0.450 (0.039) & 0.163 (0.172) & 0.369 (0.047) & 0.563 (0.040) \\
 & AMI & 0.611 (0.031) & 0.640 (0.017) & 0.167 (0.034) & 0.683 (0.028) & 0.329 (0.207) & 0.593 (0.033) & 0.795 (0.023) \\
\midrule
Image Seg. & ARI & 0.253 (0.032) & 0.294 (0.091) & 0.101 (0.043) & 0.249 (0.025) & 0.309 (0.108) & 0.280 (0.083) & 0.439 (0.049) \\
 & AMI & 0.484 (0.027) & 0.486 (0.045) & 0.263 (0.088) & 0.478 (0.049) & 0.543 (0.068) & 0.480 (0.080) & 0.611 (0.028) \\
\midrule
Rice & ARI & 0.213 (0.000) & 0.193 (0.030) & 0.116 (0.013) & 0.071 (0.026) & 0.000 (0.000) & 0.196 (0.061) & 0.465 (0.030) \\
 & AMI & 0.272 (0.005) & 0.288 (0.024) & 0.076 (0.010) & 0.124 (0.001) & 0.000 (0.000) & 0.280 (0.026) & 0.414 (0.000) \\
\midrule
TUANDROMD & ARI & 0.103 (0.026) & 0.095 (0.006) & -0.020 (0.002) & 0.107 (0.026) & 0.179 (0.051) & 0.201 (0.109) & 0.105 (0.181) \\
 & AMI & 0.115 (0.065) & 0.243 (0.010) & 0.009 (0.004) & 0.167 (0.007) & 0.244 (0.043) & 0.282 (0.050) & 0.189 (0.090) \\
\midrule
Phoneme & ARI & 0.025 (0.013) & 0.026 (0.014) & 0.012 (0.004) & 0.008 (0.009) & 0.081 (0.002) & 0.110 (0.041) & 0.050 (0.007) \\
 & AMI & 0.093 (0.005) & 0.044 (0.002) & 0.002 (0.001) & 0.076 (0.003) & 0.125 (0.011) & 0.126 (0.007) & 0.111 (0.001) \\
\midrule
Texture & ARI & 0.309 (0.036) & 0.482 (0.082) & 0.087 (0.021) & 0.086 (0.020) & 0.307 (0.176) & 0.510 (0.061) & 0.572 (0.068) \\
 & AMI & 0.599 (0.014) & 0.630 (0.036) & 0.233 (0.042) & 0.463 (0.019) & 0.576 (0.087) & 0.701 (0.030) & 0.749 (0.030) \\
\midrule
OptDigits & ARI & 0.329 (0.069) & 0.605 (0.062) & 0.064 (0.017) & 0.113 (0.037) & 0.405 (0.126) & 0.669 (0.050) & 0.724 (0.078) \\
 & AMI & 0.605 (0.036) & 0.695 (0.028) & 0.148 (0.037) & 0.470 (0.033) & 0.485 (0.068) & 0.763 (0.022) & 0.816 (0.032) \\
\midrule
Statlog & ARI & 0.275 (0.060) & 0.346 (0.065) & 0.174 (0.089) & 0.087 (0.033) & 0.328 (0.061) & 0.434 (0.059) & 0.507 (0.057) \\
 & AMI & 0.484 (0.026) & 0.482 (0.026) & 0.250 (0.102) & 0.368 (0.021) & 0.429 (0.062) & 0.551 (0.032) & 0.604 (0.026) \\
\midrule
Anuran Calls & ARI & 0.298 (0.090) & 0.350 (0.094) & 0.137 (0.127) & 0.090 (0.092) & 0.144 (0.083) & 0.285 (0.106) & 0.265 (0.076) \\
 & AMI & 0.370 (0.040) & 0.399 (0.029) & 0.107 (0.085) & 0.204 (0.038) & 0.292 (0.056) & 0.373 (0.037) & 0.408 (0.041) \\
\midrule
Isolet & ARI & 0.295 (0.066) & 0.310 (0.039) & 0.035 (0.018) & 0.308 (0.042) & 0.000 (0.000) & 0.296 (0.071) & 0.396 (0.079) \\
 & AMI & 0.634 (0.026) & 0.606 (0.013) & 0.186 (0.057) & 0.571 (0.034) & 0.000 (0.000) & 0.681 (0.026) & 0.708 (0.034) \\
\midrule
STL10 & ARI & 0.298 (0.125) & 0.454 (0.161) & 0.000 (0.000) & 0.214 (0.057) & 0.000 (0.000) & 0.341 (0.207) & 0.497 (0.144) \\
 & AMI & 0.573 (0.092) & 0.649 (0.087) & 0.000 (0.000) & 0.413 (0.058) & 0.000 (0.000) & 0.577 (0.185) & 0.705 (0.092) \\
\midrule
MNIST10K & ARI & 0.573 (0.056) & 0.602 (0.099) & 0.000 (0.000) & 0.363 (0.088) & 0.557 (0.108) & 0.644 (0.132) & 0.736 (0.067) \\
 & AMI & 0.759 (0.018) & 0.703 (0.050) & 0.000 (0.000) & 0.598 (0.039) & 0.732 (0.047) & 0.793 (0.083) & 0.860 (0.024) \\
\midrule
PenBased & ARI & 0.383 (0.058) & 0.547 (0.045) & 0.102 (0.027) & 0.035 (0.023) & 0.449 (0.142) & 0.581 (0.045) & 0.640 (0.054) \\
 & AMI & 0.631 (0.027) & 0.650 (0.022) & 0.263 (0.044) & 0.423 (0.011) & 0.625 (0.080) & 0.718 (0.020) & 0.771 (0.019) \\
\midrule
Letter & ARI & 0.040 (0.018) & 0.107 (0.017) & 0.011 (0.007) & 0.049 (0.008) & 0.135 (0.030) & 0.108 (0.032) & 0.111 (0.023) \\
 & AMI & 0.376 (0.032) & 0.480 (0.012) & 0.063 (0.024) & 0.386 (0.043) & 0.478 (0.024) & 0.445 (0.016) & 0.365 (0.027) \\
\midrule
Shuttle & ARI & 0.266 (0.197) & 0.689 (0.048) & 0.247 (0.191) & 0.020 (0.026) & 0.270 (0.194) & 0.273 (0.107) & 0.150 (0.097) \\
 & AMI & 0.348 (0.106) & 0.591 (0.046) & 0.248 (0.192) & 0.204 (0.014) & 0.395 (0.099) & 0.345 (0.062) & 0.364 (0.042) \\
\midrule
Skin & ARI & N/A & 0.081 (0.005) & N/A & 0.016 (0.002) & 0.635 (0.036) & 0.162 (0.039) & 0.174 (0.027) \\
 & AMI & N/A & 0.242 (0.017) & N/A & 0.179 (0.007) & 0.510 (0.053) & 0.267 (0.032) & 0.328 (0.014) \\
\bottomrule
\end{tabular}%
}
\par\vspace{0.35em}
\parbox{0.98\textwidth}{\footnotesize\raggedright
The values in parentheses indicate the standard deviation over 30 independent runs. \textnormal{N/A} denotes unavailable results under the convention above.}
\end{table*}

For a generic evaluation measure $Q$, let $R_{i,j}^{(Q)}$ denote the value on stage $i$ after training up to stage $j$, where $j \ge i$. The appendix reports backward transfer summaries only as supplementary evidence, and they are defined by
\begin{equation}
  \operatorname{BWT}_{Q}
  =
  \frac{1}{J-1}\sum_{i=1}^{J-1}\left(R_{i,J}^{(Q)} - R_{i,i}^{(Q)}\right),
  \qquad
  Q \in \{\mathrm{ARI}, \mathrm{AMI}\}.
  \label{eq:phida_appendix_bwt_metric}
\end{equation}
Accordingly, the appendix reports $\operatorname{BWT}_{\mathrm{ARI}}$ and $\operatorname{BWT}_{\mathrm{AMI}}$ as supplementary nonstationary summaries only.

\begin{table*}[htbp]
\centering
\scriptsize
\setlength{\tabcolsep}{2pt}
\renewcommand{\arraystretch}{0.94}
\caption{Per-dataset nonstationary streamwise summaries.}
\label{tab:app_nonstationary_streamwise}
\resizebox{\textwidth}{!}{%
\begin{tabular}{llccccccc}
\toprule
Dataset & Measure & GNG & SOINN+ & DDVFA & CAEA & CAE & IDAT & PHIDA \\
\midrule
Iris & avgInc\_ARI & 0.576 (0.161) & 0.696 (0.070) & 0.366 (0.007) & 0.400 (0.008) & 0.440 (0.130) & 0.662 (0.074) & 0.810 (0.090) \\
 & avgInc\_AMI & 0.610 (0.160) & 0.697 (0.047) & 0.421 (0.008) & 0.442 (0.022) & 0.471 (0.140) & 0.684 (0.077) & 0.830 (0.067) \\
\midrule
Seeds & avgInc\_ARI & 0.638 (0.122) & 0.517 (0.042) & 0.477 (0.081) & 0.503 (0.051) & 0.333 (0.000) & 0.641 (0.034) & 0.588 (0.040) \\
 & avgInc\_AMI & 0.645 (0.101) & 0.588 (0.044) & 0.476 (0.057) & 0.542 (0.018) & 0.333 (0.000) & 0.669 (0.023) & 0.669 (0.020) \\
\midrule
Dermatology & avgInc\_ARI & 0.199 (0.052) & 0.324 (0.039) & 0.527 (0.107) & 0.359 (0.069) & 0.167 (0.000) & 0.246 (0.039) & 0.560 (0.100) \\
 & avgInc\_AMI & 0.230 (0.051) & 0.446 (0.045) & 0.584 (0.098) & 0.459 (0.059) & 0.167 (0.002) & 0.330 (0.051) & 0.626 (0.094) \\
\midrule
Pima & avgInc\_ARI & 0.517 (0.011) & 0.507 (0.012) & 0.499 (0.007) & 0.502 (0.000) & 0.497 (0.004) & 0.521 (0.004) & 0.531 (0.007) \\
 & avgInc\_AMI & 0.526 (0.005) & 0.508 (0.003) & 0.504 (0.007) & 0.523 (0.001) & 0.510 (0.005) & 0.531 (0.004) & 0.532 (0.008) \\
\midrule
Mice Protein & avgInc\_ARI & 0.330 (0.038) & 0.371 (0.099) & 0.259 (0.056) & 0.196 (0.014) & 0.126 (0.001) & 0.359 (0.035) & 0.374 (0.028) \\
 & avgInc\_AMI & 0.442 (0.046) & 0.532 (0.049) & 0.301 (0.055) & 0.394 (0.009) & 0.139 (0.021) & 0.479 (0.040) & 0.542 (0.031) \\
\midrule
Binalpha & avgInc\_ARI & 0.128 (0.037) & 0.327 (0.041) & 0.073 (0.005) & 0.279 (0.027) & 0.085 (0.032) & 0.180 (0.039) & 0.266 (0.045) \\
 & avgInc\_AMI & 0.291 (0.060) & 0.504 (0.027) & 0.121 (0.007) & 0.490 (0.021) & 0.155 (0.056) & 0.362 (0.052) & 0.463 (0.048) \\
\midrule
Yeast & avgInc\_ARI & 0.223 (0.069) & 0.215 (0.090) & 0.159 (0.042) & 0.107 (0.001) & 0.166 (0.032) & 0.238 (0.068) & 0.213 (0.054) \\
 & avgInc\_AMI & 0.278 (0.052) & 0.229 (0.070) & 0.180 (0.044) & 0.188 (0.015) & 0.201 (0.030) & 0.283 (0.052) & 0.275 (0.045) \\
\midrule
Semeion & avgInc\_ARI & 0.346 (0.070) & 0.395 (0.072) & 0.257 (0.010) & 0.256 (0.023) & 0.141 (0.032) & 0.348 (0.074) & 0.499 (0.067) \\
 & avgInc\_AMI & 0.470 (0.068) & 0.524 (0.038) & 0.333 (0.013) & 0.486 (0.013) & 0.195 (0.035) & 0.458 (0.079) & 0.591 (0.056) \\
\midrule
MSRA25 & avgInc\_ARI & 0.434 (0.048) & 0.380 (0.043) & 0.274 (0.023) & 0.400 (0.028) & 0.205 (0.146) & 0.406 (0.044) & 0.560 (0.029) \\
 & avgInc\_AMI & 0.611 (0.046) & 0.621 (0.024) & 0.380 (0.027) & 0.553 (0.017) & 0.320 (0.189) & 0.574 (0.040) & 0.758 (0.022) \\
\midrule
Image Seg. & avgInc\_ARI & 0.354 (0.046) & 0.456 (0.130) & 0.347 (0.076) & 0.332 (0.037) & 0.402 (0.093) & 0.429 (0.077) & 0.523 (0.056) \\
 & avgInc\_AMI & 0.516 (0.062) & 0.567 (0.072) & 0.438 (0.078) & 0.479 (0.030) & 0.560 (0.066) & 0.569 (0.073) & 0.649 (0.053) \\
\midrule
Rice & avgInc\_ARI & 0.606 (0.000) & 0.597 (0.015) & 0.558 (0.006) & 0.535 (0.013) & 0.500 (0.000) & 0.598 (0.030) & 0.732 (0.015) \\
 & avgInc\_AMI & 0.636 (0.002) & 0.644 (0.012) & 0.538 (0.005) & 0.562 (0.000) & 0.500 (0.000) & 0.640 (0.013) & 0.707 (0.000) \\
\midrule
TUANDROMD & avgInc\_ARI & 0.551 (0.013) & 0.548 (0.003) & 0.490 (0.001) & 0.553 (0.013) & 0.589 (0.026) & 0.601 (0.055) & 0.553 (0.090) \\
 & avgInc\_AMI & 0.557 (0.033) & 0.622 (0.005) & 0.504 (0.002) & 0.583 (0.003) & 0.622 (0.021) & 0.641 (0.025) & 0.594 (0.045) \\
\midrule
Phoneme & avgInc\_ARI & 0.513 (0.006) & 0.513 (0.007) & 0.506 (0.002) & 0.504 (0.005) & 0.540 (0.001) & 0.555 (0.020) & 0.525 (0.003) \\
 & avgInc\_AMI & 0.546 (0.003) & 0.522 (0.001) & 0.501 (0.001) & 0.538 (0.001) & 0.563 (0.006) & 0.563 (0.004) & 0.555 (0.001) \\
\midrule
Texture & avgInc\_ARI & 0.385 (0.050) & 0.623 (0.080) & 0.245 (0.037) & 0.156 (0.016) & 0.443 (0.123) & 0.540 (0.063) & 0.612 (0.067) \\
 & avgInc\_AMI & 0.604 (0.037) & 0.680 (0.049) & 0.339 (0.039) & 0.456 (0.016) & 0.613 (0.088) & 0.672 (0.037) & 0.741 (0.038) \\
\midrule
OptDigits & avgInc\_ARI & 0.459 (0.052) & 0.712 (0.062) & 0.253 (0.039) & 0.191 (0.032) & 0.398 (0.092) & 0.643 (0.074) & 0.667 (0.067) \\
 & avgInc\_AMI & 0.641 (0.033) & 0.718 (0.037) & 0.321 (0.034) & 0.474 (0.031) & 0.422 (0.050) & 0.720 (0.038) & 0.768 (0.032) \\
\midrule
Statlog & avgInc\_ARI & 0.383 (0.054) & 0.471 (0.083) & 0.388 (0.074) & 0.214 (0.022) & 0.453 (0.074) & 0.525 (0.059) & 0.553 (0.044) \\
 & avgInc\_AMI & 0.522 (0.039) & 0.537 (0.037) & 0.418 (0.068) & 0.414 (0.012) & 0.492 (0.067) & 0.586 (0.038) & 0.626 (0.037) \\
\midrule
Anuran Calls & avgInc\_ARI & 0.405 (0.058) & 0.433 (0.080) & 0.358 (0.050) & 0.301 (0.025) & 0.341 (0.031) & 0.409 (0.067) & 0.396 (0.034) \\
 & avgInc\_AMI & 0.460 (0.040) & 0.489 (0.038) & 0.343 (0.043) & 0.367 (0.015) & 0.426 (0.026) & 0.472 (0.039) & 0.498 (0.035) \\
\midrule
Isolet & avgInc\_ARI & 0.415 (0.042) & 0.450 (0.061) & 0.173 (0.018) & 0.423 (0.060) & 0.039 (0.000) & 0.481 (0.066) & 0.580 (0.056) \\
 & avgInc\_AMI & 0.663 (0.032) & 0.644 (0.034) & 0.320 (0.019) & 0.574 (0.046) & 0.039 (0.000) & 0.704 (0.044) & 0.768 (0.041) \\
\midrule
STL10 & avgInc\_ARI & 0.457 (0.091) & 0.618 (0.151) & 0.000 (0.000) & 0.293 (0.050) & 0.000 (0.000) & 0.465 (0.215) & 0.643 (0.130) \\
 & avgInc\_AMI & 0.622 (0.064) & 0.696 (0.109) & 0.000 (0.000) & 0.466 (0.038) & 0.000 (0.000) & 0.602 (0.182) & 0.750 (0.088) \\
\midrule
MNIST10K & avgInc\_ARI & 0.435 (0.046) & 0.724 (0.076) & 0.000 (0.000) & 0.376 (0.065) & 0.550 (0.098) & 0.660 (0.127) & 0.694 (0.057) \\
 & avgInc\_AMI & 0.666 (0.018) & 0.741 (0.050) & 0.000 (0.000) & 0.553 (0.022) & 0.681 (0.049) & 0.752 (0.075) & 0.801 (0.027) \\
\midrule
PenBased & avgInc\_ARI & 0.371 (0.048) & 0.649 (0.073) & 0.335 (0.036) & 0.124 (0.015) & 0.441 (0.132) & 0.587 (0.057) & 0.618 (0.048) \\
 & avgInc\_AMI & 0.608 (0.023) & 0.687 (0.041) & 0.450 (0.030) & 0.419 (0.012) & 0.580 (0.078) & 0.689 (0.029) & 0.736 (0.026) \\
\midrule
Letter & avgInc\_ARI & 0.137 (0.017) & 0.191 (0.041) & 0.097 (0.012) & 0.085 (0.008) & 0.176 (0.030) & 0.213 (0.035) & 0.223 (0.032) \\
 & avgInc\_AMI & 0.427 (0.023) & 0.489 (0.023) & 0.143 (0.013) & 0.382 (0.034) & 0.460 (0.025) & 0.467 (0.026) & 0.431 (0.040) \\
\midrule
Shuttle & avgInc\_ARI & 0.293 (0.101) & 0.532 (0.144) & 0.275 (0.084) & 0.152 (0.009) & 0.275 (0.112) & 0.275 (0.063) & 0.235 (0.058) \\
 & avgInc\_AMI & 0.347 (0.083) & 0.529 (0.094) & 0.271 (0.057) & 0.248 (0.036) & 0.364 (0.079) & 0.342 (0.078) & 0.359 (0.069) \\
\midrule
Skin & avgInc\_ARI & N/A & 0.541 (0.003) & N/A & 0.508 (0.001) & 0.818 (0.018) & 0.581 (0.020) & 0.587 (0.014) \\
 & avgInc\_AMI & N/A & 0.621 (0.009) & N/A & 0.589 (0.004) & 0.755 (0.026) & 0.634 (0.016) & 0.664 (0.007) \\
\bottomrule
\end{tabular}%
}
\par\vspace{0.35em}
\parbox{0.98\textwidth}{\footnotesize\raggedright
The values in parentheses indicate the standard deviation over 30 independent runs. \textnormal{N/A} follows the availability convention above.}
\end{table*}

\begin{table*}[htbp]
\centering
\scriptsize
\setlength{\tabcolsep}{2pt}
\renewcommand{\arraystretch}{0.94}
\caption{Per-dataset nonstationary backward transfer summaries.}
\label{tab:app_nonstationary_bwt}
\resizebox{\textwidth}{!}{%
\begin{tabular}{llccccccc}
\toprule
Dataset & Measure & GNG & SOINN+ & DDVFA & CAEA & CAE & IDAT & PHIDA \\
\midrule
Iris & BWT\_ARI & -0.296 (0.178) & -0.210 (0.054) & -0.477 (0.073) & -0.439 (0.049) & -0.334 (0.160) & -0.145 (0.206) & -0.159 (0.110) \\
 & BWT\_AMI & -0.224 (0.180) & -0.172 (0.031) & -0.424 (0.150) & -0.396 (0.075) & -0.313 (0.150) & -0.110 (0.163) & -0.149 (0.099) \\
\midrule
Seeds & BWT\_ARI & -0.276 (0.181) & -0.316 (0.050) & -0.478 (0.139) & -0.297 (0.066) & -0.500 (0.000) & -0.236 (0.070) & -0.267 (0.076) \\
 & BWT\_AMI & -0.219 (0.147) & -0.262 (0.054) & -0.444 (0.152) & -0.269 (0.027) & -0.500 (0.000) & -0.178 (0.044) & -0.196 (0.054) \\
\midrule
Dermatology & BWT\_ARI & -0.203 (0.052) & -0.185 (0.053) & -0.323 (0.136) & -0.104 (0.183) & -0.200 (0.001) & -0.209 (0.063) & -0.088 (0.140) \\
 & BWT\_AMI & -0.175 (0.056) & -0.088 (0.043) & -0.249 (0.183) & -0.061 (0.175) & -0.197 (0.008) & -0.181 (0.083) & -0.001 (0.128) \\
\midrule
Pima & BWT\_ARI & -0.967 (0.021) & -0.987 (0.023) & -1.002 (0.013) & -0.997 (0.001) & -1.006 (0.008) & -0.958 (0.008) & -0.938 (0.013) \\
 & BWT\_AMI & -0.947 (0.011) & -0.985 (0.006) & -0.992 (0.014) & -0.955 (0.001) & -0.980 (0.011) & -0.938 (0.009) & -0.937 (0.016) \\
\midrule
Mice Protein & BWT\_ARI & -0.153 (0.048) & -0.146 (0.049) & -0.293 (0.065) & -0.095 (0.008) & -0.141 (0.002) & -0.120 (0.038) & -0.101 (0.047) \\
 & BWT\_AMI & -0.077 (0.054) & -0.045 (0.035) & -0.303 (0.066) & 0.030 (0.017) & -0.119 (0.031) & -0.067 (0.046) & -0.061 (0.044) \\
\midrule
Binalpha & BWT\_ARI & -0.107 (0.036) & -0.120 (0.026) & -0.073 (0.005) & -0.041 (0.020) & 0.010 (0.037) & -0.088 (0.065) & -0.168 (0.066) \\
 & BWT\_AMI & -0.091 (0.064) & -0.062 (0.020) & -0.098 (0.017) & -0.019 (0.017) & 0.048 (0.057) & -0.029 (0.112) & -0.111 (0.083) \\
\midrule
Yeast & BWT\_ARI & -0.144 (0.071) & -0.201 (0.095) & -0.161 (0.048) & -0.105 (0.002) & -0.100 (0.064) & -0.130 (0.098) & -0.090 (0.073) \\
 & BWT\_AMI & -0.098 (0.060) & -0.165 (0.078) & -0.152 (0.052) & -0.056 (0.018) & -0.090 (0.060) & -0.089 (0.074) & -0.063 (0.070) \\
\midrule
Semeion & BWT\_ARI & -0.239 (0.084) & -0.111 (0.056) & -0.253 (0.013) & -0.036 (0.023) & -0.062 (0.043) & -0.055 (0.056) & -0.153 (0.091) \\
 & BWT\_AMI & -0.152 (0.090) & -0.031 (0.042) & -0.205 (0.052) & 0.021 (0.021) & -0.032 (0.046) & 0.005 (0.063) & -0.070 (0.061) \\
\midrule
MSRA25 & BWT\_ARI & -0.079 (0.041) & -0.061 (0.026) & -0.233 (0.028) & 0.055 (0.025) & -0.046 (0.033) & -0.040 (0.050) & 0.003 (0.036) \\
 & BWT\_AMI & 0.000 (0.038) & 0.021 (0.015) & -0.232 (0.045) & 0.142 (0.017) & 0.010 (0.034) & 0.020 (0.048) & 0.040 (0.024) \\
\midrule
Image Seg. & BWT\_ARI & -0.118 (0.059) & -0.189 (0.080) & -0.287 (0.104) & -0.097 (0.041) & -0.108 (0.100) & -0.174 (0.080) & -0.098 (0.081) \\
 & BWT\_AMI & -0.037 (0.079) & -0.094 (0.064) & -0.204 (0.148) & -0.002 (0.045) & -0.020 (0.070) & -0.104 (0.068) & -0.044 (0.067) \\
\midrule
Rice & BWT\_ARI & -0.787 (0.000) & -0.807 (0.030) & -0.884 (0.013) & -0.930 (0.026) & -1.000 (0.000) & -0.804 (0.061) & -0.535 (0.030) \\
 & BWT\_AMI & -0.728 (0.005) & -0.712 (0.024) & -0.924 (0.010) & -0.876 (0.001) & -1.000 (0.000) & -0.720 (0.026) & -0.586 (0.000) \\
\midrule
TUANDROMD & BWT\_ARI & -0.897 (0.026) & -0.905 (0.006) & -1.020 (0.002) & -0.894 (0.026) & -0.822 (0.051) & -0.799 (0.109) & -0.895 (0.181) \\
 & BWT\_AMI & -0.885 (0.065) & -0.757 (0.010) & -0.991 (0.004) & -0.833 (0.007) & -0.756 (0.043) & -0.718 (0.050) & -0.811 (0.090) \\
\midrule
Phoneme & BWT\_ARI & -0.975 (0.013) & -0.974 (0.014) & -0.988 (0.004) & -0.992 (0.009) & -0.919 (0.002) & -0.890 (0.041) & -0.950 (0.007) \\
 & BWT\_AMI & -0.907 (0.005) & -0.956 (0.002) & -0.998 (0.001) & -0.924 (0.003) & -0.875 (0.011) & -0.874 (0.007) & -0.889 (0.001) \\
\midrule
Texture & BWT\_ARI & -0.084 (0.047) & -0.155 (0.080) & -0.174 (0.048) & -0.078 (0.014) & -0.150 (0.219) & -0.033 (0.056) & -0.043 (0.069) \\
 & BWT\_AMI & -0.006 (0.042) & -0.055 (0.051) & -0.117 (0.066) & 0.008 (0.010) & -0.040 (0.130) & 0.032 (0.037) & 0.008 (0.040) \\
\midrule
OptDigits & BWT\_ARI & -0.145 (0.093) & -0.118 (0.035) & -0.209 (0.050) & -0.086 (0.014) & 0.008 (0.084) & 0.029 (0.052) & 0.063 (0.055) \\
 & BWT\_AMI & -0.040 (0.053) & -0.026 (0.024) & -0.192 (0.062) & -0.004 (0.032) & 0.070 (0.063) & 0.047 (0.031) & 0.054 (0.025) \\
\midrule
Statlog & BWT\_ARI & -0.130 (0.061) & -0.150 (0.055) & -0.257 (0.165) & -0.153 (0.028) & -0.150 (0.074) & -0.110 (0.060) & -0.055 (0.072) \\
 & BWT\_AMI & -0.046 (0.045) & -0.065 (0.040) & -0.202 (0.165) & -0.055 (0.023) & -0.076 (0.059) & -0.042 (0.051) & -0.027 (0.051) \\
\midrule
Anuran Calls & BWT\_ARI & -0.143 (0.100) & -0.110 (0.063) & -0.294 (0.164) & -0.282 (0.111) & -0.263 (0.088) & -0.164 (0.087) & -0.175 (0.094) \\
 & BWT\_AMI & -0.120 (0.048) & -0.119 (0.042) & -0.315 (0.128) & -0.217 (0.055) & -0.179 (0.064) & -0.132 (0.048) & -0.120 (0.068) \\
\midrule
Isolet & BWT\_ARI & -0.125 (0.073) & -0.145 (0.049) & -0.144 (0.028) & -0.119 (0.056) & -0.040 (0.000) & -0.193 (0.082) & -0.192 (0.083) \\
 & BWT\_AMI & -0.030 (0.042) & -0.040 (0.031) & -0.139 (0.062) & -0.003 (0.029) & -0.040 (0.000) & -0.024 (0.044) & -0.063 (0.053) \\
\midrule
STL10 & BWT\_ARI & -0.176 (0.133) & -0.183 (0.092) & -1.000 (0.000) & -0.088 (0.026) & -1.000 (0.000) & -0.138 (0.073) & -0.162 (0.121) \\
 & BWT\_AMI & -0.054 (0.102) & -0.053 (0.064) & -1.000 (0.000) & -0.060 (0.032) & -1.000 (0.000) & -0.028 (0.072) & -0.050 (0.080) \\
\midrule
MNIST10K & BWT\_ARI & 0.153 (0.055) & -0.136 (0.061) & -1.000 (0.000) & -0.014 (0.082) & 0.008 (0.083) & -0.018 (0.085) & 0.047 (0.067) \\
 & BWT\_AMI & 0.103 (0.017) & -0.042 (0.035) & -1.000 (0.000) & 0.050 (0.034) & 0.057 (0.035) & 0.046 (0.056) & 0.065 (0.026) \\
\midrule
PenBased & BWT\_ARI & 0.013 (0.060) & -0.114 (0.046) & -0.258 (0.051) & -0.100 (0.011) & 0.009 (0.072) & -0.007 (0.049) & 0.025 (0.058) \\
 & BWT\_AMI & 0.026 (0.032) & -0.042 (0.037) & -0.208 (0.060) & 0.005 (0.009) & 0.050 (0.042) & 0.032 (0.033) & 0.039 (0.033) \\
\midrule
Letter & BWT\_ARI & -0.101 (0.021) & -0.088 (0.030) & -0.089 (0.015) & -0.038 (0.003) & -0.042 (0.029) & -0.109 (0.035) & -0.117 (0.042) \\
 & BWT\_AMI & -0.052 (0.030) & -0.009 (0.019) & -0.083 (0.027) & 0.005 (0.013) & 0.019 (0.017) & -0.022 (0.023) & -0.069 (0.048) \\
\midrule
Shuttle & BWT\_ARI & -0.031 (0.179) & 0.183 (0.157) & -0.032 (0.246) & -0.154 (0.023) & -0.006 (0.127) & -0.002 (0.088) & -0.100 (0.084) \\
 & BWT\_AMI & 0.001 (0.100) & 0.072 (0.092) & -0.026 (0.256) & -0.052 (0.037) & 0.037 (0.073) & 0.004 (0.071) & 0.006 (0.078) \\
\midrule
Skin & BWT\_ARI & N/A & -0.919 (0.005) & N/A & -0.984 (0.002) & -0.365 (0.036) & -0.838 (0.039) & -0.826 (0.027) \\
 & BWT\_AMI & N/A & -0.758 (0.017) & N/A & -0.821 (0.007) & -0.490 (0.053) & -0.733 (0.032) & -0.672 (0.014) \\
\bottomrule
\end{tabular}%
}
\par\vspace{0.35em}
\parbox{0.98\textwidth}{\footnotesize\raggedright
The values in parentheses indicate the standard deviation over 30 independent runs. \textnormal{N/A} follows the availability convention above.}
\end{table*}

\subsection{Auxiliary Summaries of Learned Nodes, Output Granularity, and Runtime}
\label{app:structural_tables}

This subsection reports auxiliary summaries that complement the ARI and AMI values in Appendix~\ref{app:benchmark_tables}. Tables~\ref{tab:app_nodes_clusters_stationary} and \ref{tab:app_nodes_clusters_nonstationary} report counts of learned nodes and the number of clusters produced at the final evaluation point. Tables~\ref{tab:app_fit_time_stationary} and \ref{tab:app_fit_time_nonstationary} report average computation time under the common execution environment. Counts of learned nodes summarize the size of the state of learned nodes, and cluster counts summarize output granularity. Computation time summarizes implementation cost rather than an additional clustering performance objective. PHIDA may require higher computation time than IDAT on some datasets because it periodically rebuilds the PH graph and updates the PH-constrained component hierarchy. The reported times should therefore be read together with the complexity characterization in Section~\ref{sec:complexity} and the ARI and AMI values in Appendix~\ref{app:benchmark_tables}.

\begin{table*}[htbp]
\centering
\scriptsize
\setlength{\tabcolsep}{3pt}
\renewcommand{\arraystretch}{0.95}
\caption{Average number of nodes and final clusters in the stationary setting.}
\label{tab:app_nodes_clusters_stationary}
\resizebox{\textwidth}{!}{%
\begin{tabular}{llccccccccccccc}
\toprule
Dataset & Measure & GNG & SOINN+ & DDVFA & CAEA & TBM-P & CAE & IDAT & TC & SR-PCM-HDP & TableDC & G-CEALS & AuToMATo & PHIDA \\
\midrule
Iris & \#Nodes & 50.8 (0.9) & 24.9 (7.0) & -- & 40.4 (2.7) & 3.2 (0.8) & 62.7 (28.3) & 8.8 (1.5) & -- & -- & -- & -- & -- & 11.4 (2.7) \\
(\#Classes: 3) & \#Clusters & 4.6 (2.2) & 9.3 (3.8) & 2.2 (0.5) & 25.7 (3.3) & 2.2 (0.4) & 48.1 (28.7) & 5.6 (1.4) & 2.0 (0.0) & 5.0 (0.2) & 3.0 (0.0) & 3.0 (0.0) & 2.0 (0.0) & 3.8 (0.9) \\
\midrule
Seeds & \#Nodes & 32.0 (0.0) & 29.2 (9.1) & -- & 54.5 (5.1) & 4.6 (1.0) & 187.4 (14.7) & 12.1 (2.4) & -- & -- & -- & -- & -- & 15.9 (3.1) \\
(\#Classes: 3) & \#Clusters & 1.0 (0.2) & 7.9 (5.3) & 13.7 (1.4) & 38.8 (2.1) & 3.7 (0.9) & 179.2 (19.8) & 6.8 (1.9) & 7.0 (0.0) & 3.0 (0.2) & 3.0 (0.0) & 3.0 (0.0) & 3.0 (0.0) & 4.7 (1.0) \\
\midrule
Dermatology & \#Nodes & 54.0 (0.0) & 46.8 (14.3) & -- & 4.7 (4.7) & 8.8 (0.4) & 285.7 (79.9) & 12.7 (2.9) & -- & -- & -- & -- & -- & 21.0 (4.4) \\
(\#Classes: 6) & \#Clusters & 1.0 (0.2) & 16.9 (7.1) & 13.7 (1.8) & 1.2 (0.6) & 8.8 (0.4) & 253.7 (90.5) & 6.1 (2.0) & 10.0 (0.0) & 7.0 (0.0) & 6.0 (0.0) & 6.0 (0.0) & 1.0 (0.0) & 5.6 (1.4) \\
\midrule
Pima & \#Nodes & 110.8 (0.4) & 53.9 (11.7) & -- & 312.8 (9.0) & 7.5 (0.6) & 300.6 (160.0) & 13.1 (3.4) & -- & -- & -- & -- & -- & 14.3 (2.2) \\
(\#Classes: 2) & \#Clusters & 1.2 (0.4) & 11.6 (4.1) & 112.8 (5.6) & 234.9 (9.1) & 7.5 (0.6) & 218.1 (161.8) & 6.1 (1.7) & 12.0 (0.0) & 5.0 (0.0) & 2.0 (0.0) & 2.0 (0.0) & 2.0 (0.0) & 3.3 (1.0) \\
\midrule
Mice Protein & \#Nodes & 35.6 (0.6) & 76.7 (18.6) & -- & 66.0 (0.0) & 113.3 (3.6) & 876.4 (282.1) & 21.4 (3.9) & -- & -- & -- & -- & -- & 34.5 (8.5) \\
(\#Classes: 8) & \#Clusters & 1.6 (0.8) & 15.5 (7.5) & 374.8 (4.9) & 66.0 (0.0) & 105.0 (4.6) & 843.8 (313.9) & 9.3 (2.1) & 15.0 (0.0) & 5.4 (0.9) & 8.0 (0.0) & 8.0 (0.0) & 2.0 (0.0) & 8.8 (2.0) \\
\midrule
Binalpha & \#Nodes & 235.9 (0.3) & 77.6 (24.8) & -- & 63.2 (8.3) & 165.9 (3.9) & 312.6 (159.9) & 27.5 (14.9) & -- & -- & -- & -- & -- & 28.2 (16.1) \\
(\#Classes: 36) & \#Clusters & 1.0 (0.0) & 39.9 (13.4) & 224.3 (11.5) & 16.7 (4.4) & 132.9 (6.2) & 213.0 (135.2) & 9.6 (5.9) & 327.0 (0.0) & 1.0 (0.0) & 36.0 (0.0) & 36.0 (0.0) & 4.1 (0.3) & 5.0 (3.1) \\
\midrule
Yeast & \#Nodes & 484.0 (4.0) & 61.4 (16.0) & -- & 18.8 (3.5) & 47.7 (2.7) & 230.9 (83.4) & 32.3 (4.4) & -- & -- & -- & -- & -- & 46.4 (6.4) \\
(\#Classes: 10) & \#Clusters & 21.3 (4.7) & 5.6 (2.7) & 27.7 (2.6) & 1.9 (1.0) & 34.6 (4.3) & 100.8 (65.4) & 9.2 (2.3) & 2.0 (0.0) & 3.0 (0.0) & 10.0 (0.0) & 10.0 (0.0) & 3.0 (0.0) & 10.7 (2.8) \\
\midrule
Semeion & \#Nodes & 160.2 (0.4) & 85.2 (22.2) & -- & 117.0 (0.0) & 157.4 (3.6) & 652.7 (8.7) & 39.2 (21.2) & -- & -- & -- & -- & -- & 43.6 (26.8) \\
(\#Classes: 10) & \#Clusters & 1.0 (0.2) & 44.8 (12.7) & 1427.3 (4.8) & 117.0 (0.0) & 138.7 (6.3) & 498.2 (10.6) & 18.7 (11.7) & 347.0 (0.0) & 1.0 (0.0) & 10.0 (0.0) & 10.0 (0.0) & 4.0 (0.2) & 6.3 (2.8) \\
\midrule
MSRA25 & \#Nodes & 86.8 (0.4) & 140.1 (27.6) & -- & 8.2 (2.8) & 82.1 (4.1) & 484.9 (202.1) & 39.4 (10.7) & -- & -- & -- & -- & -- & 54.1 (8.6) \\
(\#Classes: 12) & \#Clusters & 4.1 (1.3) & 43.7 (10.5) & 66.5 (4.4) & 2.1 (1.4) & 69.1 (5.7) & 327.5 (176.6) & 23.4 (7.1) & 24.0 (0.0) & 2.0 (0.0) & 12.0 (0.0) & 12.0 (0.0) & 9.0 (1.6) & 14.2 (3.0) \\
\midrule
Image Seg. & \#Nodes & 331.1 (0.8) & 94.3 (31.2) & -- & 51.0 (0.0) & 85.8 (4.8) & 194.0 (62.2) & 25.5 (5.5) & -- & -- & -- & -- & -- & 19.6 (2.6) \\
(\#Classes: 7) & \#Clusters & 1.5 (0.5) & 9.2 (6.0) & 14.4 (2.5) & 51.0 (0.0) & 25.1 (4.4) & 81.8 (48.3) & 10.2 (2.3) & 2.0 (0.0) & 11.0 (0.8) & 7.0 (0.0) & 7.0 (0.0) & 4.0 (0.2) & 5.9 (1.1) \\
\midrule
Rice & \#Nodes & 11.8 (5.0) & 129.0 (38.5) & -- & 92.0 (0.0) & 88.8 (6.8) & 1315.1 (424.1) & 17.7 (2.8) & -- & -- & -- & -- & -- & 12.9 (1.3) \\
(\#Classes: 2) & \#Clusters & 4.1 (2.0) & 15.9 (5.0) & 44.2 (2.8) & 92.0 (0.0) & 64.9 (5.4) & 981.9 (498.5) & 5.7 (2.4) & 25.0 (0.0) & 2.0 (0.0) & 2.0 (0.0) & 2.0 (0.0) & 1.0 (0.0) & 3.2 (1.1) \\
\midrule
TUANDROMD & \#Nodes & 121.9 (0.3) & 78.0 (22.4) & -- & 163.3 (3.5) & 97.6 (6.6) & 73.0 (75.0) & 7.1 (1.5) & -- & -- & -- & -- & -- & 12.2 (1.6) \\
(\#Classes: 2) & \#Clusters & 4.4 (0.9) & 30.5 (12.9) & 16.0 (2.3) & 131.7 (5.0) & 44.3 (5.5) & 6.9 (2.1) & 3.2 (1.4) & 5.0 (0.0) & 1.0 (0.0) & 2.0 (0.0) & 2.0 (0.0) & 360.9 (84.8) & 3.2 (0.8) \\
\midrule
Phoneme & \#Nodes & 2000.0 (0.0) & 150.3 (34.3) & -- & 26.9 (6.7) & 427.5 (11.0) & 245.3 (48.1) & 34.7 (5.8) & -- & -- & -- & -- & -- & 42.8 (4.4) \\
(\#Classes: 2) & \#Clusters & 19.7 (4.8) & 10.2 (4.7) & 120.0 (0.0) & 1.1 (0.3) & 134.8 (12.5) & 78.8 (38.6) & 10.7 (2.9) & 37.0 (0.0) & 6.0 (0.0) & 2.0 (0.0) & 2.0 (0.0) & 3.7 (0.7) & 10.4 (2.0) \\
\midrule
Texture & \#Nodes & 28.7 (4.9) & 174.9 (55.7) & -- & 30.4 (7.7) & 93.0 (4.9) & 177.1 (54.2) & 43.9 (5.3) & -- & -- & -- & -- & -- & 56.2 (4.7) \\
(\#Classes: 11) & \#Clusters & 10.2 (1.8) & 19.8 (9.2) & 503.5 (3.2) & 3.5 (1.5) & 76.1 (4.9) & 63.5 (38.7) & 15.8 (2.8) & 11.0 (0.0) & 12.1 (0.6) & 11.0 (0.0) & 11.0 (0.0) & 3.1 (0.3) & 9.0 (3.2) \\
\midrule
OptDigits & \#Nodes & 1833.2 (5.0) & 228.8 (49.0) & -- & 15.0 (0.0) & 201.1 (4.5) & 525.8 (177.8) & 70.2 (12.1) & -- & -- & -- & -- & -- & 90.5 (14.0) \\
(\#Classes: 10) & \#Clusters & 100.4 (11.7) & 57.8 (18.0) & 9.4 (2.9) & 15.0 (0.0) & 146.7 (7.1) & 309.1 (148.5) & 27.0 (3.6) & 10.0 (0.0) & 1.0 (0.0) & 10.0 (0.0) & 10.0 (0.0) & 7.1 (0.3) & 13.1 (2.0) \\
\midrule
Statlog & \#Nodes & 19.8 (8.0) & 124.7 (42.1) & -- & 4.6 (2.6) & 133.3 (3.2) & 165.1 (81.2) & 36.3 (4.7) & -- & -- & -- & -- & -- & 70.1 (5.2) \\
(\#Classes: 6) & \#Clusters & 2.9 (1.5) & 34.7 (14.8) & 1310.6 (7.3) & 1.0 (0.2) & 110.2 (6.2) & 96.4 (63.4) & 16.8 (3.3) & 22.0 (0.0) & 7.0 (0.0) & 6.0 (0.0) & 6.0 (0.0) & 4.0 (0.0) & 7.6 (2.1) \\
\midrule
Anuran Calls & \#Nodes & 39.7 (8.8) & 218.7 (63.2) & -- & 64.7 (5.3) & 120.8 (5.2) & 254.6 (118.0) & 26.1 (5.5) & -- & -- & -- & -- & -- & 43.9 (3.1) \\
(\#Classes: 4) & \#Clusters & 6.1 (2.0) & 60.7 (20.0) & 431.9 (8.1) & 19.8 (3.4) & 79.1 (5.9) & 126.7 (85.6) & 12.9 (2.2) & 5.0 (0.0) & 5.8 (0.4) & 4.0 (0.0) & 4.0 (0.0) & 12.4 (0.5) & 10.3 (1.6) \\
\midrule
Isolet & \#Nodes & 41.6 (8.8) & 197.9 (38.6) & -- & 85.0 (0.0) & 282.0 (0.2) & 1558.9 (489.1) & 71.5 (17.3) & -- & -- & -- & -- & N/A & 98.5 (25.9) \\
(\#Classes: 26) & \#Clusters & 15.2 (3.3) & 102.1 (22.7) & 2000.0 (0.0) & 85.0 (0.0) & 282.0 (0.2) & 1187.1 (664.5) & 28.0 (6.1) & 51.0 (0.0) & 1.0 (0.0) & 26.0 (0.0) & 26.0 (0.0) & N/A & 12.1 (2.0) \\
\midrule
STL10 & \#Nodes & 33.4 (5.1) & 94.5 (38.1) & -- & 28.0 (0.0) & 180.0 (0.0) & 2000.0 (0.0) & 40.2 (6.2) & -- & -- & -- & -- & N/A & 65.4 (14.9) \\
(\#Classes: 10) & \#Clusters & 10.4 (1.9) & 53.5 (23.9) & 1998.9 (1.2) & 28.0 (0.0) & 180.0 (0.0) & 2000.0 (0.0) & 12.9 (2.5) & 2.0 (0.0) & 1.0 (0.0) & 10.0 (0.0) & 10.0 (0.0) & N/A & 10.0 (2.4) \\
\midrule
MNIST10K & \#Nodes & 2000.0 (0.0) & 197.5 (55.4) & -- & 17.0 (0.0) & 56.4 (4.2) & 298.9 (22.6) & 70.2 (10.5) & -- & -- & -- & -- & N/A & 100.1 (9.4) \\
(\#Classes: 10) & \#Clusters & 36.1 (4.7) & 30.1 (9.9) & 8.0 (0.9) & 17.0 (0.0) & 27.3 (3.9) & 27.9 (20.9) & 37.8 (5.4) & 79.0 (0.0) & 10.9 (0.7) & 10.0 (0.0) & 10.0 (0.0) & N/A & 11.5 (1.9) \\
\midrule
PenBased & \#Nodes & 153.4 (10.2) & 274.8 (53.1) & -- & 24.0 (0.0) & 300.1 (12.2) & 371.5 (96.0) & 39.6 (5.3) & -- & -- & -- & -- & -- & 56.5 (2.3) \\
(\#Classes: 10) & \#Clusters & 20.6 (3.4) & 29.1 (6.4) & 1286.7 (28.0) & 24.0 (0.0) & 144.8 (9.3) & 124.6 (73.5) & 16.3 (2.6) & 12.0 (0.0) & 6.9 (0.8) & 10.0 (0.0) & 10.0 (0.0) & 13.0 (0.0) & 11.5 (2.0) \\
\midrule
Letter & \#Nodes & 208.0 (14.1) & 260.2 (55.2) & -- & 224.4 (18.4) & 1964.4 (22.2) & 657.1 (238.4) & 72.6 (8.0) & -- & -- & -- & -- & -- & 65.1 (4.7) \\
(\#Classes: 26) & \#Clusters & 15.8 (3.8) & 12.1 (5.1) & 2000.0 (0.2) & 39.6 (11.9) & 1543.8 (27.5) & 289.2 (190.9) & 8.9 (2.2) & 13020.0 (0.0) & 1.7 (0.5) & 26.0 (0.0) & 26.0 (0.0) & 68.7 (2.2) & 8.8 (2.3) \\
\midrule
Shuttle & \#Nodes & 2000.0 (0.0) & 329.3 (66.0) & -- & 208.0 (0.0) & 65.1 (3.8) & 468.6 (186.7) & 43.6 (16.1) & -- & -- & -- & -- & -- & 61.0 (5.3) \\
(\#Classes: 7) & \#Clusters & 12.2 (3.8) & 11.1 (3.1) & 34.8 (1.8) & 208.0 (0.0) & 9.1 (1.9) & 30.4 (27.8) & 7.2 (2.6) & 2.0 (0.0) & 2.3 (0.6) & 7.0 (0.0) & 7.0 (0.0) & 4.0 (0.0) & 10.0 (1.7) \\
\midrule
Skin & \#Nodes & 1999.7 (0.5) & 940.0 (193.7) & N/A & 346.8 (66.1) & N/A & 261.2 (55.7) & 25.9 (3.8) & N/A & -- & -- & -- & -- & 69.2 (3.9) \\
(\#Classes: 2) & \#Clusters & 225.8 (15.0) & 109.6 (26.8) & N/A & 157.5 (29.8) & N/A & 14.1 (5.4) & 9.1 (2.5) & N/A & 5.6 (0.6) & 2.0 (0.0) & 2.0 (0.0) & 14769.0 (575.2) & 7.9 (1.8) \\
\bottomrule
\end{tabular}%
}
\par\vspace{0.35em}
\parbox{0.98\textwidth}{\footnotesize\raggedright
The values in parentheses indicate the standard deviation over 30 independent runs. \textnormal{N/A} follows the availability convention above. For available methods that do not explicitly maintain learned nodes or prototypes, \#Nodes is not applicable and is shown as \texttt{--}. Anuran Calls uses only the Family label.}
\end{table*}

\begin{table*}[htbp]
\centering
\scriptsize
\setlength{\tabcolsep}{3pt}
\renewcommand{\arraystretch}{0.95}
\caption{Average number of nodes and final clusters in the nonstationary setting.}
\label{tab:app_nodes_clusters_nonstationary}
\resizebox{\textwidth}{!}{%
\begin{tabular}{llccccccc}
\toprule
Dataset & Measure & GNG & SOINN+ & DDVFA & CAEA & CAE & IDAT & PHIDA \\
\midrule
Iris & \#Nodes & 48.6 (1.0) & 39.1 (6.0) & -- & 13.0 (0.0) & 92.7 (45.5) & 7.7 (1.2) & 16.1 (5.9) \\
(\#Classes: 3) & \#Clusters & 4.1 (2.2) & 18.5 (3.9) & 1.3 (0.5) & 13.0 (0.0) & 85.9 (49.7) & 4.1 (1.1) & 4.9 (1.1) \\
\midrule
Seeds & \#Nodes & 69.5 (1.0) & 59.5 (3.7) & -- & 43.0 (0.0) & 210.0 (0.0) & 12.7 (2.1) & 25.0 (3.6) \\
(\#Classes: 3) & \#Clusters & 5.8 (2.2) & 22.5 (5.3) & 3.4 (0.9) & 43.0 (0.0) & 210.0 (0.0) & 6.0 (1.5) & 7.8 (2.0) \\
\midrule
Dermatology & \#Nodes & 120.5 (1.1) & 110.5 (16.9) & -- & 37.0 (0.0) & 364.7 (3.3) & 14.1 (3.7) & 28.5 (5.5) \\
(\#Classes: 6) & \#Clusters & 9.7 (3.2) & 55.9 (13.1) & 1.4 (0.5) & 37.0 (0.0) & 363.8 (5.8) & 7.1 (3.1) & 8.3 (2.3) \\
\midrule
Pima & \#Nodes & 250.4 (3.1) & 69.4 (17.9) & -- & 129.5 (19.3) & 289.2 (66.3) & 14.0 (5.0) & 18.2 (2.5) \\
(\#Classes: 2) & \#Clusters & 15.0 (4.0) & 15.6 (7.0) & 1.8 (1.3) & 70.5 (21.1) & 203.8 (90.7) & 7.4 (3.0) & 4.6 (0.5) \\
\midrule
Mice Protein & \#Nodes & 350.2 (2.8) & 344.4 (27.7) & -- & 202.0 (0.0) & 1008.8 (91.8) & 31.0 (6.5) & 100.3 (8.0) \\
(\#Classes: 8) & \#Clusters & 36.1 (5.8) & 167.3 (20.7) & 1.0 (0.0) & 202.0 (0.0) & 1004.8 (97.0) & 12.8 (2.5) & 14.8 (3.9) \\
\midrule
Binalpha & \#Nodes & 457.1 (3.5) & 364.7 (28.3) & -- & 65.0 (0.0) & 1120.7 (81.2) & 42.8 (7.3) & 80.7 (18.8) \\
(\#Classes: 36) & \#Clusters & 21.2 (4.5) & 226.0 (19.7) & 27.8 (10.7) & 65.0 (0.0) & 1050.7 (108.3) & 18.7 (6.5) & 10.8 (3.5) \\
\midrule
Yeast & \#Nodes & 480.0 (3.6) & 171.1 (57.9) & -- & 235.0 (0.0) & 335.4 (216.2) & 31.8 (6.4) & 59.8 (13.5) \\
(\#Classes: 10) & \#Clusters & 22.1 (8.1) & 36.1 (20.0) & 2.0 (1.0) & 235.0 (0.0) & 176.2 (185.9) & 8.2 (3.0) & 8.8 (2.4) \\
\midrule
Semeion & \#Nodes & 517.4 (3.3) & 344.3 (40.4) & -- & 87.0 (0.0) & 1272.7 (116.5) & 44.2 (5.6) & 115.2 (16.8) \\
(\#Classes: 10) & \#Clusters & 26.2 (6.3) & 180.2 (30.8) & 77.6 (38.0) & 87.0 (0.0) & 1197.0 (150.0) & 25.2 (5.9) & 12.6 (2.6) \\
\midrule
MSRA25 & \#Nodes & 574.1 (5.0) & 543.2 (42.3) & -- & 202.0 (0.0) & 1039.6 (505.1) & 41.0 (5.3) & 139.7 (15.6) \\
(\#Classes: 12) & \#Clusters & 59.2 (7.8) & 255.7 (29.0) & 1.7 (0.6) & 202.0 (0.0) & 950.2 (518.5) & 19.0 (3.4) & 24.6 (3.4) \\
\midrule
Image Seg. & \#Nodes & 747.7 (5.7) & 608.5 (54.2) & -- & 202.0 (0.0) & 204.6 (64.3) & 36.7 (9.3) & 52.4 (7.6) \\
(\#Classes: 7) & \#Clusters & 58.2 (11.4) & 223.7 (40.2) & 1.0 (0.0) & 202.0 (0.0) & 63.0 (43.3) & 14.5 (3.6) & 12.2 (2.6) \\
\midrule
Rice & \#Nodes & 771.3 (0.4) & 216.3 (22.5) & -- & 57.2 (66.1) & 2000.0 (0.0) & 23.4 (0.5) & 17.9 (1.5) \\
(\#Classes: 2) & \#Clusters & 6.7 (0.4) & 20.7 (6.3) & 1.0 (0.0) & 57.2 (66.1) & 2000.0 (0.0) & 7.2 (3.5) & 3.0 (0.0) \\
\midrule
TUANDROMD & \#Nodes & 213.8 (284.7) & 147.3 (33.3) & -- & 16.9 (4.9) & 88.5 (93.6) & 9.4 (3.0) & 22.3 (4.9) \\
(\#Classes: 2) & \#Clusters & 24.9 (36.9) & 66.6 (24.3) & 2.3 (0.4) & 16.9 (4.9) & 19.3 (6.3) & 5.2 (1.5) & 3.6 (0.5) \\
\midrule
Phoneme & \#Nodes & 1757.0 (7.5) & 271.0 (84.7) & -- & 207.1 (72.4) & 199.0 (34.9) & 35.6 (3.0) & 46.3 (4.4) \\
(\#Classes: 2) & \#Clusters & 97.9 (10.2) & 30.8 (15.9) & 3.5 (0.9) & 207.1 (72.4) & 58.6 (39.4) & 12.4 (0.5) & 11.5 (2.0) \\
\midrule
Texture & \#Nodes & 1773.3 (7.5) & 1336.6 (76.6) & -- & 350.1 (72.9) & 374.9 (144.1) & 128.6 (16.3) & 195.6 (25.7) \\
(\#Classes: 11) & \#Clusters & 126.3 (12.4) & 451.8 (53.2) & 1.0 (0.0) & 99.2 (35.2) & 82.6 (84.9) & 34.3 (4.7) & 17.1 (5.0) \\
\midrule
OptDigits & \#Nodes & 1833.5 (5.1) & 1180.7 (81.0) & -- & 155.4 (53.9) & 1652.9 (648.9) & 129.3 (15.7) & 225.2 (31.8) \\
(\#Classes: 10) & \#Clusters & 100.4 (12.6) & 385.8 (55.4) & 1.0 (0.0) & 30.1 (10.3) & 1183.7 (542.6) & 31.2 (5.9) & 22.5 (7.7) \\
\midrule
Statlog & \#Nodes & 2000.0 (0.0) & 639.3 (123.4) & -- & 285.3 (50.9) & 541.1 (565.6) & 45.8 (8.5) & 119.2 (15.3) \\
(\#Classes: 6) & \#Clusters & 91.2 (11.7) & 252.4 (53.2) & 12.8 (19.9) & 146.2 (47.1) & 382.0 (480.1) & 17.0 (3.8) & 10.3 (2.2) \\
\midrule
Anuran Calls & \#Nodes & 2000.0 (0.0) & 546.7 (131.8) & -- & 174.8 (29.8) & 297.7 (125.9) & 27.5 (7.2) & 77.0 (13.2) \\
(\#Classes: 4) & \#Clusters & 243.0 (27.9) & 188.1 (54.9) & 2.8 (1.6) & 174.8 (29.8) & 170.1 (89.3) & 11.1 (4.4) & 11.0 (2.3) \\
\midrule
Isolet & \#Nodes & 2000.0 (0.0) & 1119.5 (70.1) & -- & 202.0 (0.0) & 2000.0 (0.0) & 76.2 (8.9) & 311.0 (121.6) \\
(\#Classes: 26) & \#Clusters & 132.0 (14.6) & 556.5 (54.6) & 3.9 (6.8) & 202.0 (0.0) & 2000.0 (0.0) & 24.4 (3.4) & 21.9 (6.4) \\
\midrule
STL10 & \#Nodes & 2000.0 (0.2) & 359.9 (63.5) & -- & 244.1 (8.0) & 0.0 (0.0) & 37.7 (8.4) & 121.0 (34.4) \\
(\#Classes: 10) & \#Clusters & 72.1 (11.5) & 166.5 (29.7) & 0.0 (0.0) & 174.5 (6.7) & 0.0 (0.0) & 10.1 (3.9) & 12.7 (3.9) \\
\midrule
MNIST10K & \#Nodes & 2000.0 (0.0) & 1773.1 (288.1) & -- & 1200.7 (448.8) & 1564.8 (461.7) & 124.5 (32.8) & 250.5 (24.9) \\
(\#Classes: 10) & \#Clusters & 123.9 (11.8) & 568.1 (136.8) & 0.0 (0.0) & 144.8 (59.7) & 146.2 (51.9) & 32.3 (7.6) & 17.8 (4.3) \\
\midrule
PenBased & \#Nodes & 2000.0 (0.0) & 2420.3 (197.0) & -- & 936.7 (235.8) & 898.6 (509.5) & 132.1 (29.2) & 193.4 (16.2) \\
(\#Classes: 10) & \#Clusters & 156.2 (13.6) & 798.4 (143.0) & 2.1 (1.3) & 188.1 (36.3) & 276.4 (371.1) & 32.5 (5.8) & 21.3 (4.1) \\
\midrule
Letter & \#Nodes & 1999.9 (0.3) & 6251.3 (353.9) & -- & 567.0 (193.1) & 882.5 (314.4) & 189.0 (20.4) & 264.1 (50.1) \\
(\#Classes: 26) & \#Clusters & 174.5 (18.6) & 2561.1 (306.4) & 1.4 (0.7) & 166.2 (41.1) & 228.5 (108.4) & 51.0 (9.0) & 19.4 (4.9) \\
\midrule
Shuttle & \#Nodes & 2000.0 (0.0) & 1683.6 (1064.1) & -- & 389.2 (70.7) & 503.1 (191.0) & 131.6 (46.5) & 176.6 (26.5) \\
(\#Classes: 7) & \#Clusters & 79.2 (25.4) & 110.0 (84.8) & 2.5 (1.8) & 226.1 (83.4) & 55.9 (22.8) & 20.3 (7.5) & 15.8 (5.0) \\
\midrule
Skin & \#Nodes & N/A & 1187.6 (74.2) & N/A & 104.6 (36.4) & 407.4 (130.0) & 42.2 (11.5) & 115.7 (4.4) \\
(\#Classes: 2) & \#Clusters & N/A & 111.4 (27.9) & N/A & 3.6 (2.7) & 21.0 (7.5) & 17.0 (2.5) & 12.5 (2.0) \\
\bottomrule
\end{tabular}%
}
\par\vspace{0.35em}
\parbox{0.98\textwidth}{\footnotesize\raggedright
The values in parentheses indicate the standard deviation over 30 independent runs. \textnormal{N/A} follows the availability convention above. For available methods that do not explicitly maintain learned nodes or prototypes, \#Nodes is not applicable and is shown as \texttt{--}. Anuran Calls uses only the Family label.}
\end{table*}

\begin{sidewaystable}[p]
\centering
\scriptsize
\setlength{\tabcolsep}{3pt}
\renewcommand{\arraystretch}{0.95}
\caption{Average computation time (sec.) for the stationary setting.}
\label{tab:app_fit_time_stationary}
\resizebox{\textheight}{!}{%
\begin{tabular}{lccccccccccccc}
\toprule
Dataset & GNG & SOINN+ & DDVFA & CAEA & TBM-P & CAE & IDAT & TC & SR-PCM-HDP & TableDC & G-CEALS & AuToMATo & PHIDA \\
\midrule
Iris & 0.031 (0.014) & 0.499 (0.352) & 0.101 (0.066) & 0.009 (0.005) & 0.313 (0.191) & 0.060 (0.033) & 0.573 (0.397) & 0.073 (0.000) & 0.137 (0.066) & 0.281 (0.014) & 1.055 (0.260) & 0.364 (0.009) & 0.037 (0.003) \\
\midrule
Seeds & 0.003 (0.001) & 0.038 (0.009) & 0.035 (0.015) & 0.007 (0.003) & 0.041 (0.015) & 0.045 (0.019) & 0.125 (0.093) & 0.029 (0.000) & 0.040 (0.007) & 0.271 (0.013) & 0.953 (0.078) & 0.505 (0.010) & 0.052 (0.004) \\
\midrule
Dermatology & 0.006 (0.001) & 0.089 (0.045) & 0.061 (0.023) & 0.008 (0.004) & 0.205 (0.059) & 0.169 (0.096) & 0.015 (0.004) & 0.032 (0.000) & 0.160 (0.026) & 0.277 (0.010) & 1.812 (0.070) & 1.316 (0.019) & 0.100 (0.004) \\
\midrule
Pima & 0.016 (0.003) & 0.246 (0.083) & 0.989 (0.066) & 0.041 (0.005) & 0.076 (0.014) & 0.283 (0.299) & 0.202 (0.143) & 0.065 (0.000) & 0.119 (0.007) & 0.349 (0.007) & 2.893 (0.077) & 2.366 (0.090) & 0.272 (0.012) \\
\midrule
Mice Protein & 0.013 (0.001) & 0.138 (0.017) & 3.163 (0.369) & 0.120 (0.033) & 0.407 (0.220) & 6.820 (3.908) & 0.037 (0.012) & 0.124 (0.000) & 0.611 (0.027) & 0.395 (0.004) & 6.084 (0.544) & 15.748 (0.179) & 0.317 (0.036) \\
\midrule
Binalpha & 0.450 (0.113) & 0.240 (0.054) & 5.778 (0.388) & 0.470 (0.073) & 1.446 (0.253) & 1.665 (0.756) & 0.126 (0.041) & 0.174 (0.000) & 2.617 (0.127) & 0.459 (0.003) & 5.547 (0.072) & 346.921 (21.271) & 0.737 (0.445) \\
\midrule
Yeast & 0.184 (0.009) & 0.201 (0.026) & 0.295 (0.025) & 0.016 (0.002) & 0.182 (0.018) & 0.090 (0.055) & 0.235 (0.137) & 0.155 (0.000) & 0.479 (0.058) & 0.434 (0.007) & 6.137 (0.644) & 8.167 (0.072) & 0.339 (0.014) \\
\midrule
Semeion & 0.348 (0.101) & 0.237 (0.059) & 35.812 (7.627) & 0.588 (0.083) & 1.603 (0.287) & 3.704 (0.160) & 0.454 (0.159) & 0.169 (0.000) & 2.440 (0.143) & 0.464 (0.002) & 6.452 (0.053) & 280.942 (14.404) & 1.026 (0.471) \\
\midrule
MSRA25 & 0.193 (0.095) & 0.353 (0.046) & 3.138 (0.134) & 0.060 (0.015) & 0.884 (0.297) & 3.483 (1.391) & 0.089 (0.008) & 0.313 (0.000) & 3.822 (0.207) & 0.512 (0.002) & 7.304 (0.081) & 240.163 (17.759) & 0.952 (0.089) \\
\midrule
Image Seg. & 0.184 (0.048) & 0.336 (0.019) & 0.248 (0.028) & 0.025 (0.003) & 0.253 (0.018) & 0.102 (0.031) & 0.101 (0.035) & 0.620 (0.000) & 1.163 (0.049) & 0.569 (0.003) & 9.395 (0.867) & 16.275 (1.278) & 0.646 (0.039) \\
\midrule
Rice & 0.075 (0.055) & 1.987 (1.542) & 1.403 (0.394) & 0.150 (0.118) & 1.308 (0.764) & 32.109 (26.606) & 0.092 (0.013) & 1.926 (0.000) & 1.569 (0.242) & 0.756 (0.024) & 14.205 (0.937) & 9.989 (0.081) & 1.619 (0.109) \\
\midrule
TUANDROMD & 0.268 (0.018) & 1.015 (0.135) & 1.056 (0.095) & 1.130 (0.064) & 8.987 (0.591) & 0.274 (0.159) & 0.206 (0.049) & 0.757 (0.000) & 14.864 (1.931) & 0.903 (0.002) & 20.537 (2.177) & 858.196 (9.262) & 2.257 (0.205) \\
\midrule
Phoneme & 39.564 (0.889) & 0.848 (0.063) & 61.953 (3.914) & 0.044 (0.008) & 1.647 (0.111) & 0.155 (0.032) & 0.107 (0.005) & 3.364 (0.000) & 2.663 (0.605) & 1.005 (0.002) & 23.285 (2.764) & 22.671 (0.284) & 1.236 (0.032) \\
\midrule
Texture & 0.045 (0.008) & 0.778 (0.088) & 63.391 (3.181) & 0.087 (0.014) & 1.667 (0.172) & 0.496 (0.143) & 0.357 (0.120) & 2.527 (0.000) & 12.586 (1.832) & 1.021 (0.005) & 22.658 (0.208) & 141.755 (2.704) & 1.698 (0.149) \\
\midrule
OptDigits & 13.492 (0.375) & 0.675 (0.094) & 1.982 (0.186) & 0.095 (0.004) & 2.696 (0.469) & 2.789 (0.725) & 0.415 (0.071) & 2.375 (0.000) & 15.052 (2.037) & 1.047 (0.002) & 22.202 (2.859) & 876.142 (40.033) & 2.098 (0.161) \\
\midrule
Statlog & 0.042 (0.008) & 0.559 (0.050) & 134.121 (11.656) & 0.045 (0.007) & 6.734 (0.307) & 0.658 (0.300) & 0.162 (0.008) & 2.462 (0.000) & 14.306 (2.337) & 1.155 (0.002) & 24.203 (0.145) & 295.601 (5.468) & 2.090 (0.138) \\
\midrule
Anuran Calls & 0.043 (0.006) & 0.676 (0.086) & 14.536 (1.064) & 0.249 (0.053) & 10.788 (0.394) & 0.767 (0.330) & 0.159 (0.008) & 2.802 (0.000) & 6.619 (0.533) & 1.221 (0.002) & 28.387 (0.429) & 165.177 (1.274) & 1.756 (0.039) \\
\midrule
Isolet & 0.926 (0.318) & 3.663 (0.579) & 871.905 (11.956) & 5.025 (0.198) & 141.844 (7.740) & 155.178 (38.635) & 2.092 (0.488) & 5.677 (0.000) & 216.715 (14.729) & 1.519 (0.005) & 29.301 (1.723) & N/A & 21.702 (5.596) \\
\midrule
STL10 & 3.346 (0.365) & 11.476 (3.068) & 4849.056 (417.613) & 13.662 (0.414) & 1598.144 (17.633) & 206.278 (6.984) & 14.605 (0.930) & 8.786 (0.000) & 1930.724 (17.000) & 3.033 (0.001) & 55.165 (6.192) & N/A & 195.967 (33.582) \\
\midrule
MNIST10K & 440.067 (2.370) & 27.055 (7.210) & 7.179 (1.586) & 19.165 (0.948) & 476.371 (39.722) & 64.610 (2.548) & 15.450 (1.550) & 7.958 (0.000) & 2379.645 (40.314) & 3.107 (0.002) & 49.721 (0.638) & N/A & 308.211 (31.506) \\
\midrule
PenBased & 0.285 (0.028) & 1.410 (0.138) & 178.398 (7.010) & 0.371 (0.013) & 9.702 (0.474) & 0.956 (0.222) & 0.655 (0.526) & 10.985 (0.000) & 16.443 (1.030) & 1.782 (0.002) & 43.283 (4.565) & 154.448 (0.229) & 2.584 (0.037) \\
\midrule
Letter & 0.864 (0.108) & 4.630 (0.569) & 611.091 (11.083) & 0.864 (0.060) & 177.996 (5.306) & 4.285 (1.737) & 0.617 (0.136) & 41.642 (0.000) & 61.473 (3.188) & 3.256 (0.002) & 89.733 (4.372) & 836.984 (1.324) & 4.664 (0.105) \\
\midrule
Shuttle & 370.124 (4.658) & 14.224 (2.003) & 9.972 (0.639) & 1.048 (0.037) & 153.415 (11.385) & 2.908 (1.203) & 1.530 (0.182) & 411.928 (0.000) & 309.148 (5.600) & 8.368 (0.004) & 210.079 (1.400) & 697.210 (1.565) & 27.408 (1.885) \\
\midrule
Skin & 1630.409 (39.423) & 48.444 (6.790) & N/A & 11.033 (0.675) & N/A & 4.584 (0.611) & 5.041 (0.236) & N/A & 4295.890 (24.384) & 34.038 (0.014) & 1047.524 (7.299) & 1074.300 (5.385) & 82.802 (2.306) \\
\bottomrule
\end{tabular}%
}
\par\vspace{0.35em}
\parbox{0.98\textheight}{\footnotesize\raggedright
The values in parentheses indicate the standard deviation over 30 independent runs. Times are reported in seconds. \textnormal{N/A} follows the availability convention above and is not treated as zero runtime.}
\end{sidewaystable}

\begin{table*}[htbp]
\centering
\scriptsize
\setlength{\tabcolsep}{3pt}
\renewcommand{\arraystretch}{0.95}
\caption{Average computation time (sec.) for the nonstationary setting.}
\label{tab:app_fit_time_nonstationary}
\resizebox{\textwidth}{!}{%
\begin{tabular}{lccccccc}
\toprule
Dataset & GNG & SOINN+ & DDVFA & CAEA & CAE & IDAT & PHIDA \\
\midrule
Iris & 0.005 (0.003) & 0.421 (0.209) & 0.054 (0.028) & 0.002 (0.001) & 0.053 (0.025) & 0.070 (0.020) & 0.040 (0.002) \\
\midrule
Seeds & 0.005 (0.003) & 0.040 (0.010) & 0.036 (0.025) & 0.002 (0.001) & 0.061 (0.011) & 0.011 (0.003) & 0.058 (0.003) \\
\midrule
Dermatology & 0.017 (0.006) & 0.097 (0.038) & 0.042 (0.004) & 0.007 (0.001) & 0.282 (0.013) & 0.016 (0.004) & 0.117 (0.009) \\
\midrule
Pima & 0.051 (0.005) & 0.223 (0.074) & 0.124 (0.006) & 0.020 (0.002) & 0.063 (0.010) & 0.026 (0.008) & 0.267 (0.005) \\
\midrule
Mice Protein & 0.152 (0.012) & 0.245 (0.023) & 0.266 (0.258) & 0.162 (0.006) & 6.880 (3.909) & 0.056 (0.017) & 0.557 (0.034) \\
\midrule
Binalpha & 0.533 (0.050) & 0.886 (0.088) & 0.788 (0.048) & 0.419 (0.020) & 3.721 (1.589) & 0.210 (0.045) & 1.692 (0.403) \\
\midrule
Yeast & 0.278 (0.021) & 0.233 (0.028) & 0.071 (0.003) & 0.035 (0.002) & 0.111 (0.070) & 0.057 (0.012) & 0.390 (0.034) \\
\midrule
Semeion & 0.631 (0.061) & 0.455 (0.059) & 2.709 (0.059) & 0.436 (0.014) & 3.334 (0.350) & 0.240 (0.014) & 2.106 (0.227) \\
\midrule
MSRA25 & 0.878 (0.062) & 0.856 (0.090) & 0.264 (0.017) & 1.052 (0.037) & 3.896 (2.080) & 0.137 (0.010) & 1.859 (0.192) \\
\midrule
Image Seg. & 1.099 (0.060) & 0.540 (0.053) & 0.073 (0.004) & 0.079 (0.004) & 0.099 (0.022) & 0.090 (0.024) & 1.022 (0.145) \\
\midrule
Rice & 10.576 (0.228) & 2.019 (1.685) & 0.749 (0.562) & 0.179 (0.140) & 59.640 (2.063) & 0.084 (0.009) & 1.913 (0.183) \\
\midrule
TUANDROMD & 4.527 (6.555) & 1.090 (0.103) & 0.267 (0.017) & 0.628 (0.071) & 0.310 (0.271) & 0.189 (0.022) & 2.972 (0.467) \\
\midrule
Phoneme & 15.238 (0.849) & 0.809 (0.039) & 0.928 (0.024) & 0.092 (0.004) & 0.119 (0.023) & 0.110 (0.006) & 1.263 (0.025) \\
\midrule
Texture & 16.608 (0.722) & 2.712 (0.291) & 0.248 (0.015) & 0.651 (0.065) & 0.791 (0.347) & 0.212 (0.023) & 2.992 (0.439) \\
\midrule
OptDigits & 18.570 (0.692) & 2.485 (0.252) & 1.714 (0.046) & 0.392 (0.082) & 5.377 (1.924) & 0.314 (0.030) & 3.250 (0.413) \\
\midrule
Statlog & 26.680 (0.730) & 1.008 (0.209) & 4.125 (0.072) & 1.097 (0.092) & 1.633 (1.439) & 0.194 (0.020) & 2.850 (0.520) \\
\midrule
Anuran Calls & 579.941 (4.223) & 0.960 (0.172) & 0.904 (0.124) & 0.291 (0.022) & 0.695 (0.249) & 0.173 (0.015) & 2.084 (0.162) \\
\midrule
Isolet & 55.666 (2.321) & 17.564 (1.602) & 15.692 (0.267) & 11.562 (0.293) & 93.732 (2.741) & 3.051 (0.341) & 44.209 (10.237) \\
\midrule
STL10 & 221.722 (6.743) & 27.859 (4.981) & 0.000 (0.000) & 83.028 (2.541) & 0.000 (0.000) & 14.797 (1.521) & 212.651 (37.478) \\
\midrule
MNIST10K & 266.191 (5.445) & 171.092 (39.817) & 0.000 (0.000) & 154.310 (30.070) & 155.613 (39.656) & 35.186 (4.369) & 545.362 (45.349) \\
\midrule
PenBased & 71.329 (1.070) & 7.230 (0.922) & 1.045 (0.020) & 2.107 (0.204) & 2.042 (1.421) & 0.885 (0.617) & 8.706 (2.113) \\
\midrule
Letter & 161.541 (1.351) & 99.775 (11.068) & 2.766 (0.058) & 1.882 (0.558) & 2.801 (1.178) & 0.853 (0.124) & 18.843 (3.760) \\
\midrule
Shuttle & 519.634 (34.046) & 38.195 (17.225) & 1.947 (0.052) & 2.958 (0.384) & 2.782 (1.028) & 2.351 (0.506) & 58.207 (13.195) \\
\midrule
Skin & N/A & 52.234 (1.809) & N/A & 2.359 (1.122) & 6.461 (3.449) & 5.553 (0.371) & 105.757 (2.332) \\
\bottomrule
\end{tabular}%
}
\par\vspace{0.35em}
\parbox{0.98\textwidth}{\footnotesize\raggedright
The values in parentheses indicate the standard deviation over 30 independent runs. Times are reported in seconds. \textnormal{N/A} follows the availability convention above and is not treated as zero runtime.}
\end{table*}

\clearpage
\subsection{Discussion by Dataset}
\label{app:structure_quality_tradeoffs}

Tables~\ref{tab:app_nodes_clusters_stationary} and
\ref{tab:app_nodes_clusters_nonstationary} characterize counts of learned nodes and the number of clusters produced at the final evaluation point, while
Tables~\ref{tab:app_stationary_quality},
\ref{tab:app_nonstationary_final_quality}, and
\ref{tab:app_nonstationary_streamwise} report the corresponding
ARI and AMI values. These auxiliary quantities should therefore be interpreted jointly with these values, not as standalone superiority criteria.

PHIDA exhibits an informative tradeoff between ARI and AMI values,
counts of learned nodes, and the number of clusters produced at the final
evaluation point. Relative to IDAT, PHIDA tends to use more learned nodes
but often yields fewer clusters at the final evaluation point. Compared with competitors that maintain large graphs, such as SOINN+, GNG, CAE, and CAEA, PHIDA
often has a smaller number of clusters produced at the final evaluation
point under the reported structural summaries. This behavior is useful
when the resulting ARI and AMI values also improve, because it indicates that
PHIDA obtains a smaller final partition without simply increasing the
number of clusters produced at the final evaluation point. When an increased
number of learned nodes is not accompanied by higher ARI or AMI values, the
same behavior should instead be read as part of the tradeoff between
ARI and AMI values, counts of learned nodes, and the number of clusters
produced at the final evaluation point, not as evidence of superiority.

The descriptive cases in
Tables~\ref{tab:app_nodes_clusters_nonstationary},
\ref{tab:app_dataset_stats},
\ref{tab:app_nonstationary_final_quality}, and
\ref{tab:app_nonstationary_streamwise} suggest a scope condition for
the present PHIDA design. PHIDA favors fewer output clusters based on support modes
over very fine local graph partitioning. This is beneficial
when groups of learned nodes align with the external classes. It may
be less favorable when the benchmark labels require finer local boundary
resolution or do not align cleanly with modes based on node support density. For
example, Letter suggests a possible case where AMI may reflect under resolution.
PHIDA produces fewer clusters at the final evaluation point than the 26 external labels, whereas
several stronger AMI competitors preserve substantially finer output
structures. Shuttle and Skin suggest regimes where finer local structure
is beneficial. The similar final cluster counts on Phoneme suggest that the loss there is unlikely to be explained by output
granularity alone and may instead reflect limited alignment between modes based on node support density and the external label structure. These
observations are descriptive rather than causal. Ground-truth labels are used only for supplementary interpretation.

\clearpage
\subsection{Additional Critical-Difference Diagrams}
\label{app:additional_figures}

\subsubsection{Stationary Ablation Critical-Difference Diagrams}
\label{app:ablation_cd_stationary}

\begin{figure}[htbp]
  \centering
  \begin{subfigure}[t]{0.48\textwidth}
    \centering
    \includegraphics[width=\linewidth]{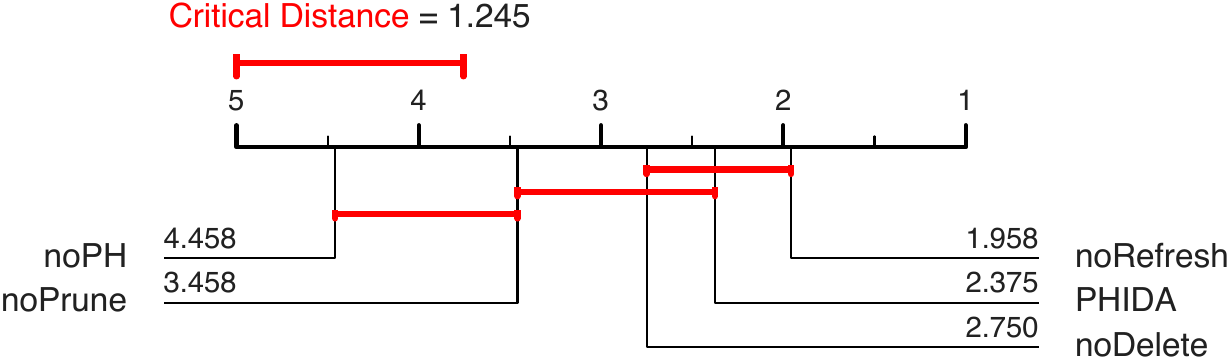}
    \caption{Final ARI}
  \end{subfigure}\hfill
  \begin{subfigure}[t]{0.48\textwidth}
    \centering
    \includegraphics[width=\linewidth]{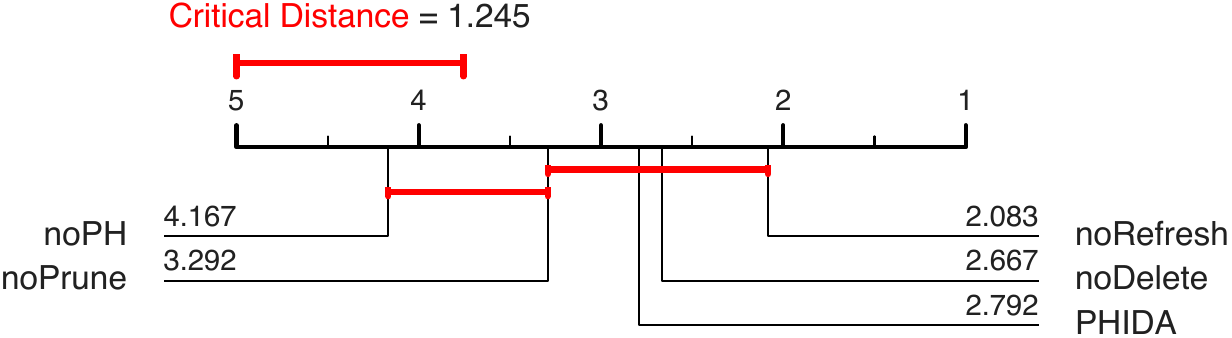}
    \caption{Final AMI}
  \end{subfigure}
  \caption{
  Stationary ablation critical-difference diagrams.
  Lower average rank is better.
  The stationary setting is used as supplementary evidence.
  The noPH variant remains the weakest average rank profile, while noRefresh ranks highly in the stationary setting because periodic PH component view refresh is primarily motivated by evolving streams.
  Ablation switches are defined in Appendix~\ref{app:ablation_variant_protocol}.
  }
\label{fig:ablation_cd_stationary}
\end{figure}

Figure~\ref{fig:ablation_cd_stationary} reports stationary ablation critical-difference diagrams as supplementary figures. The main ablation discussion focuses on the primary nonstationary protocol.

\subsubsection{Nonstationary Backward Transfer Critical-Difference Diagrams}
\label{app:bwt_cd_nonstationary}

\begin{figure}[htbp]
\centering
\begin{subfigure}[t]{0.48\textwidth}
  \centering
  \includegraphics[width=\linewidth]{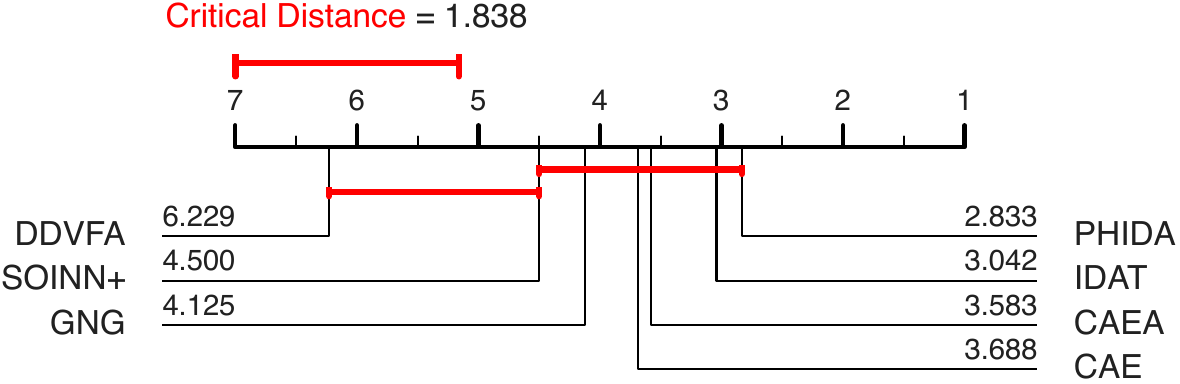}
  \caption{BWT\_ARI}
\end{subfigure}\hfill
\begin{subfigure}[t]{0.48\textwidth}
  \centering
  \includegraphics[width=\linewidth]{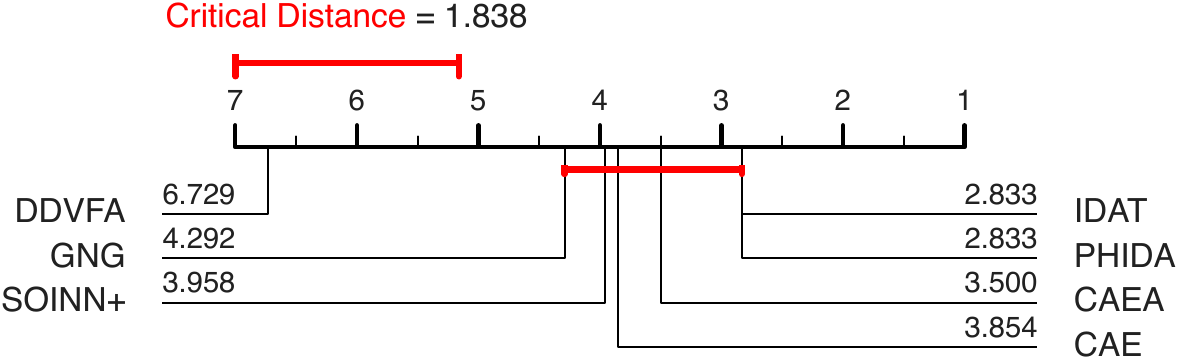}
  \caption{BWT\_AMI}
\end{subfigure}
\caption{Critical difference diagrams based on BWT\_ARI and BWT\_AMI in the nonstationary setting. Smaller average rank is better.}
\label{fig:cd_nonstationary_bwt}
\end{figure}

Figure~\ref{fig:cd_nonstationary_bwt} shows the supplementary nonstationary backward transfer rank overview associated with Table~\ref{tab:app_nonstationary_bwt} of Appendix~\ref{app:benchmark_tables}.

\subsubsection{CD Diagrams for Mapping Comparisons Based on the IDA Component}
\label{app:ida_component_mapping_control_cd}

This subsection reports CD diagrams for additional mapping comparisons based on the IDA component. These methods apply Ward, HDBSCAN, or K-means mappings to the state of learned nodes produced without the PH component view. Ward and K-means use the true number of classes directly. HDBSCAN uses the true number of classes to choose the hierarchy cut. These methods are additional comparisons over states of learned nodes and mapping rules rather than internal PHIDA ablations. They are not baselines evaluated without using the true number of classes. Lower average rank is better in all diagrams.

\begin{figure}[htbp]
  \centering
  \begin{subfigure}[t]{0.48\textwidth}
    \centering
    \includegraphics[width=\linewidth]{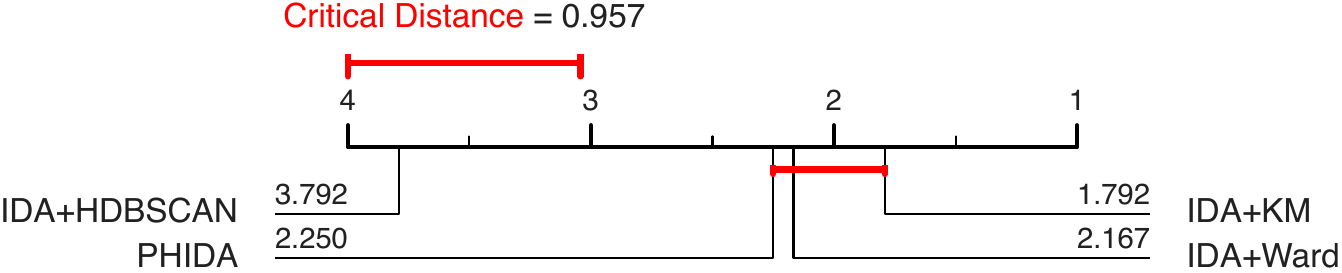}
    \caption{Final ARI}
  \end{subfigure}\hfill
  \begin{subfigure}[t]{0.48\textwidth}
    \centering
    \includegraphics[width=\linewidth]{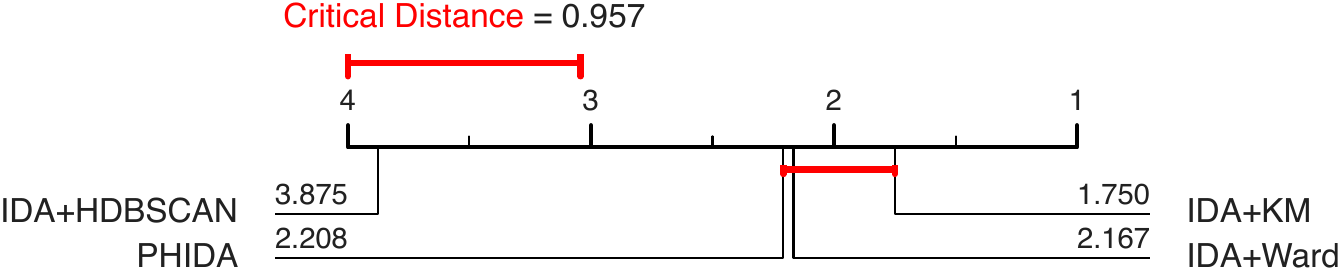}
    \caption{Final AMI}
  \end{subfigure}
  \caption{Stationary CD diagrams for mapping comparisons based on the IDA component. IDA+Ward, IDA+HDBSCAN, and IDA+KM use the IDA component without the PH component view and then apply mappings to the representatives of learned nodes. Ward and K-means use the true number of classes directly. HDBSCAN uses the true number of classes to choose a hierarchy cut. Lower average rank is better.}
  \label{fig:cd_ida_component_mapping_controls_stationary}
\end{figure}

Figure~\ref{fig:cd_ida_component_mapping_controls_stationary} shows the stationary CD diagrams for the mapping comparisons based on the IDA component using final ARI and final AMI.

\begin{figure*}[htbp]
  \centering
  \begin{subfigure}[t]{0.48\textwidth}
    \centering
    \includegraphics[width=\linewidth]{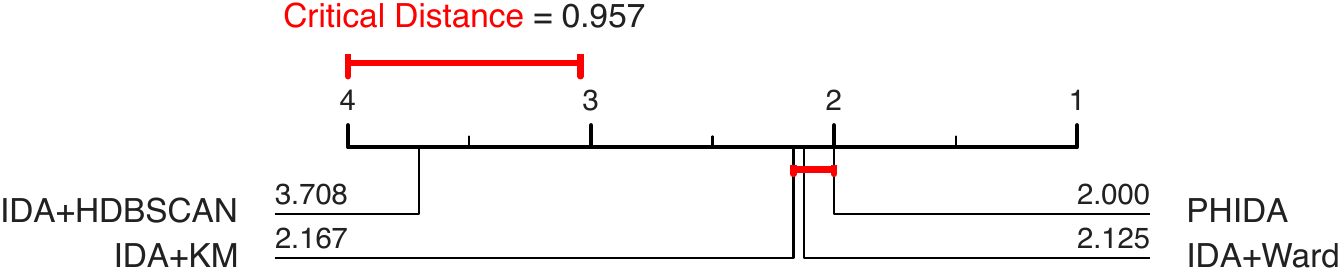}
    \caption{Final ARI}
  \end{subfigure}\hfill
  \begin{subfigure}[t]{0.48\textwidth}
    \centering
    \includegraphics[width=\linewidth]{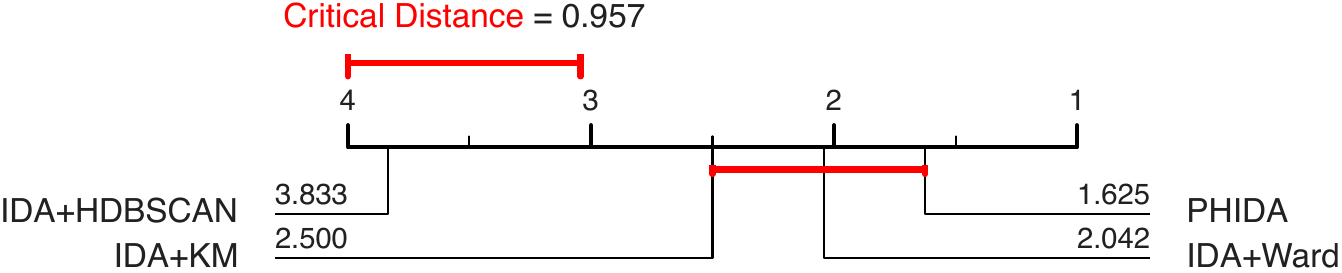}
    \caption{Final AMI}
  \end{subfigure}

  \vspace{0.6em}

  \begin{subfigure}[t]{0.48\textwidth}
    \centering
    \includegraphics[width=\linewidth]{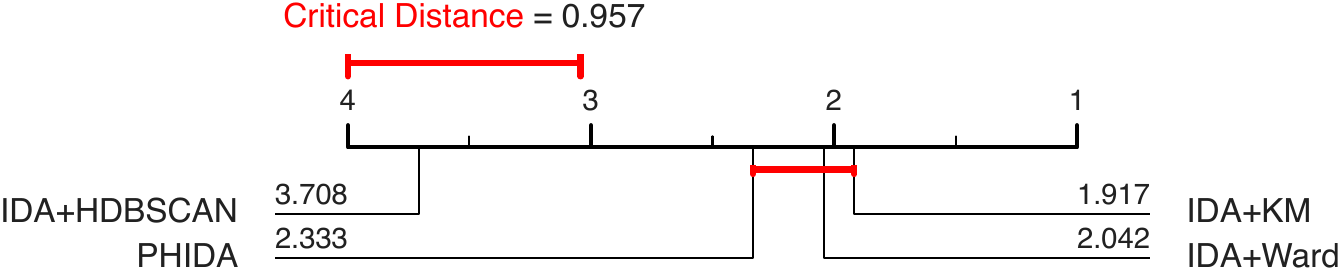}
    \caption{avgInc\_ARI}
  \end{subfigure}\hfill
  \begin{subfigure}[t]{0.48\textwidth}
    \centering
    \includegraphics[width=\linewidth]{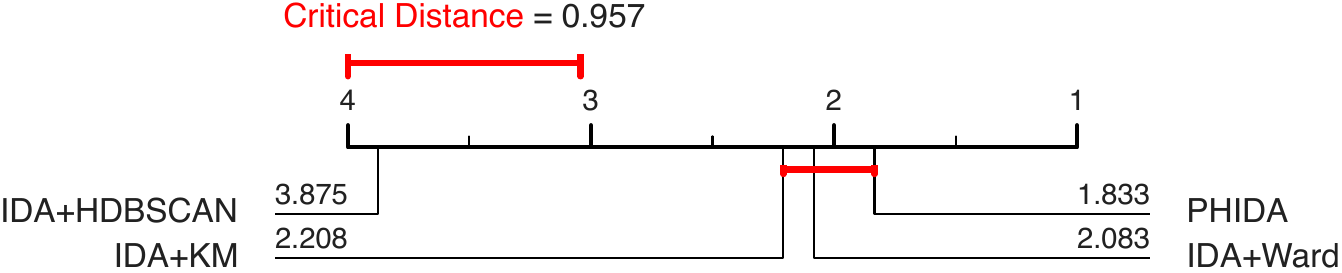}
    \caption{avgInc\_AMI}
  \end{subfigure}
  \caption{Nonstationary CD diagrams for mapping comparisons based on the IDA component. The target number of clusters at each evaluation stage is the number of classes observed up to that stage. IDA+Ward and IDA+KM use this target directly. IDA+HDBSCAN uses the closest available hierarchy cut with at least the target number of clusters when such a cut exists and otherwise uses the closest available cut. Lower average rank is better.}
  \label{fig:cd_ida_component_mapping_controls_nonstationary}
\end{figure*}

Figure~\ref{fig:cd_ida_component_mapping_controls_nonstationary} shows the nonstationary CD diagrams for the mapping comparisons based on the IDA component using final ARI, final AMI, avgInc\_ARI, and avgInc\_AMI.

The mapping comparisons test whether conventional mapping applied after node construction by IDA matches PHIDA when the conventional mappings are given the true number of classes. The CD diagrams give a mixed but informative result. PHIDA has the best average rank for nonstationary final ARI, nonstationary final AMI, and nonstationary avgInc\_AMI, while IDA+KM has the best average rank for nonstationary avgInc\_ARI and for the two stationary summaries. IDA+HDBSCAN has higher average ranks than IDA+Ward and IDA+KM across these diagrams. Because the additional methods use the true number of classes, their competitiveness is expected. The main observation is that conventional mappings with this extra information on the state of learned nodes produced by the IDA component without the PH component view do not uniformly dominate PHIDA, and PHIDA retains favorable average ranks in several nonstationary summaries. These targeted comparisons complement the internal PHIDA ablations by assessing conventional mappings on the IDA state of learned nodes.

\section{Code, Compute Resources, Assets, and Broader Impacts}
\label{app:code_compute_assets_impacts}

\paragraph{Code and data access.}
A supplementary code package provides the PHIDA implementation with its IDA node learning component, the common evaluation runner, utilities for continual metrics, README/instructions, and scripts for generating results. The package also provides the wrappers or reimplementations used to execute the compared algorithms when redistribution is permitted. Otherwise, it provides acquisition and execution instructions consistent with the original source terms. The code package does not claim ownership of external benchmark datasets or baseline algorithms. Benchmark datasets are obtained from their public sources, such as dataset entries hosted on OpenML, UCI, KEEL, and GitHub repositories specific to each dataset, under the shared dataset collection and preprocessing assumptions. When redistribution is not covered by the original source terms, the package provides instructions for obtaining the data from the original public source rather than redistributing the data itself.

\paragraph{Compute resources.}
All reported experiments were executed through the common benchmark pipeline on an Apple M2 Ultra processor with 128GB RAM running macOS. The experiments used Python 3.10.7. The runner scripts use joblib/loky parallel execution, with up to 10 workers in the standard runner and up to 15 workers in the batch CSV runner. Unavailable entries include out-of-memory termination, timeout, absence of valid hyperparameters, or failure to build a predictive model.

\paragraph{Assets and licenses.}
The external assets used in this study are existing public benchmark datasets and baseline method descriptions or implementations. Dataset creators and repositories are credited through the benchmark references, such as dataset entries hosted on OpenML, UCI, and KEEL, and through the cited benchmark protocol. IDAT~\citep{masuyama25} is the published paired ART baseline with available code used as the reference method. Baseline methods are credited by citation, and any code obtained from GitHub or repositories provided by the method authors is used according to the license and terms specified by each original repository. The supplementary package distributes the PHIDA code and the common benchmark code used for reproducibility to the extent permitted by the relevant licenses. The paper does not claim ownership over external datasets, external baseline algorithms, or external code and respects the original source terms.

\paragraph{Broader impacts.}
PHIDA may help make online clustering more usable in streaming and nonstationary settings because it reduces manual design of output clusters and improves reproducibility of node-to-cluster mapping. Potential beneficial uses include exploratory analysis of evolving tabular data, monitoring, and scientific workflows where labels are limited. Possible negative impacts arise if unsupervised clusters are treated as ground-truth categories in high stakes decisions, or if biases present in benchmark data or similar deployment data are amplified by automatic grouping. PHIDA is not designed as a decision making system by itself, and applications that involve people, sensitive attributes, or consequential interventions should require domain validation, auditing, and human oversight.

\end{document}